\documentclass[10pt,twocolumn,letterpaper]{article}
\usepackage[pagenumbers]{cvpr} 

\AtBeginDocument{%
  }

\usepackage[dvipsnames]{xcolor}

\usepackage[ruled]{algorithm2e} 

\SetAlFnt{\small}
\SetAlCapFnt{\small}
\SetAlCapNameFnt{\small}
\SetAlCapHSkip{0pt}

\usepackage{booktabs} 

\usepackage{caption}
\usepackage{subcaption}
\usepackage{float}
\usepackage{esint} 
\usepackage{comment}
\usepackage{graphicx}
\usepackage{arydshln}
\usepackage{multirow}
\usepackage{rotating}

\expandafter\def\expandafter\normalsize\expandafter{%
    \normalsize%
    \setlength\abovedisplayskip{-8pt}
    \setlength\belowdisplayskip{2pt}
    \setlength\abovedisplayshortskip{-8pt}
    \setlength\belowdisplayshortskip{2pt}

    \setlength\floatsep{2pt}
    \setlength\dblfloatsep{0pt} 
    \setlength\intextsep{2pt} 
    \setlength\abovecaptionskip{2pt} 
    \setlength\belowcaptionskip{2pt} 
}

\newcommand{\rgb}{\textcolor{red}{R}\textcolor{teal}{G}\textcolor{blue}{B} }

\usepackage{xr}
\usepackage{ifthen}
\newboolean{boolIncludeAppendix}

\definecolor{cvprblue}{rgb}{0.21,0.49,0.74}
\usepackage[pagebackref,breaklinks,colorlinks,citecolor=cvprblue]{hyperref}

\begin{document}

\title{Towards Physically-Based Sky-Modeling For Image Based Lighting}

\author{
    Ian J.~Maquignaz \\
    Université Laval \\
    Québec, Québec \\
    Canada \\
    {\tt\small ian.maquignaz.1@ulaval.ca}
}


\twocolumn[{%
\renewcommand\twocolumn[1][]{#1}%
\vspace{-1.5cm}
\maketitle

\begin{center}
    \centering
    \vspace{-1cm}
    \captionsetup{type=figure}
    \includegraphics[width=0.8\textwidth,keepaspectratio]{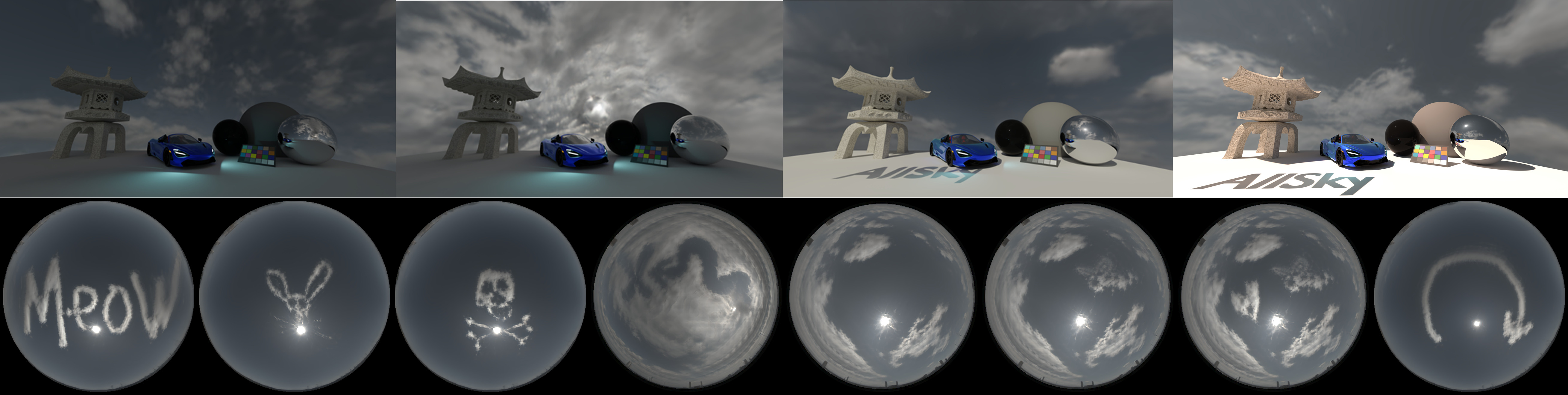}
    \captionof{figure}{IBL renders of AllSky environment maps and AllSky environment maps generated from user-drawn labels.}
\end{center}%
}]

\begin{abstract}
\vspace{-0.4cm}

Accurate environment maps are a key component for rendering photorealistic outdoor scenes with coherent illumination.
They enable captivating visual arts, immersive virtual reality, and a wide range of engineering and scientific applications.
Recent works have extended sky-models to be more comprehensive and inclusive of cloud formations but, as we demonstrate, existing methods fall short in faithfully recreating natural skies.
Though in recent years the visual quality of DNN-generated High Dynamic Range Imagery (HDRI) has greatly improved, the environment maps generated by DNN sky-models do not re-light scenes with the same tones, shadows, and illumination as physically captured HDR imagery.
In this work, we demonstrate progress in HDR literature to be tangential to sky-modelling as current works cannot support both photorealism and the 22 f-stops required for the Full Dynamic Range (FDR) of outdoor illumination.
We achieve this by proposing AllSky, a flexible all-weather sky-model learned directly from physically captured HDRI which we leverage to study the input modalities, tonemapping, conditioning, and evaluation of sky-models.
Per user-controlled positioning of the sun and cloud formations, AllSky expands on current functionality by allowing for intuitive user control over environment maps and achieves state-of-the-art sky-model performance.
Through our proposed evaluation, we demonstrate existing DNN sky-models are not interchangeable with physically captured HDRI or parametric sky-models, with current limitations being prohibitive of scalability and accurate illumination in downstream applications.
\end{abstract}

\vspace{-0.5cm}
\section{Introduction}
\label{sec:introduction}

Illumination is central to human perception of physical spaces and the visual quality of media and film \cite{lighting_lalonde24,lighting_lalonde21,Säks_2024}.
Early works modelling skydomes combined data from varied sources into pre-computed and parametric sky models for engineering and scientific applications, with the first sky models \cite{MOON_1940,PEREZ_1993} modelling only luminance.
With the advent of the digital age, a new paradigm of applications spurred interest in sky models to enable a wide range of emerging digital applications.
Nishita et al.~\cite{NISHITA_1993} proposed the first colour sky model enabling the generation of extraterrestrial views of the Earth for space flight simulators, and Image-Based Lighting (IBL) techniques~\cite{IBL} were proposed to render synthetic objects into real and virtual scenes.

Though conventional Low Dynamic Range (LDR) imagery is suitable for some applications, High Dynamic Range (HDR) ~\cite{HDR_IBL_BOOK} imagery is integral to sky-models and the capture of outdoor scenes.
HDR images capture a greater range of illumination and, in the particular case of outdoor lighting, can capture the estimated $22$ f-stops of exposure necessary for the highlights and shadows of an average real-world outdoor scene \cite{JENSEN_NightSky,HDR_IBL_BOOK,STUMPFEL_HDR_Sky_Capture}.
To distinguish the proportionality of scene exposure captured by an HDRI, we define the following:
\begin{enumerate}
     \item \textbf{Low Dynamic Range (LDR)} Imagery: Display-referenced images with compressed dynamic range which can be clipped and displayed in 8-bit colour.

    \item \textbf{High Dynamic Range (HDR)} Imagery: Scene-referenced measures of illumination with uncompressed dynamic range and precision greater than LDR 8-bit colour for later display as LDR. This includes imagery captured by conventional cameras in 12-bit RAW.

    \item \textbf{Extended Dynamic Range (EDR)} Imagery: HDR images captured using techniques such as LDR-bracketing for greater exposure range than a singular image from a conventional camera.

    \item \textbf{Full Dynamic Range (FDR)} Imagery / Physically-Captured Imagery: HDR images that \textbf{\textit{fully}}-capture the dynamic range of a reference scene without saturation of the exposure range.
\end{enumerate}

\begin{figure}[H]
    \centering
    \includegraphics[width=\linewidth,keepaspectratio]{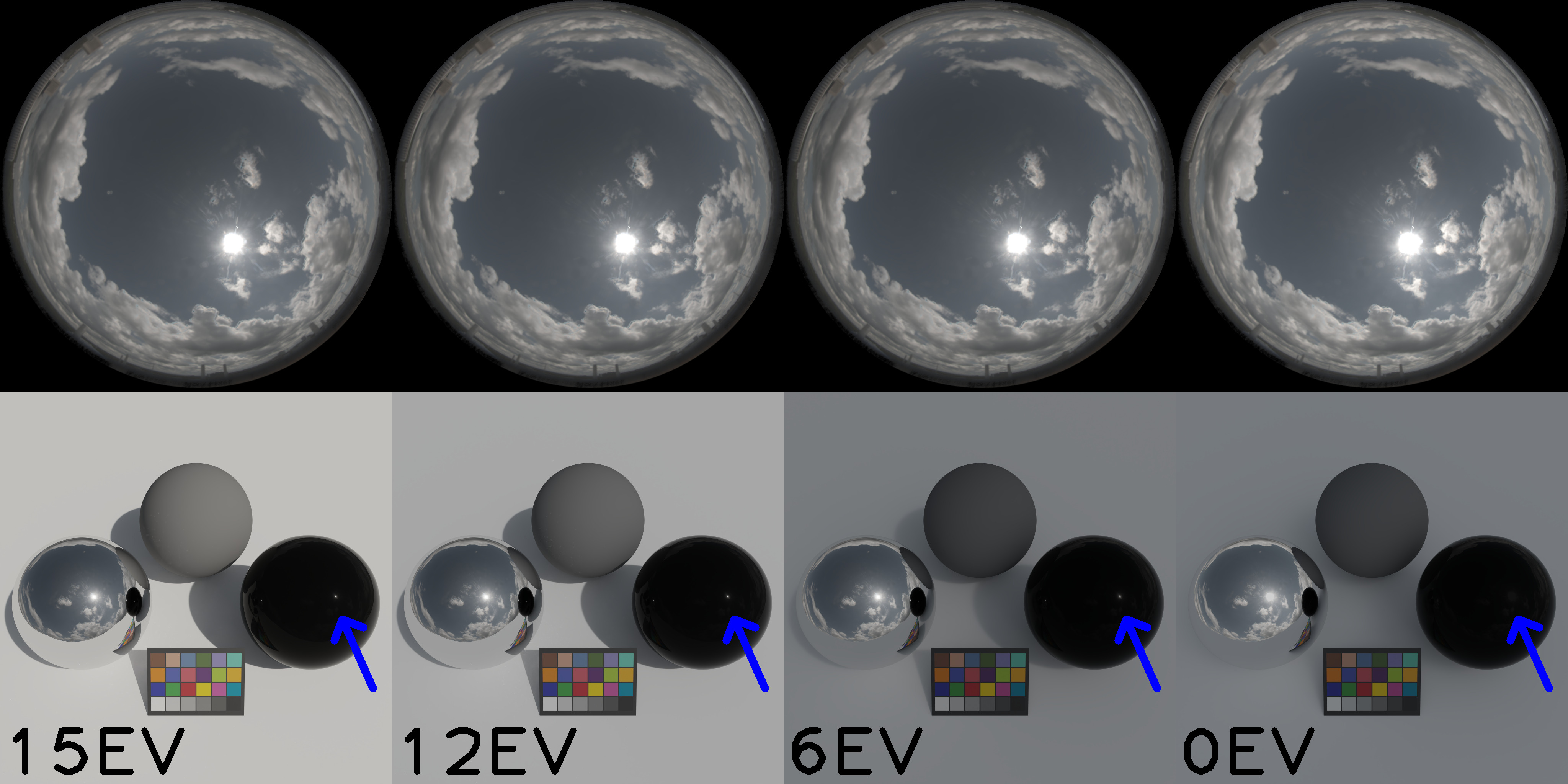}
    \caption{
    Each column illustrates the impact of an incremental clipping of exposure with exposure equalization to the 15EV ground truth.
    Though the environment-maps are visually unaltered, renderings have altered tones, shadows and light transmission (\textcolor{blue}{sun transmission}; black glass orb).
    }
    \label{fig:grid_demo_EV}
\end{figure}

This distinction between HDRIs is necessary given the challenges in applying HDR literature to sky-modeling \cite{Zhang_2017_ICCV,glowGAN, nerfInTheDark,hilliard2024360uformerhdrilluminationestimation,Phongthawee_2024_CVPR,HDR_from_LDR}.
In this work, we measure HDR as the Exposure Value (EV) of HDRI given $EV=log_2(|I|_{max} - |I|_{min} + 1)$, where $|I|$ is grayscale intensity.
We find that HDR literature generally does not provide sufficient detail to determine supported exposure ranges, focuses on exposure ranges $\leq6EV$ when real-world outdoor scenes are $\geq13EV$, omit daytime outdoor imagery in results and, if included, demonstrates saturated clouds and solar features.
Such limitations are problematic, given exposure range is key to the illumination provided by environment-maps and current sky-models struggle to reproduce the photorealism and illumination of real-world skies.
As demonstrated by \cref{fig:grid_demo_EV}, incrementally clipping the EV of an HDRI while equalizing exposure to the FDR 15EV ground truth results in visually indiscernible alterations to the environment-maps (\cref{fig:grid_demo_EV}, top), but significant alteration to illumination in IBL scenes through softer tones, shadows, and light transmission (\cref{fig:grid_demo_EV}, bottom).
Despite advances in HDR image generation using deep neural networks (DNNs), current sky-models struggle photorealistically reproduce the physical accuracy and illumination of real-world skies.
These limitations hinder their effectiveness in downstream applications such as IBL, where accurate lighting is crucial.

In this work, we propose AllSky, a conditional generative sky-model trained directly on physically captured FDR imagery and enabling user-controlled positioning of the sun and cloud formations.
We leverage this model to study the input modalities, tonemapping, conditioning, and evaluation of sky-models and propose a novel framework for evaluating DNN sky-models for coherence with real-world luminance.
We demonstrate AllSky achieves state-of-the-art (SOTA) illumination with sustained visual quality, and identify shortfalls in DNN sky-modeling which impede their scalability and interchangeability with physically captured imagery and parametric counterparts.
Our findings improve performance in downstream applications, but show recent improvements to DNN HDRI generation to be tangential to sky-modeling.
Significant work remains to make sky-models scalable to higher resolutions while sustaining the exposure range of the sun.

\section{Background}
\label{sec:background}

Physically-captured FDR images offer unsurpassed quality and photorealism for IBL renderings, but physical capture is difficult, labour-intensive \cite{KIDER_captureFramework} and the resulting imagery is both inflexible and of fixed temporal- and geo-locality.
Though HDR and EDR images can be suitable, partial-capture of skydome dynamic range generally saturates the solar spectrum in favor of capturing lucid cloud formations.
Sky-models have progressed over the past four decades to offer a wide range of accurate, versatile, and portable models providing artists, engineers and scientists a definitive cost-advantage to physical capture \cite{BRUNETON_2017_clearSky_eval}.
Early works included numerical and parametric models
\cite{PEREZ_1993,NISHITA_1996,PREETHAM_1999,ONEAL_2005,HABER_2005,BRUNETON_2008,ELEK_2010,HOSEK_13,HOSEK_13Sun,LM_2014,PSM_21} which generated clear and overcast skies via a limited set of user-parameters tied to physical properties but were unable to model atmospheric formations.

With the advent of deep learning, the fitting of analytical models to features and modalities can be automated with parameterization to a finite set of latent or user-intuitive parameters.
Notably, this concept has been proposed for lighting estimation \cite{YANNICK_2017, YANNICK_2019_SKYNET, ZHANG_2019_panonet, relighting_hdsky, Yu2021DualAA}, where LDR images can be used to guide the generation of HDR environment-maps for relighting virtual objects and scenes.

Recent deep learning approaches have been proposed to augment parametric clear-skies with photorealistic clouds.
CloudNet \cite{DEEPCLOUDS_22} proposed a method for augmenting Ho{\v{s}}ek-Wilkie~\cite{HOSEK_13} parametric skies per user-controlled cloud placement.
SkyGAN~\cite{SKYGAN_2022} proposed fitting the Prague Sky Model (PSM) ~\cite{PSM_21} model to real-world photographs, enabling the re-generation of PSM skies with cloud formations.
This formulation was also demonstrated by LM-GAN \cite{LM-GAN_2023}, fitting physical captures to the 11-parameter Lalonde-Matthews (LM) outdoor lighting model \cite{LM_2014} and producing photorealistic (though lower-resolution) weathered skies.
Most recently, Text2Light \cite{text2light} introduced textually controlled generation of indoor and outdoor environment-maps via diffusion, though its textual conditioning and diversity are dubious.

The objective of physically-accurate sky-models is to generate FDR environment-maps which emulate the illumination of clear (13EV+), cloudy and/or overcast daylight skies of FDR environment-maps, but many challenges remain to achieve this objective.
To mitigate problems modeling illumination, many works propose post-generation `tricks' to augment EDR to FDR solar illumination, including compositing with a parametric model sun and further aggregated inverse-tonemapping \cite{SKYGAN_2022, DEEPCLOUDS_22, text2light}.
Modelling and control over non-uniform atmospheric formations such as clouds remain a challenge and works that augment clear-skies with atmospheric formations are not proven to retain a tractable relationship to the parameters of the underlying clear sky-models.
A common alternative is to include complex cloud formations through volumetric cloud rendering \cite{DeepScatering,aasberg2024real} and cloud simulations \cite{BRUNETON_2008_clouds} post sky-model generation.
While this multi-step approach can be versatile and photorealistic, it is labour and computationally expensive \cite{SIGGRAPH_course_2020}.

As a result, the modeling of outdoor FDR imagery remains a challenge without a clear solution \cite{Apple_Paper}.
Regardless of the approach, the current limitations result in many situations where physical capture is unsurpassed and the preferable method to provide both photorealism, weather variations and illumination \cite{HOSEK_13Extrasolar, FORZA}.

\section{Methodology}
\label{sec:methodology}

We define our tonemapping operators, sky-model, and losses in terms of design considerations for sky-modelling.

\subsection{Tone-mapping Dynamic Range}
\begin{figure}[ht]
    \centering
    \includegraphics[width=\linewidth,keepaspectratio]{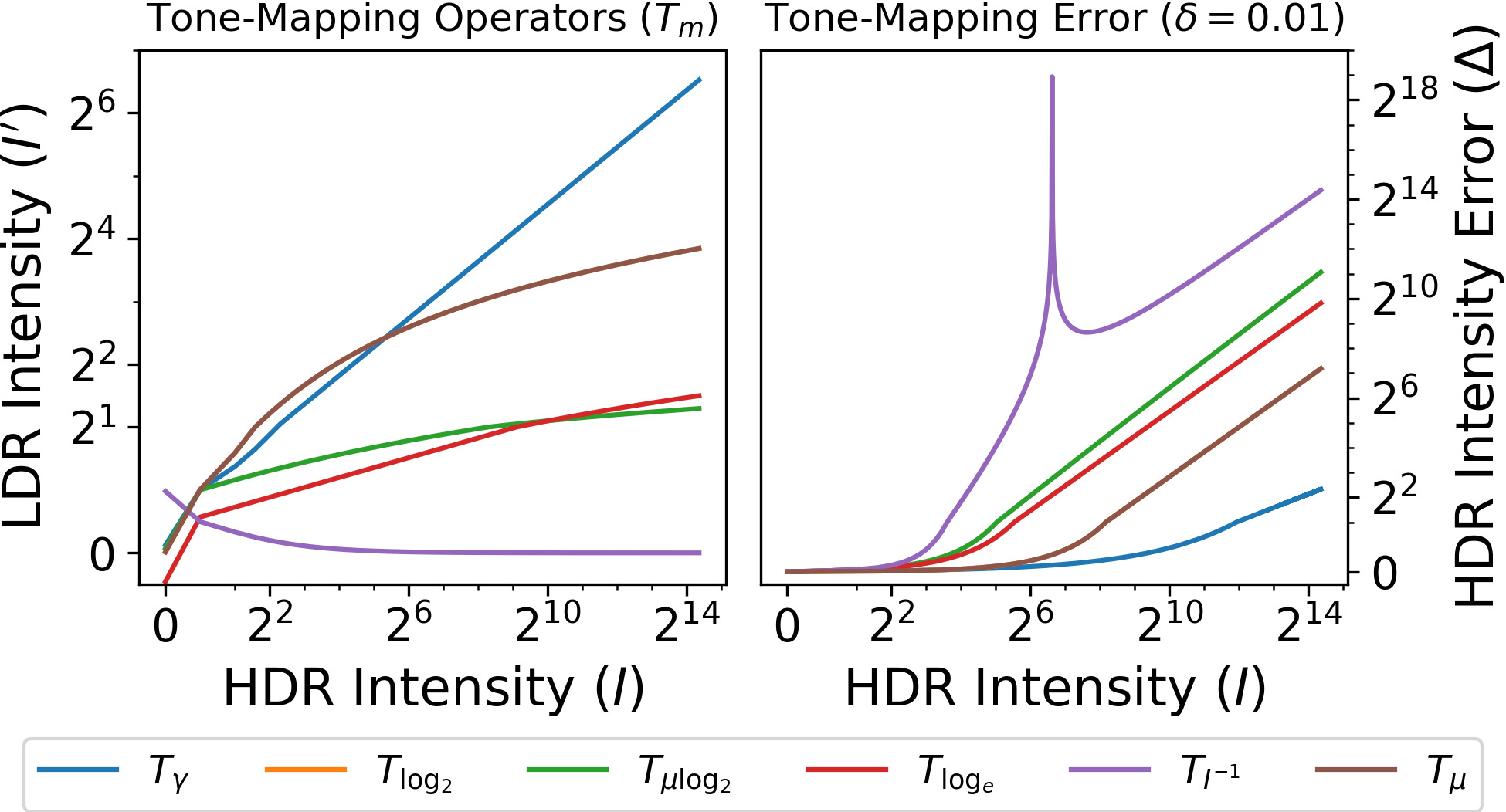}
    \caption{
    Tone-Mapping operator ($T_m$) compression (left) and respective non-linearity between error ($\delta$=$0.01$) in LDR compressed space and error ($\Delta$) in uncompressed HDR space (right).
    }
    \label{fig:plt_tm}
\end{figure}

Tone-mapping operators ($T_m$) compress HDRI to a visible, displayable, or latent colour-space more favourable for DNN training given $I' = {T_{m}}(I)$.
We investigate a range of operators, including:
Power-Law ($T_\gamma$),
logarithmic ($T_{log_n}$),
$\mu$-law ($T_\mu$),
and variations thereof as shown in \cref{app:sec::AllSky:::training} and \cref{fig:plt_tm} (left).
With the growing popularity of mixed-tone-mappers (combinations of tone-mappers) such as Hybrid Log-Gamma for display of 8K broadcasts \cite{HLG}, we propose $\mu$-lawLog$_2$ ($T_{\mu \log_2}$) to better preserve texture-rich cloud and skydome components (generally $EV \lessapprox 1$) and aggressively compress texture-poor high-exposure regions which are generally saturated in LDR imagery.

\begin{align}
\mu\text{-lawLog}_2 \text{: }& T_{\mu \log_2} \left(I\right) = \log_2\left[ \frac{\log_e(1+\mu I)}{\log_e(1.0+\mu)} +1\right] \label{eq:tm_muLawLog2}
\end{align}

Each operator is a bijection allowing for the recovery of the original image via $I = {T_{m}}^{-1}(I')$, but given $\Delta I = | I - {T_m}^{-1}({T_m}(I)-\delta)|$ these operators introduce a non-linearity between error ($\delta$) in LDR compressed space and error ($\Delta$) in uncompressed HDR space (right).
Most pronounced with the solar disk, a small error ($\delta$) in LDR space results in a large error ($\Delta$) in HDR space with a profound impact to illumination in IBL rendering.

\subsection{Model}
\begin{figure}[ht]
  \centering
  \includegraphics[width=\linewidth,keepaspectratio]{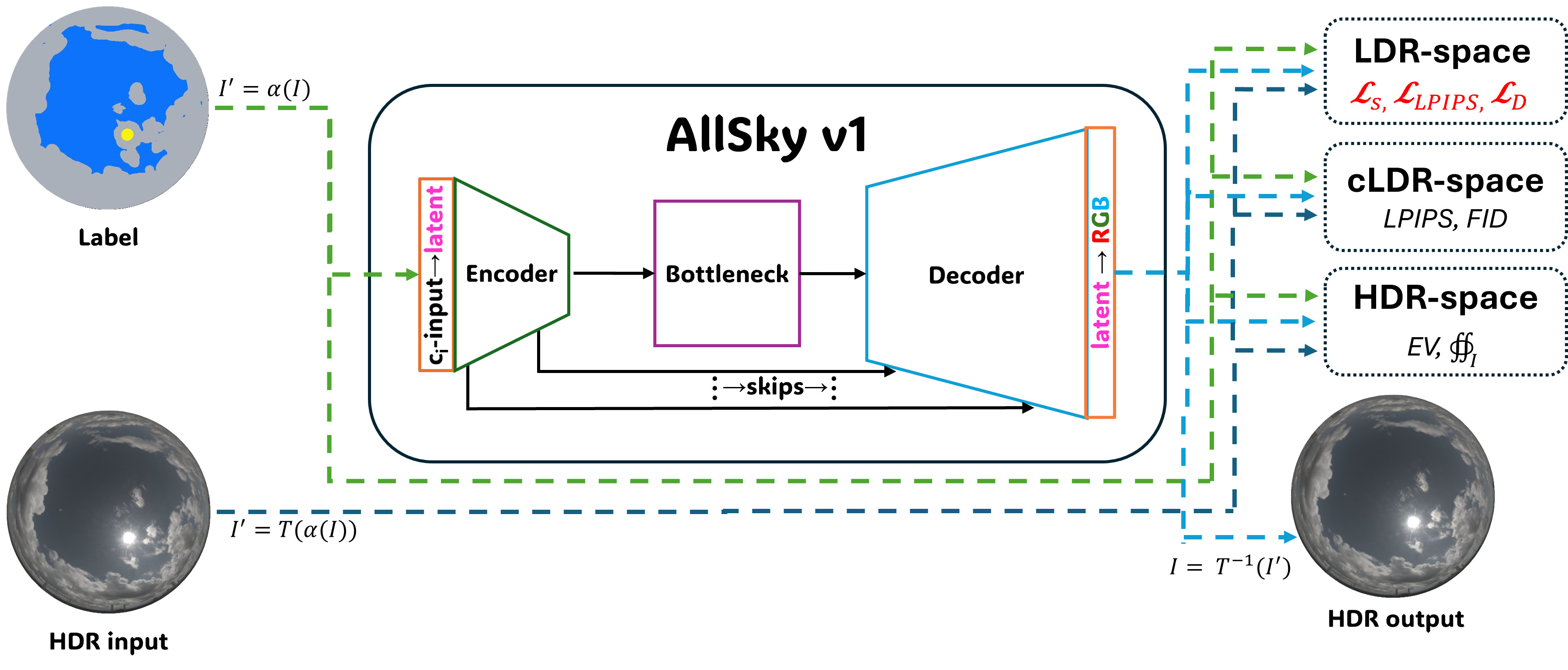}
    \caption{
        AllSky baseline architecture, accepting arbitrary discrete ($\mathbb{Z}^{+}$) or continuous ($\mathbb{R}$) $c_{i}$-channel input labels.
        The $c_{o}$-channel output can be evaluated by losses and metrics in $LDR$ compressed space, ${cLDR}$ compressed and clipped LDR-space, and ${HDR}$ inverse-tone-mapped linear space.
    }
    \label{fig:dia_AllSky}
    \vspace{-8pt}
\end{figure}

As a baseline and backbone, we implement UNetFixUp \cite{UNetFixup} as a configurable and proven U-Net architecture for outdoor image editing.
The model can be configured for arbitrary discrete ($\mathbb{Z}^{+}$) or continuous ($\mathbb{R}$) $c_{i}$-channel input and $c_{o}$-channel output.
We propose organizing sky-model metrics as shown in \cref{fig:dia_AllSky}, allowing dynamic-range-aware evaluation through losses and metrics in LDR compressed space, cLDR compressed and clipped LDR-space, and HDR inverse-tonemapped linear HDR-space.
Though losses can be placed in LDR-, cLDR- and/or HDR-space, choosing saturation points (clipping) is ambiguous and dependent on tonemapper selection, and HDR-space losses are prone to exploding gradients.

\subsection{Losses}
\label{sec:methodology::losses}

In experimentation, we observed that an $L_1$-loss on a dataset of clear-sky imagery guides the model toward a median clear-sky image with few visual artifacts, but IBL renderings exhibited poor illumination.
To explain this discrepancy, we explored sky-model losses and the guidance they provided given a dataset of FDR skydomes.

\begin{figure}[ht]
    \centering \includegraphics[width=0.7\linewidth,keepaspectratio]{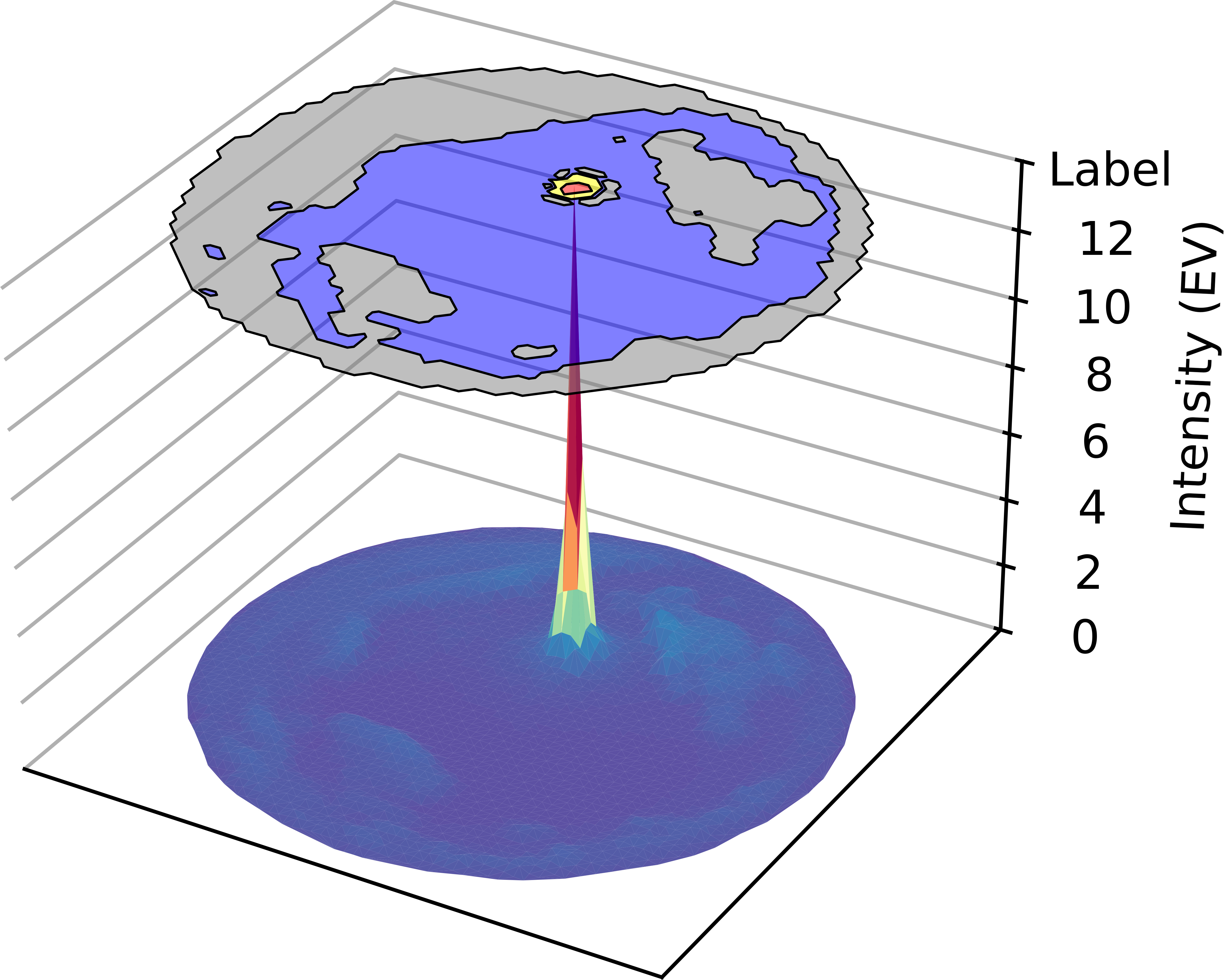}
    \caption{3D surface of skydome illumination, coloured from \textcolor{blue}{low} to \textcolor{red}{high} intensity. Disk above 3D surface illustrates skydome segmentation. The sun (\textcolor{red}{red} spike) is a small subset of pixels whose intensities belittle the remainder of the skydome. HDRDB sample from June 7th, 2016 at 1:54PM \cite{LavalHDRdb}.}
    \label{fig:plt_3D_surface}
\end{figure}

\begin{figure}[ht]
    \centering
    \includegraphics[width=\linewidth,keepaspectratio]{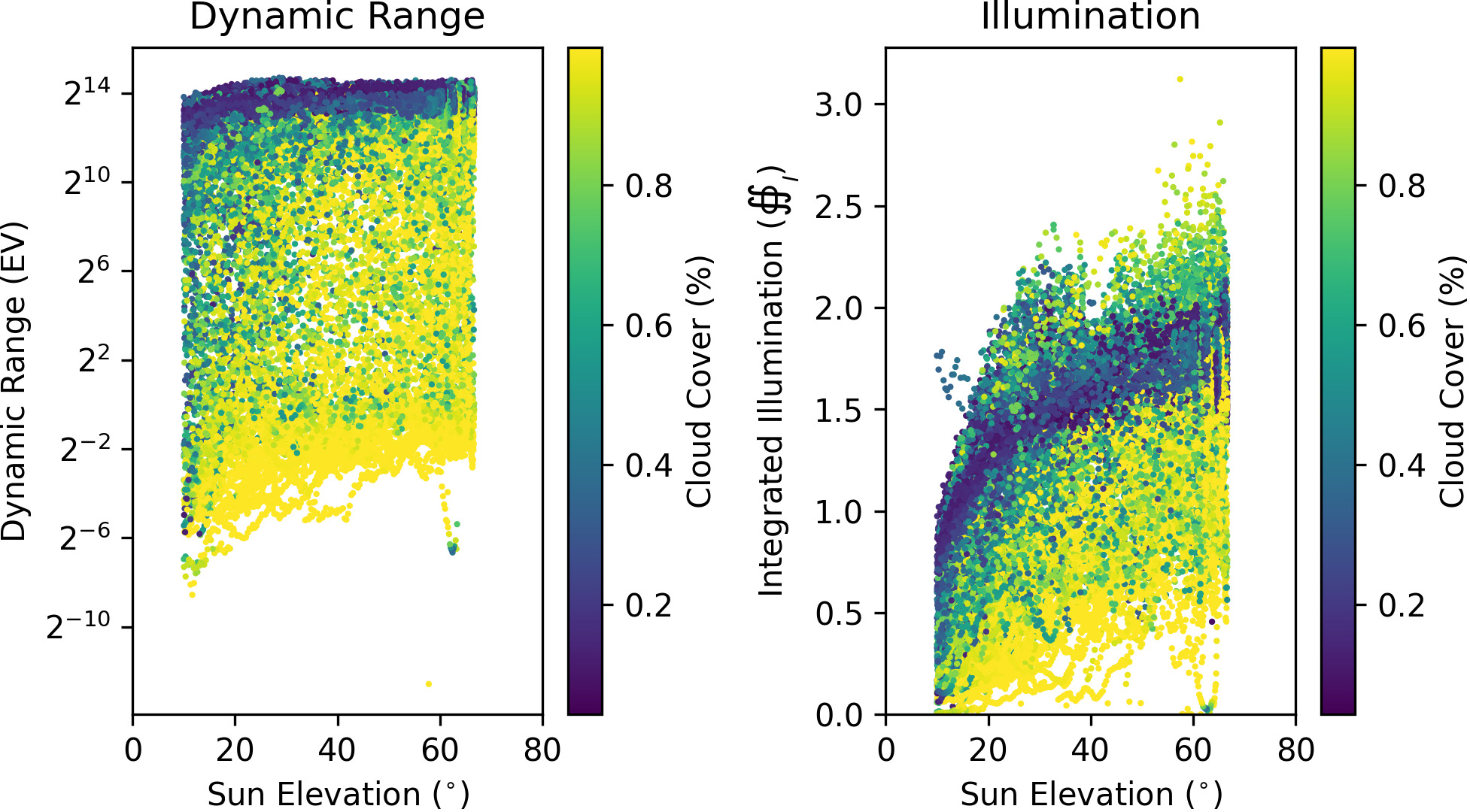}
    \caption{Distribution of exposure range and illumination by sun elevation and cloud cover in HDRDB \cite{LavalHDRdb}.}
    \label{fig:plt_illumination}
    \vspace{-16pt}
\end{figure}

\subsubsection{Supervised Losses}
\label{sec:methodology::losses:::supervised}

As shown in \cref{fig:plt_3D_surface}, the sun (\textcolor{red}{red}) is represented by a small subset of pixels whose intensities belittle the remainder of the skydome (\textcolor{blue}{blue}/\textcolor{teal}{green}).
The four-pixels solar region with intensity $EV{\geq}11$ of the $512^2$ skydome represents 55\% of the environment-maps' illumination.
The remaining 45\% of illumination is provided by the remaining 99\%+ of pixels with intensities $EV{\leq}8$, representing an amalgamation of refractive mediums (i.e.\ skydome, clouds, etc.).
Conventional global losses (e.g.\ an $L_1$-loss) are insensitive to the solar exposure range of due to the class-imbalance it entail.
Although greater sensitivity is a potential solution (e.g.\ $L_2$ instead of $L_1$) the potential loss-imbalances and/or gradient explosion must be mitigated.
In this regard, we propose implement class-segmented losses, as \cref{eq:selective}:

\begin{equation}
\mathcal{L}(I_{real},I_{fake}, M) = \mathcal{L}(I_{real}\subset M,I_{fake} \subset M)
\label{eq:selective}
\end{equation}
This allows conventional metrics (e.g.\ L$_1$) to be selectively applied via a segmentation mask ($M$) to target regions which would otherwise be saturated by pixels from outside their distributions (e.g.\ border, skydome).

\subsubsection{Unsupervised Losses}
\label{sec:methodology::losses:::unsupervised}

From an extraterrestrial point-of-view (POV), the sun is a $0.5^{\circ}$ angular-diameter disk with near-constant illumination \cite{seeds_astronomy} but, from a terrestrial POV, its size and radiant intensity are attenuated by a stochastic atmosphere.
As illustrated in \cref{fig:plt_illumination}, there is no intuitive relationship between solar-elevation, cloud cover, dynamic range and/or illumination.
Similarly, cloud formations consist of 27 categories \cite{NOAA_clouds} with variable textures and altitude-specific classification.
In the absence of labelled datasets and cloud-formation classification tools, current DNN sky-models unconditionally generate cloud formations and exhibit little diversity in cloud formations. Unintentionally, where segmentation morphology is deterministic of cloud formation, some models achieve a greater diversity by overfitting to input masks.

In the absence of input modalities to condition solar radiant intensity and/or cloud textures, skies are nondeterministic systems with intractable variability.
As a result, supervised losses (e.g. $L_1$, LPIPS \cite{LPIPS}) guide per an invalid assumption of exactness in reconstruction.
To mitigate, we propose a conditional n-Layer discriminator inspired by \cite{Pix2Pix_2016,stablevideodiffusionscaling} with adaptive gradient penalty \cite{arjovsky2017wassersteingan, gulrajani2017improvedtrainingwassersteingans,kodali2017convergencestabilitygans}, adaptive weighting of the adversarial loss \cite{rombach2021highresolution} and a modified hinge loss.
Given DNN sky-models are ignorant of their operation within LDR-space and the non-linear relationship between LDR- and HDR-space, unsupervised learning also mitigates ill-adapted conditioning by visual losses created for alternate LDR-spaces (e.g. LPIPS \cite{LPIPS} adaptability to logarithmic LDR-space has not been previously studied).

\section{Experimental Configuration}
\label{sec:experiments}

For the purpose of comparison, we train all AllSky, CloudNet, SkyGAN and SkyNet against the Laval HDR Sky database (HDRDB \cite{LavalHDRdb}, \cref{app:sec::dataset}), which consists of approximately 32k FDR HDRI.
We evaluate Text2Light on the FDR Laval Outdoor Dataset (LOD) \cite{YANNICK_2019_SKYNET} using the author's textual CLIP prompts.
Images herein are gamma ($\gamma=2.2$) tone-mapped and target the four primary skydome configurations: clear, cloudy with unobstructed sun, cloudy with obstructed sun, and overcast skies.

\subsection{Image-Processing}
We develop a pre-processing pipeline to re-size and augment HDRI, optimizing for retention of physically captured characteristics.
For evaluation, we quantify pixel intensity as linear luminance per BT.709 \cite{BT709} and further quantify global-illumination as Integrated Illumination ($\oiint_I$):

\begin{equation}
    \oiint_I(I) = \sum{\Omega|I|}
    \label{eq:IntegratedIllumination}
\end{equation}
Where $\Omega$ is the environment-map's solid angles.
As HDRDB, LOD and other datasets previously used to train DNN sky-models are not radiometrically calibrated, we assume a linear camera response and do not report illumination in metric quantities.

\subsubsection{Scaling environment-maps}
\begin{figure}[ht]
    \centering
    \includegraphics[width=\linewidth,keepaspectratio]{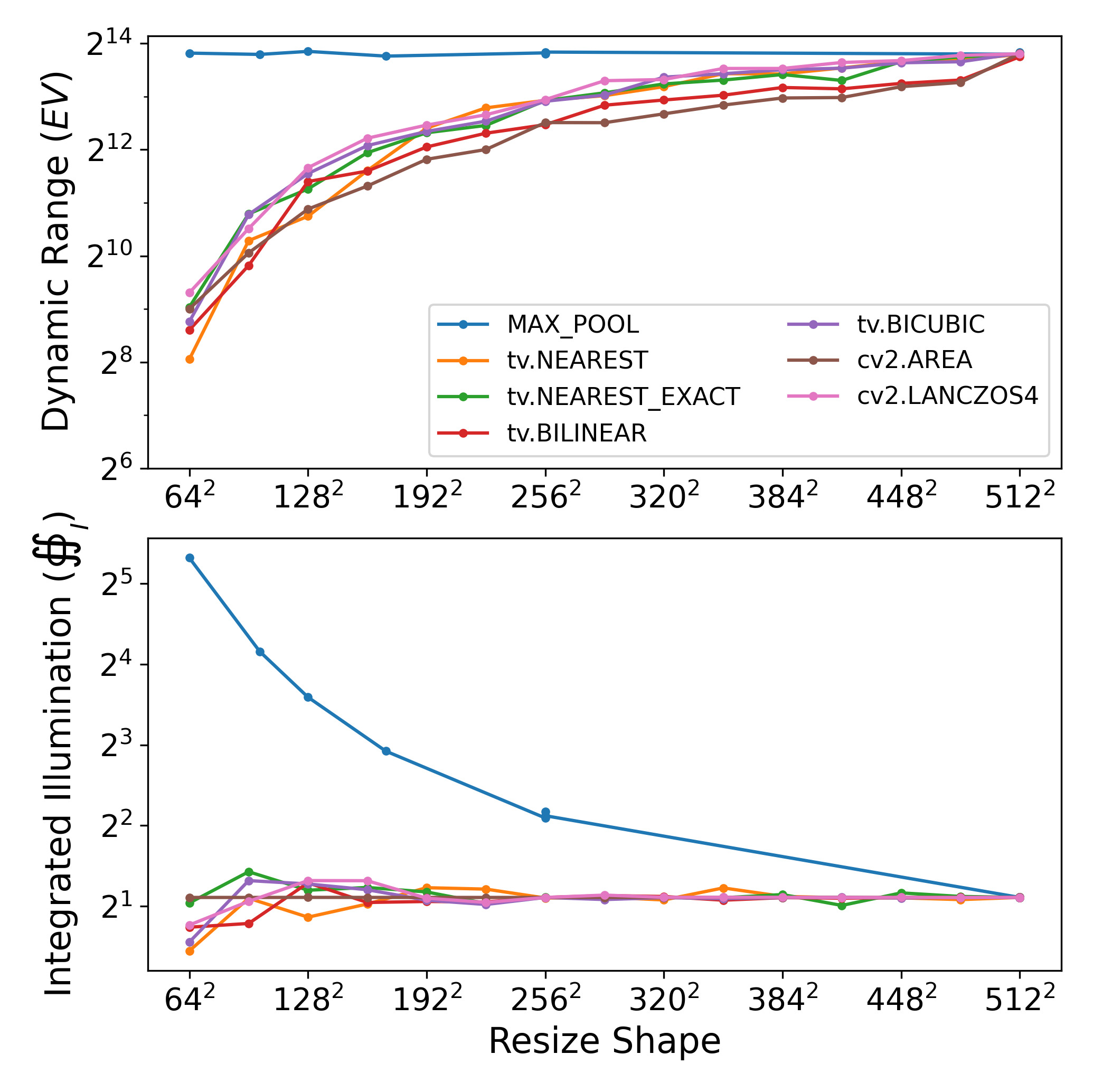}
    \caption{Impact of resizing a $512^2$ physically captured HDRI on exposure range and illumination. Mean of 100 augmentations of an HDRI from June 7th, 2016 at 12:52PM \cite{LavalHDRdb}.}
    \label{fig:plt_scale}
\end{figure}

We observe DNN sky-models are inherently scale-sensitive due to the impact of interpolation on the exposure range of physically captured HDRI and study this observation in \cref{app:sec::envmaps}.
As plotted in \cref{fig:plt_scale}, interpolation methods available through PyTorch (tv.) and OpenCV (cv2.) impact environment-map illumination and exposure range \cite{PyTorch,opencv}.
\footnote{See \cref{app:sec::envmaps:::scaling_env_maps} for the impact to shadows, tones, and light transmission in renderings.}

Environment-map format conversions are avoided where possible and, along with augmentation, are applied prior to downsampling.
Where otherwise unavoidable, linear spline interpolation is applied and downsampling favours factors of $f(x)=1/2^x$ to enable inter-area interpolation (retaining over 99\% of illumination energy).
To prevent discontinuities, environment-maps are skyangular formatted (See \cref{app:sec::text2light:::seams} for more detail on seams).

\subsubsection{Segmentation}

\begin{figure*}[ht]
  \centering
    \begin{subfigure}[b]{0.18\textwidth}
     \centering
     \includegraphics[width=\textwidth]{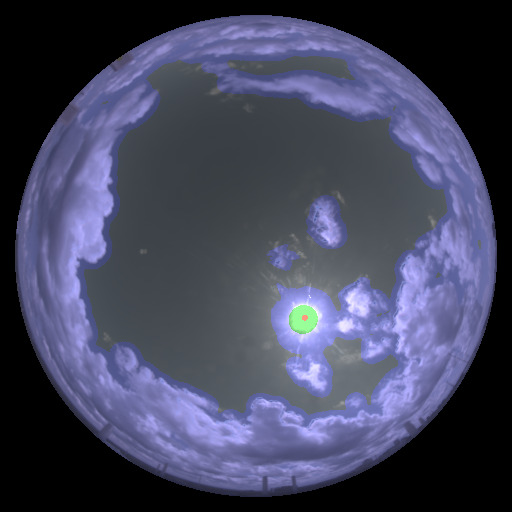}
     \caption{Overlay}
     \label{fig:maskDemo_overlay}
    \end{subfigure}
    \hfill
    \begin{subfigure}[b]{0.18\textwidth}
     \centering
     \includegraphics[width=\textwidth]{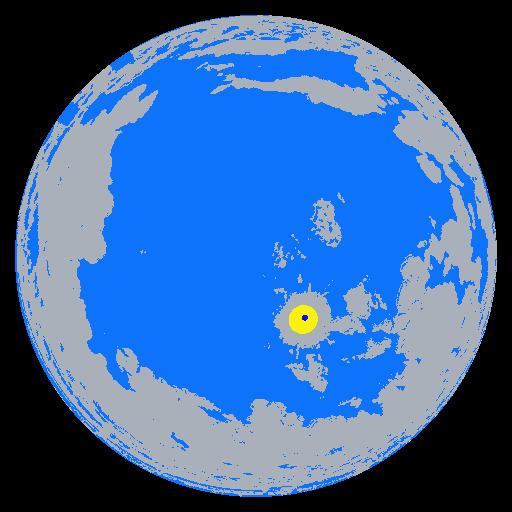}
     \caption{$\mathbb{Z}^{+}$ Crude}
     \label{fig:maskDemo_disc_crude}
    \end{subfigure}
    \hfill
    \begin{subfigure}[b]{0.18\textwidth}
     \centering
     \includegraphics[width=\textwidth]{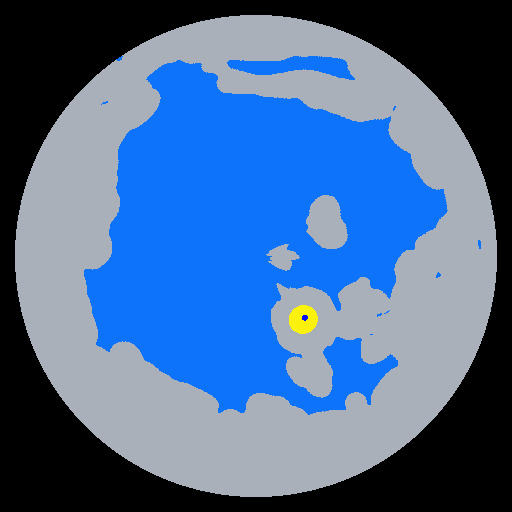}
     \caption{$\mathbb{Z}^{+}$ Hand-drawn}
     \label{fig:maskDemo_disc_handdrawn}
    \end{subfigure}
    \hfill
    \begin{subfigure}[b]{0.18\textwidth}
     \centering
     \includegraphics[width=\textwidth]{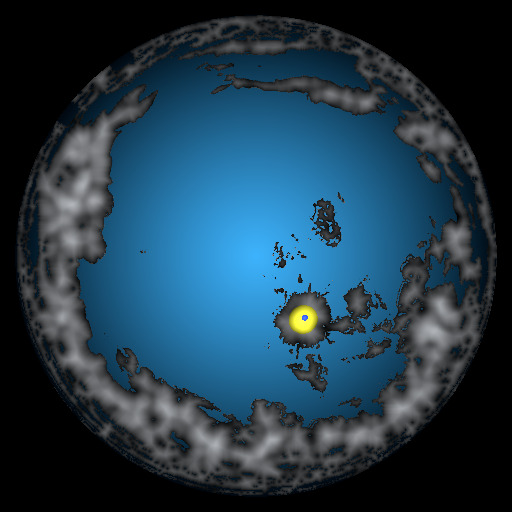}
     \caption{$\mathbb{R}$ Crude}
     \label{fig:maskDemo_cont_crude}
    \end{subfigure}
    \hfill
    \begin{subfigure}[b]{0.18\textwidth}
     \centering
     \includegraphics[width=\textwidth]{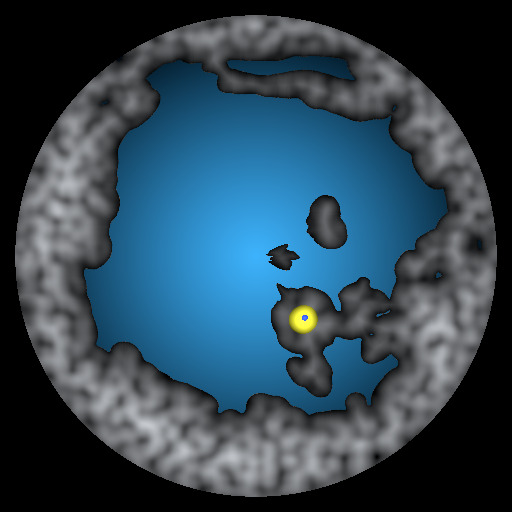}
     \caption{$\mathbb{R}$ Handdrawn}
     \label{fig:maskDemo_cont_handdrawn}
    \end{subfigure}
    \caption{Discrete ($\mathbb{Z}^{+}$, \cref{fig:maskDemo_disc_crude,fig:maskDemo_disc_handdrawn}) labels are 1-channel composites of solar, cloud, skydome (clear-sky), and border masks.
    Continuous ($\mathbb{R}$, \cref{fig:maskDemo_cont_crude,fig:maskDemo_cont_handdrawn}) labels augment the segmentation with distance-field modulation of perlin noise.
    Cloud segmentation is produced via colour ratio \cite{clouds_segmentation} and eroded from `crude' to `hand-drawn'.}
\label{fig:grid_labels}
\vspace{-0.5cm}
\end{figure*}

Solar positioning is refined from ephemeris calculations \cite{pySolar}, labelling the extraterrestrial $0.5^{\circ}$ sun as the solar disk.
To reflect terrestrial imagery, we extend the masked solar region to a diameter of $5^{\circ}$ to include the solar corona and atmospheric attenuation of the extraterrestrial solar disc.

Segmentation of cloud formations was achieved by thresholding \rgb via the ratio  $Y = \frac{\textcolor{blue}{B}-\textcolor{red}{R}}{\textcolor{blue}{B}+\textcolor{red}{R}}$ proposed by Dev et al.~\cite{clouds_segmentation}.
We find that applying this ratio to $\mu$-lawLog$_2$ ($T_{\mu \log_2}$) tone-mapped HDRI provides visually robust segmentation, but note variability with illumination intensity (e.g.\ sunrise/sunset) and seasonality \cite{koehler1991status}.
Cloud masks are morphologically processed to produce `crude' masks which can be further eroded by a parametric circular brush of kernel size $k$ to reduce complexity.
We characterize erosion by $k$=15 as emulating a `hand-drawn' sketch.

The final `discrete'  ($\mathbb{Z}^{+}$) label is a 1-channel composite of solar, cloud, skydome (clear-sky), and border masks.
We take inspiration from CloudNet \cite{DEEPCLOUDS_22} and create augmented `continuous' ($\mathbb{R}$) labels, infilling the solar regions as cosine similarity to the sun, the skydome as solid angles ($\Omega$), and clouds as distance fields modulated by Perlin noise.
A colourized illustration of `discrete' and `continuous' labels is provided in \cref{fig:grid_labels}.

\subsection{Baseline Models}
In this work we compare CloudNet, SkyGAN, Text2Light, and AllSky through training against HDRDB \cite{LavalHDRdb}.
We reproduce CloudNet and SkyGAN per the authors' source code with minimal modifications to the data-processing pipeline.
We evaluate Text2Light's \textit{outdoor} functionality  the author's published model and the LOD dataset \cite{YANNICK_2019_SKYNET}.

\subsubsection{CloudNet}
CloudNet was reproduced per the authors' source code with model and parameters unchanged \cite{DEEPCLOUDS_22}.
We adapted the data-processing pipeline to our segmentation of HDRDB, reproducing the input histogram-equalized Ho\v{s}ek-Wilkie \cite{HOSEK_13} clear skies, cloud distance field maps, and inverted tone-mapping ($T_{I^{-1}}$) per \cref{eq:CloudNet_compression}.
Originally trained on a small dataset of 629 HDRI for 7,500 epochs, we implemented a framework to facilitate a 3,500 epoch training.

\begin{equation}
    \text{Inverted: T}_{I^{-1}}\left(I\right)
    = 1 / \left(1 + I + 0.01\right)
    \label{eq:CloudNet_compression}
\end{equation}

We evaluated CloudNet both with and without the sun-pass-through mechanism, thus repeating the model's evaluation with the Ho\v{s}ek-Wilkie clear sky solar disk summed into generated environment-maps.
To evaluate the impact of CloudNet's input modalities, we train a modified baseline per our discrete ($\mathbb{Z}^+$) input labels.

\subsubsection{SkyGAN}
SkyGAN was reproduced per the authors' source code with model and parameters unchanged \cite{SKYGAN_2022}.
We adapt the data-processing pipeline to HDRDB, reproducing the input PSM \cite{PSM_21} clear skies, and natural logarithmic (${T}_{\log_e}$) tone-mapping per \cref{eq:SkyGAN_compression} with $\alpha=0.22$, $\beta=2.5$ and $\epsilon=1^{-3}$).

\begin{equation}
    \text{Natural Logarithmic: T}_{\log_e}\left(I\right) = \left[ log(I + \epsilon) + \beta \right] * \alpha
    \label{eq:SkyGAN_compression}
\end{equation}

We completed a 3,500 epoch training of SkyGAN and a variant which bypassed the clear sky encoder.
In addition, we adapt StyleGAN3 \cite{styleGAN3_2021} (SkyGAN's backbone) to HDRDB and ablate select tone-mapping operators through 1,000 epoch trainings.

\subsubsection{Text2Light}
Evaluation of Text2Light's \textit{outdoor} functionality was completed via unaltered implementation the author's published model, checkpoints and textual CLIP prompts with no upscaling ($\beta{=}1$) \cite{text2light}.
Text2Light's generation pipeline consists of three stages: 1) LDR generation via diffusion, 2) LDR to HDR augmentation via MLP Inverse Tone-Mapping Operator (iTMO), and 3) HDR to FDR parametric boosting per \cref{eq:text2light_parametric_boost} with constants $\rho$=4,  $\vartheta$=0.83,  $\gamma$=1 and $\beta$= 0.7.\footnote{$I_{b}$ denotes a batch of images, $\overline{\overline{I_{b}}}$ denotes the per-image-mean of $I_{b}$, $\overline{I_b'}$ is the mean of $I_b'$ and $\left[X\right]_a^b$ clips the values of $X$ such that $a \leq x \leq b$.}

\vspace{-\baselineskip}
\begin{equation}
\begin{split}
    M = \left[ \overline{\overline{I_{b}}} / \max\left( \overline{\overline{I_{b}}} \right) - \vartheta \right]_0^1  \\
    I_b' = I_b + \left( I_b M \rho \right) \\
    I_b' = e^{(I_b' - \overline{I_b'})\gamma - \beta}
\end{split}
\label{eq:text2light_parametric_boost}
\end{equation}

We evaluated each stage of Text2Light's generation pipeline, normalizing the resulting HDR image pairs for exposure via $|I_{real}| = \alpha|I_{fake}|$ and reformatting the environment-maps to skyangular for visual metrics.

\subsubsection{AllSky}
In \cref{app:sec::AllSky}, we leverage AllSky's configurability to ablate the input modalities, tone-mapping operators, and the losses of SkyNet, CloudNet, and SkyGAN.
As a baseline, we recreate the training of \textit{SkyNet} \cite{YANNICK_2019_SKYNET} which implemented a U-Net architecture guided only by a global $L_1$-loss.
We create a variants emulating CloudNet (continuous labels ($\mathbb{R}$) and LPIPS loss), and SkyGAN (implementing our n-Layer discriminator).
All variants implemented class-selective $L_1$-losses for border and skydome regions.

Per the findings of our ablations in \cref{app:sec::AllSky}, we complete 400-epoch trainings of AllSky at $256^2$, and $512^2$ resolution and report their performance in \cref{tab:quantitative_results}.
These models are trained with $\mathbb{Z}^{+}$ discrete labels per guidance from LPIPS, our n-Layer discriminator, and class-selective $L_1$-losses for border and skydome regions.

\subsection{Metrics}

An objective of this work is to demonstrate the inadequacy of current DNN sky-model evaluations in regards to solar illumination and characterize the trade-off between accurate illumination and visual quality exhibited by current models.
In this regard, we evaluate the sensitivity of commonly reported metrics in \cref{app:sec::metrics}, finding that SSIM, Multi-Scale SSIM, and HDR-VDP3 offer no meaningful sensitivity to exposure range, while other supervised losses such as $L_1$, $L_2$, and EMD offer limited and diminishing sensitivity with increasing resolution.
We also note an assumption of reconstruction exactness introduces significant noise from penalizes valid natural stochasticity in style and positioning.

We evaluate the performance of DNN sky-models for visual quality (LPIPS \cite{LPIPS} and FID \cite{FID}), exposure range (EV), and illumination ($\oiint_I$).
Given the generalization that atmospheric textures are low dynamic range, we assume visual quality metrics are insensitive to tone-mapping and, for consistency, evaluate all visual metrics in Gamma ($\gamma$) cLDR-space.
We quantify exposure range and illumination as EV and $\oiint_I$ for environment-maps in uncompressed HDR space.
To demonstrate the importance and sensitivity of these metrics, we report them alongside $L_1$, $L_2$, and PSNR as many works evaluate using these metrics exclusively \cite{YANNICK_2019_SKYNET,DEEPCLOUDS_22,text2light}.
For greater sensitivity, we implement PSNR with base $\log_2$ (PSNR$_{\log_2}$).

\section{Experimental Results}

In this work we adapt CloudNet, and SkyGAN to HDRDB without modification to authors' defined pre- and post-processing pipelines.
As a result, each baseline models exhibit different ground truth exposure ranges (EV) and illumination ($\oiint_I$) despite being tested against an identical subset of samples.

\begin{table*}[htb]
    \centering
  \caption{
  Comparative results for Text2Light \cite{text2light}, CloudNet \cite{DEEPCLOUDS_22}, SkyGAN \cite{SKYGAN_2022}, and AllSky (ours).
  AllSky offers unsurpassed visual quality (LPIPS, FID) for both crude (C) and hand-drawn (H) labels, and accurate FDR illumination (EV, $\oiint_I$) superseded only by SkyGAN.
  The results further demonstrate that $L_1$, $L_2$ and PNSR offer little correlation to visual quality or illumination.
  \underline{Underlined} values indicate category best. Exposure matching is denoted by (*).
  }
  \label{tab:quantitative_results}
  \begin{tabular}{@{}r|c|cc|cccc@{}}
    & \multicolumn{1}{c|} {\textbf{LDR}}
    & \multicolumn{2}{c|}{\textbf{$T_{\gamma}$-cLDR}}
    & \multicolumn{4}{c}{\textbf{HDR}} \\
    \cline{2-8}
    & PSNR$_{\log_2}$ $\uparrow$ & LPIPS $\downarrow$ & FID $\downarrow$ & $L_1$ $\downarrow$ & $L_2$ $\downarrow$ & EV $\leftarrow$ & $\oiint_I \leftarrow$ \\
    \toprule
      Ground Truth $512^2$     & -    & -  & -  & -    & -   & 12.85 & 3.54 \\
      \hdashline
    Text2Light SRiTMO Boosted  & 46.5 & 0.51 & - & 0.27 & \underline{816}  & 8.4 & 0.16 \\
    Text2Light SRiTMO Boosted* & 46.5 & \underline{0.42} & - & 0.32 & 1380 & \underline{12.92} & 3.73 \\
    Text2Light LDR+Boost*          & \underline{47.3} & \underline{0.42} & - &  \underline{0.25} & 827 & 7.96  & \underline{3.69} \\
    \bottomrule

     Ground Truth $512^2$      &-&-&-&-&-& 12.7 & 1.05 \\
     \hdashline
     CloudNet                    & \underline{99.4} & \underline{0.17} & 35.2   & \underline{0.07} & \underline{414} & 6.3  & 0.53 \\
     CloudNet w/ Sun              & \underline{99.4} & \underline{0.17} & \underline{35.1}  & 0.17 & 1907 & \underline{14.4} & 2.07 \\
     CloudNet w/o Clear Sky       & 92.1 & 0.18 & 41.5  & 0.08 & 418 & 6.7  & \underline{0.64} \\
     CloudNet w/o Clear Sky w/Sun  & 92.1 & 0.18 & 41.5  & 0.18 & 1909 & \underline{14.4} & 2.18 \\
    \bottomrule
    Ground Truth $256^2$      &-&-&-&-&-& 11.12 & 1.15 \\
    \hdashline
    SkyGAN                   &-&-& \underline{55} &-&-& 10.7 & 0.83 \\
    \begin{tabular}{rr}
    SkyGAN & w/o Clear Sky  
    \end{tabular}
    &-&-&70&-&-& \underline{11.3} & \underline{1.01} \\
    \hdashline
    \begin{tabular}{rr}
    StyleGAN & $T_{\mu\log_2}$  
    \end{tabular}
    &-&-&\underline{49}&-&-& \underline{7.7}  & \underline{1.04} \\
    \bottomrule

    Ground Truth $256^2$   &-&-&-&-&-& 8.09 & 1.13 \\
    \hdashline
    \begin{tabular}{rrrr}
    AllSky & $\mathbb{Z}^{+}$ & $T_{\mu\log_2}$ & C 
    \end{tabular}
     & \underline{104.3} & \underline{0.13} & \underline{14.9} & \underline{0.07} & \underline{965} & \underline{7.24} & \underline{0.85} \\
    \begin{tabular}{rrrr}
    AllSky & $\mathbb{Z}^{+}$ & $T_{\mu\log_2}$ & H 
    \end{tabular}
    & 98.1 & 0.16 & 22.6 & 0.43 & 344704 & 9.58 & 4.49 \\
    \bottomrule

    Ground Truth $512^2$   &-&-&-&-&-& 9.11 & 1.14 \\
    \hdashline
    \begin{tabular}{rrrr}
    AllSky & $\mathbb{Z}^{+}$ & $T_{\mu\log_2}$ & C 
    \end{tabular}
    & \underline{102.8} & \underline{0.14} & \underline{17.1} & \underline{0.06} & \underline{467} & \underline{8.23} & \underline{0.82} \\
    \begin{tabular}{rrrr}
    AllSky & $\mathbb{Z}^{+}$ & $T_{\mu\log_2}$ & H 
    \end{tabular}
    & 93.3 & 0.18 & 24.8 & 0.11 & 9919 & 9.95 & 1.52 \\
    \bottomrule
\end{tabular}
\vspace{-6pt}
\end{table*}

\subsection{CloudNet}
\textit{CloudNet} and \textit{CloudNet w/o Clear Sky} differ little in \cref{tab:quantitative_results}, suggesting clear sky and cloud distance field map input modalities contribute negligibly to model performance.
Though the Ho\v{s}ek-Wilkie model provides intuitive control over clear skies generation through parameters for turbidity and ground albedo, CloudNet keeps these parameters constant and after histogram equalization, retains no relationship to these parameters.
The aggregation of the clear-sky sun (see \textit{w/ Sun}) significantly improves EV, but overshoots illumination and introduces a solar disc that pierces through cloud formations.
This differs from natural terrestrial imagery where the solar disk is attenuated by the atmosphere and diffused through cloud formations.
Given the disparity between generated and FDR ground truth illumination (EV, $\oiint$), environment maps are characterizable as EDR.

\subsection{SkyGAN}
SkyGAN's exposure range and illumination are shown to be adversely impacted by its clear sky input modality, as its removal improves both exposure range and illumination.
SkyGAN exhibits EDR exposure range (75\% of ground truth in linear space) but offers poor visual quality and illumination in comparison to StyleGAN3.

Visually in \cref{app:sec::SkyGAN}, SkyGAN and StyleGAN both exhibit texture-less cloud `smears'.
We believe this to be the result of discriminator emphasis on solar features.
In this regard, StyleGAN3 baselines with less aggressive tone-mapping operator ($T_\gamma$, $T_{\log_2}$) collapsed and we exclusively report StyleGAN3 with $T_{\mu\log_2}$.

\subsection{Text2Light}

As shown by \textit{Text2Light SRiTMO Boosted*} in \cref{tab:quantitative_results}, Text2Light generates EDR environment-maps which can be boosted and exposure matched to provide an exposure range and illumination close to the FDR ground truth.
As demonstrated through \textit{Text2Light SRiTMO Boosted}, exposure matching to the ground truth is imperative to correct under-exposure.
We note that our exposure matching is dependent on reference FDR imagery and not manually reproducible through conventional exposure and/or offset sliders.

We ablate Text2Light's iTMO by directly augmenting generated LDR imagery to FDR via the parametric boost \cref{eq:text2light_parametric_boost}.
As shown by \textit{Text2Light LDR+Boost*}, this sample-specific manually tuned heuristic boosts LDR to EDR with lower exposure range, but matches or improves visual quality.
We find the results to be inconsistent and scene dependent, producing incoherent results when imagery lacks texture below the horizon.
Though inconsistency and manual parameter tuning may be accommodatable in applications such as IBL rendering, it is prohibitive of many downstream scientific and engineering applications which require real-world accuracy.

In \cref{app:sec::text2light:::matches} we demonstrate user control via CLIP prompts to be erratic and with little stochasticity.
One run of the authors' textual prompts generated 80 image pairs consisting of 17 unique images and the remaining reappearing between 2 and 12 times within the generated set.

\subsection{AllSky}

Through our extended work in \cref{app:sec::AllSky}, we ablate AllSky for discrete ($\mathbb{Z}^{+}$) and continuous ($\mathbb{R}$) input modalities, tone-mapping operators ($T_m$), and losses.
Continuous ($\mathbb{R}$) labels quantitatively provide an advantage with adversarial losses, but visual examination shows perlin noise propagating into the textures of cloud formations. 
As this is visually unappealing, we find discrete ($\mathbb{Z}^{+}$) labels to be favourable.

We perform an ablation study to determine the tractability provided by cloud masks by iteratively eroding the morphology of cloud segmentation.
Though illumination (EV, $\oiint_I$) is sustained, visual performance (LPIPS, FID) is incrementally lost as cloud masks are eroded from crude (C) to hand-drawn (H).
We repeat this study to determine performance when used by a human-operator, providing models with hand-drawn (H) cloud masks regardless of cloud-mask morphology given during training. 
This resulted in lost visual performance (LPIPS, FID) proportional to the difference in erosion between training and hand-drawn (H) cloud masks, demonstrating models trained for favourable visual performance with crude (C) cloud masks do not sustain the same performance when used by a human operator.

In \cref{app:sec::AllSky:::ablation_tone-mappers}, we find tone-mapping directly impacts the visual quality, exposure range and illumination of sky-model, with our $T_{\mu\log_2}$ tone-mapper improving visual quality and illumination but note aggressive tonemapping introduces instability in exposure range due to the non-linearity between error in LDR and HDR space.
Though unideal, aggressive tonemapping mitigates the `holes' in place of the solar disk produces by weaker tonemappers ($T_{I^-1}$, and $T_{\gamma}$).

Through our ablation of $L_1$, LPIPS, and adversarial losses, we find alleviating the comparative improves visual quality and illumination, but adversarial losses disproportionately differentiating through solar features and impede visual quality.

Combining these findings, we created AllSky as reported in \cref{tab:quantitative_results}.
We apply our discrete ($\mathbb{Z}^{+}$) labels, our $T_{\mu\log_2}$ tone-mapping, our class-selective $L_1$ for border and skydome regions, LPIPS, and our n-Layer discriminator.
As shown, AllSky demonstrates visual quality (LPIPS, FID) for both crude (C) and hand-drawn (H) labels and accurate illumination (EV, $\oiint_I$).
AllSky sustains visual quality with FDR illumination accuracy is exceeded only by our variant of SkyGAN and without manually-tuned boosting as required by Text2Light.

\section{Discussion}
\label{sec:discusion}

The work herein studies state-of-the-art (SOTA) sky-models and presents ablations of current methodologies with the intention of deepening our understanding of the shortfalls of current DNN sky-models.
Sky-modelling extends beyond the conventional paradigm of High Dynamic Range literature through a requirement for greater exposure range and we demonstrate in \cref{tab:quantitative_results} that current methods exhibit a trade-off between photorealism (visual quality) and real-world Full Dynamic Range (FDR).
We demonstrate this to directly impact Image Based Lighting (IBL) where renderings with saturated illumination produce skewed tones, shadows, and light transmission.

Though our proposed model AllSky out-performs current SOTA for all visual benchmarks, we note that higher-resolutions will exacerbate solar class-imbalance and exposure range, leading to lost visual quality and greater instability in exposure range and illumination.
This can be observed through AllSky in \cref{tab:quantitative_results} as resolution is doubled from $256^2$ to $512^2$.

We demonstrate that careful consideration should be taken in evaluating the role and functionality of sky-model input modalities.
We demonstrate cloud masks can provide tractability to deterministically reconstruct cloud formations and, if not manually replicatable by a user, result in lost visual quality.
We further demonstrate that augmenting inputs with clear-sky and/or other priori does not add user-configurability or improve performance.
Though textual prompts can facilitate human interaction in many applications, we show Text2Light's implementation to be dubious and unsupporting of diversity.

Through our study of metrics, we find that commonly reported metrics provide little insight into the exposure range and illumination of generated environment-maps.
Given the stochasticity of clouds formations and variability of solar intensity, we find that comparative metrics penalize natural variation which is unimpeding of photorealism and illumination.
As shown in \cref{tab:quantitative_results}, our proposed metrics for exposure range (EV) and illumination ($\oiint_I$) offer sensitivity which correlate to better visual quality in downstream applications such as IBL rendering.

\section{Conclusions}
\label{sec:conclusions}

In this work we propose a sky-model trained directly on physically captured FDR imagery (AllSky) and leverage its configurability to study the limitations and evaluation of DNN sky-models.
Through our framework, we demonstrate DNN sky-models are subject to class-imbalances and exposure range limitations which are uncharacterized by common metrics and can result in a trade-off between visual quality and illumination. 
Though AllSky surpasses the current SOTA in both visual quality and user-controllability, we demonstrate existing models cannot generate environment-maps with the same visual quality and illumination as physically captured HDRI or parametric sky-models. 
Significant work remains to enable DNN sky-models to simultaneously emulate photorealistic weathered skies and accurate natural illumination at higher resolutions.

{
    \small
    \bibliographystyle{ieeenat_fullname}
    \bibliography{main}
}

\clearpage
\appendix
    \begin{figure*}[tbp]
    \centering
    \includegraphics[width=\linewidth,height=0.98\textheight,keepaspectratio]{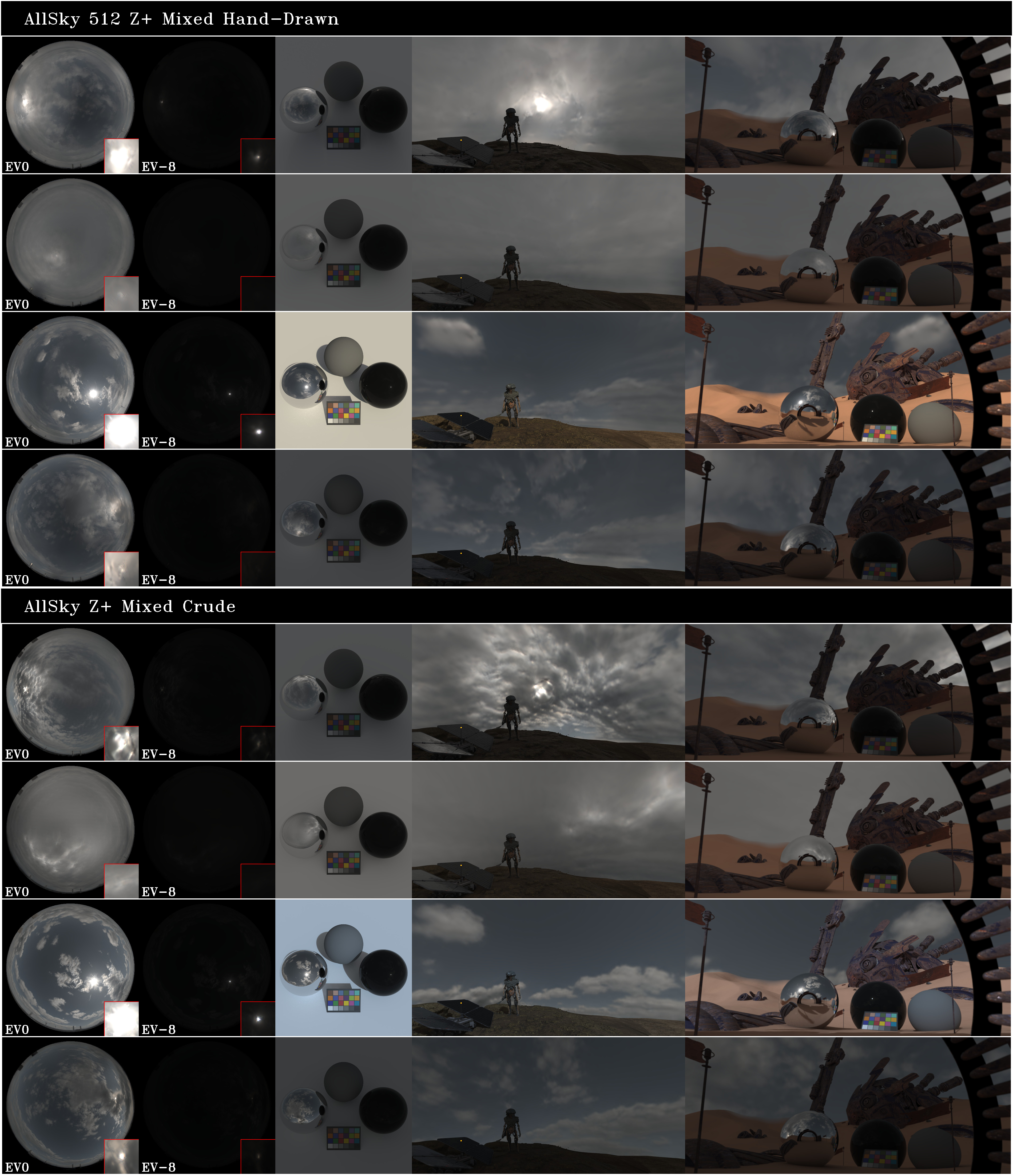}
    \caption{IBL renders with environment-maps from AllSky with $T_{\mu\log_2}$ tonemapping (Mixed) and discrete $\mathbb{Z}^+$ labels. Our `Hand-drawn' labels (top) generate environment maps with the same illumination but lesser photorealism than our `Crude' labels (bottom).}
    \label{fig:grid_AllSky_master}
\end{figure*}

\begin{figure*}[tbp]
    \centering
    \includegraphics[width=\linewidth,height=0.6\textheight,keepaspectratio]{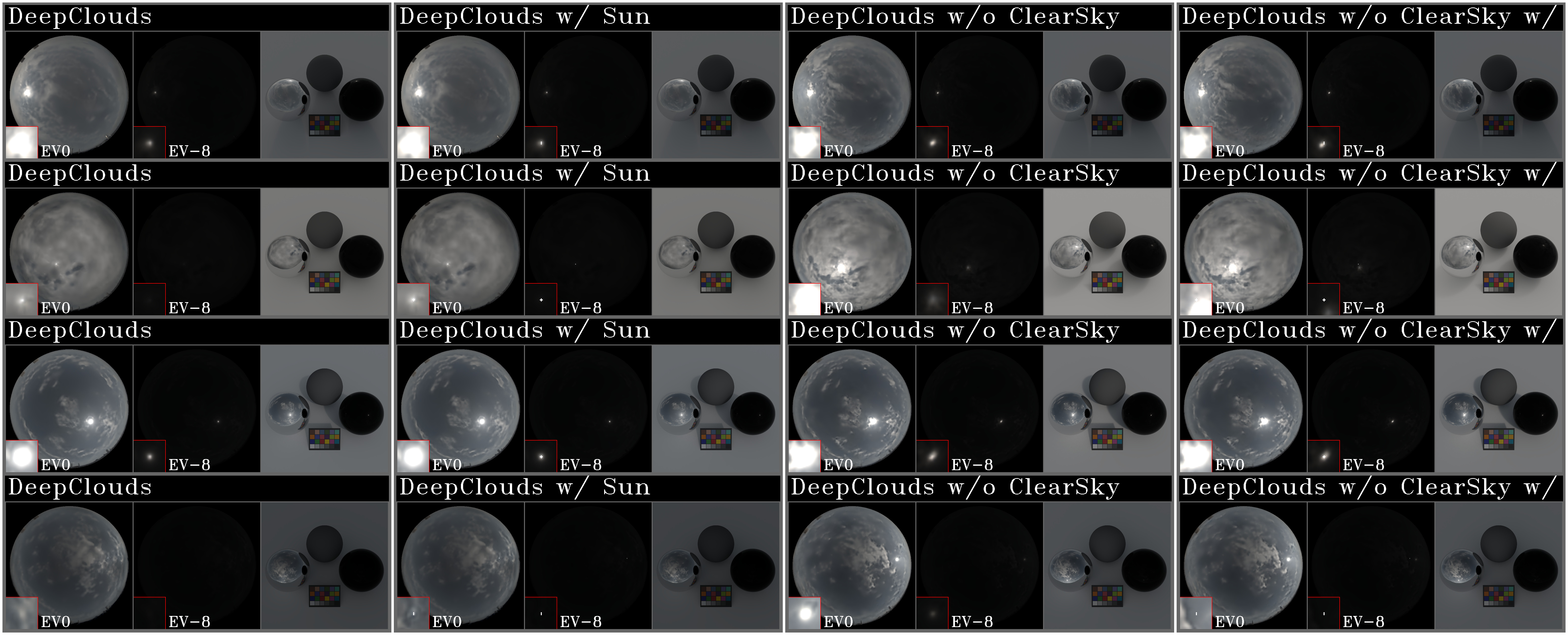}
    \caption{IBL renders with environment-maps from CloudNet and its variants. The pass-through HW sun (w/ Sun) significantly improves propagation of the solar radiant intensity to EV-8. Though FID performance is lost with the removal of the HW clear sky input (w/o ClearSky) in \cref{tab:quantitative_results}, visual inspection shows this lost performance to be dubious.}
    \label{fig:grid_CloudNet_master}
\end{figure*}

\begin{figure*}[tbp]
    \centering
    \includegraphics[width=\linewidth,height=0.6\textheight,keepaspectratio]{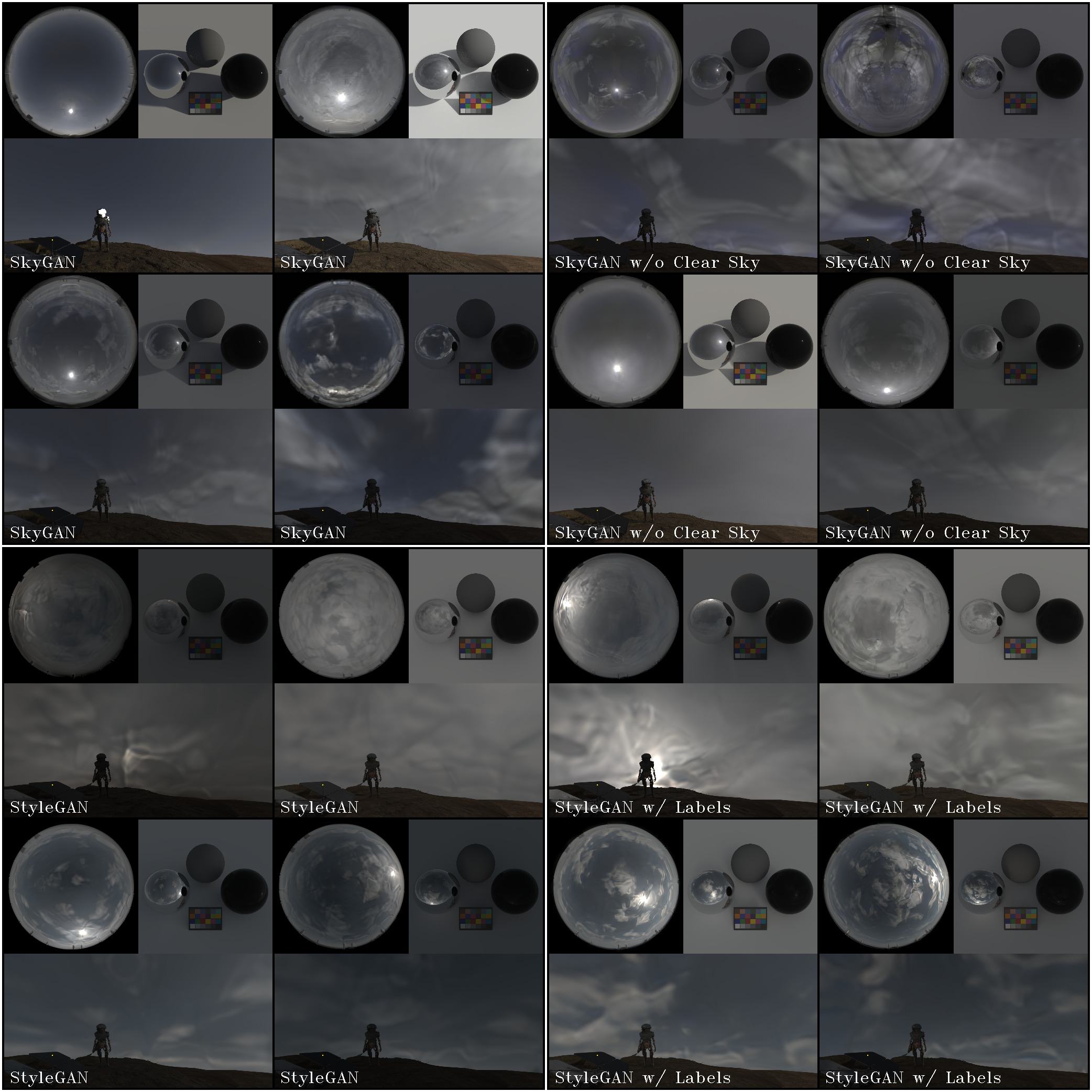}
    \caption{IBL renders with environment-maps from SkyGAN and StyleGAN. Though SkyGAN offers solar radiant intensity which casts strong shadows, visual quality is poor. Though visual quality is marginally improved by StyleGAN, both models are unable generate environments maps with both photorealism and accurate illumination.}
    \label{fig:grid_SkyGAN_master}
\end{figure*}
    \onecolumn
\section{Environment-maps}
\label{app:sec::envmaps}

\begin{figure*}    \includegraphics[width=\textwidth,height=.9\textheight,keepaspectratio]{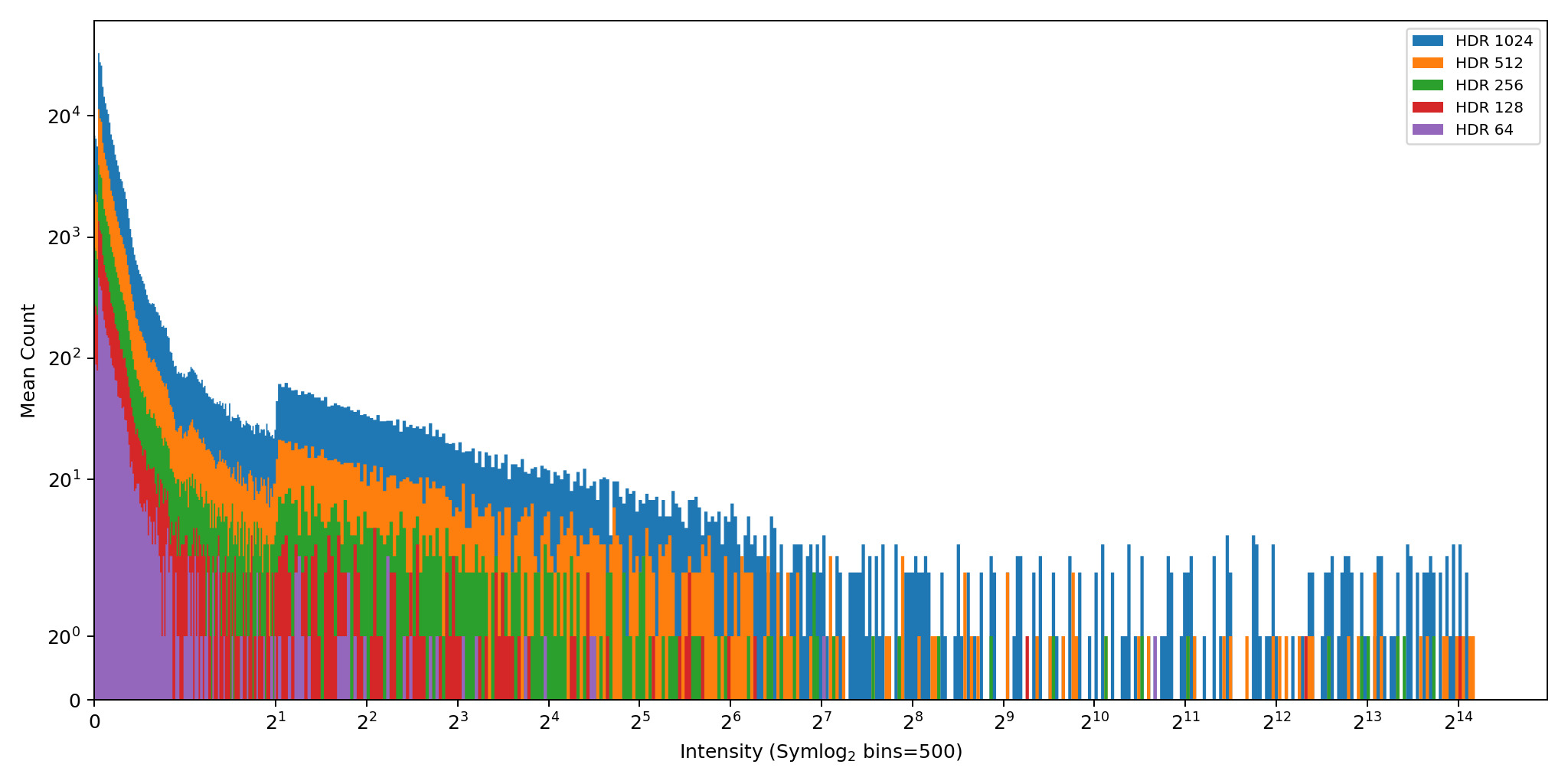}
    \caption{Histogram demonstrating the loss of intensity from inter-linear downsampling a latlong environment-map. HDRDB sample HDRI from June 7th, 2016 at 1:54PM \cite{LavalHDRdb}.}
    \label{app:fig::scaling_histogram}
\end{figure*}

\begin{table*}[htb]
\centering
\caption{
[Left] Impact of inter-linear downsampling an environment-map on exposure range (EV) and Illumination ($\oiint_I$) in latlong and sky-angular format.
[Right] Impact of inter-linear upsampling an environment-map back to native resolution (1024 latlong, $512^2$ sky-angular) on exposure range (EV) and Illumination ($\oiint_I$).
Upsampling does not recover lost exposure range or illumination.
HDRDB sample HDRI from June 7th, 2016 at 1:54PM.
}
\label{app:tab::resolution_DR_II}
\begin{tabular}{rr|cccc|lrr|cccc|}
\cline{3-6} \cline{10-13}
\multicolumn{1}{l}{} &
  \multicolumn{1}{l|}{} &
  \multicolumn{4}{c|}{Native} &
   &
  \multicolumn{1}{l}{} &
  \multicolumn{1}{l|}{} &
  \multicolumn{4}{c|}{Upsampled} \\ \cline{1-6} \cline{8-13}
\multicolumn{1}{|r|}{\multirow{3}{*}{Latlong}} &
  Shape &
  \multicolumn{1}{c|}{1024} &
  \multicolumn{1}{c|}{512} &
  \multicolumn{1}{c|}{256} &
  128 &
  \multicolumn{1}{l|}{} &
  \multicolumn{1}{r|}{\multirow{3}{*}{Latlong}} &
  Shape &
  \multicolumn{1}{c|}{\textbf{1024}} &
  \multicolumn{1}{c|}{512} &
  \multicolumn{1}{c|}{256} &
  128 \\ \cline{2-6} \cline{9-13}
\multicolumn{1}{|r|}{} &
  EV &
  \multicolumn{1}{c|}{14.2} &
  \multicolumn{1}{c|}{13.7} &
  \multicolumn{1}{c|}{14.1} &
  10.7 &
  \multicolumn{1}{l|}{} &
  \multicolumn{1}{r|}{} &
  EV &
  \multicolumn{1}{r|}{13.4} &
  \multicolumn{1}{c|}{13.6} &
  \multicolumn{1}{c|}{13.8} &
  10.5 \\ \cline{2-6} \cline{9-13}
\multicolumn{1}{|r|}{} &
  $\oiint_I$ &
  \multicolumn{1}{c|}{1.99} &
  \multicolumn{1}{c|}{1.99} &
  \multicolumn{1}{c|}{2.45} &
  1.34 &
  \multicolumn{1}{l|}{} &
  \multicolumn{1}{r|}{} &
  $\oiint_I$ &
  \multicolumn{1}{c|}{1.99} &
  \multicolumn{1}{c|}{1.99} &
  \multicolumn{1}{c|}{2.45} &
  1.34 \\ \cline{1-6} \cline{8-13}
\multicolumn{1}{|r|}{\multirow{3}{*}{Sky-Angular}} &
  Shape &
  \multicolumn{1}{c|}{$512^2$} &
  \multicolumn{1}{c|}{$256^2$} &
  \multicolumn{1}{c|}{$128^2$} &
  $64^2$ &
  \multicolumn{1}{l|}{} &
  \multicolumn{1}{r|}{\multirow{3}{*}{Sky-Angular}} &
  Shape &
  \multicolumn{1}{c|}{\textbf{$512^2$}} &
  \multicolumn{1}{c|}{$256^2$} &
  \multicolumn{1}{c|}{$128^2$} &
  $64^2$ \\ \cline{2-6} \cline{9-13}
\multicolumn{1}{|r|}{} &
  EV &
  \multicolumn{1}{c|}{13.5} &
  \multicolumn{1}{c|}{11.7} &
  \multicolumn{1}{c|}{9.7} &
  3.7 &
  \multicolumn{1}{l|}{} &
  \multicolumn{1}{r|}{} &
  EV &
  \multicolumn{1}{c|}{13.5} &
  \multicolumn{1}{c|}{11.1} &
  \multicolumn{1}{c|}{9.3} &
  3.6 \\ \cline{2-6} \cline{9-13}
\multicolumn{1}{|r|}{} &
  $\oiint_I$ &
  \multicolumn{1}{c|}{1.97} &
  \multicolumn{1}{c|}{1.48} &
  \multicolumn{1}{c|}{1.26} &
  0.85 &
  \multicolumn{1}{l|}{} &
  \multicolumn{1}{r|}{} &
  $\oiint_I$ &
  \multicolumn{1}{c|}{1.97} &
  \multicolumn{1}{c|}{1.48} &
  \multicolumn{1}{c|}{1.23} &
  0.82 \\ \cline{1-6} \cline{8-13}
\end{tabular}
\end{table*}

\subsection{Scaling}
\label{app:sec::envmaps:::scaling_env_maps}

\begin{figure*}
    \includegraphics[width=\textwidth,keepaspectratio]{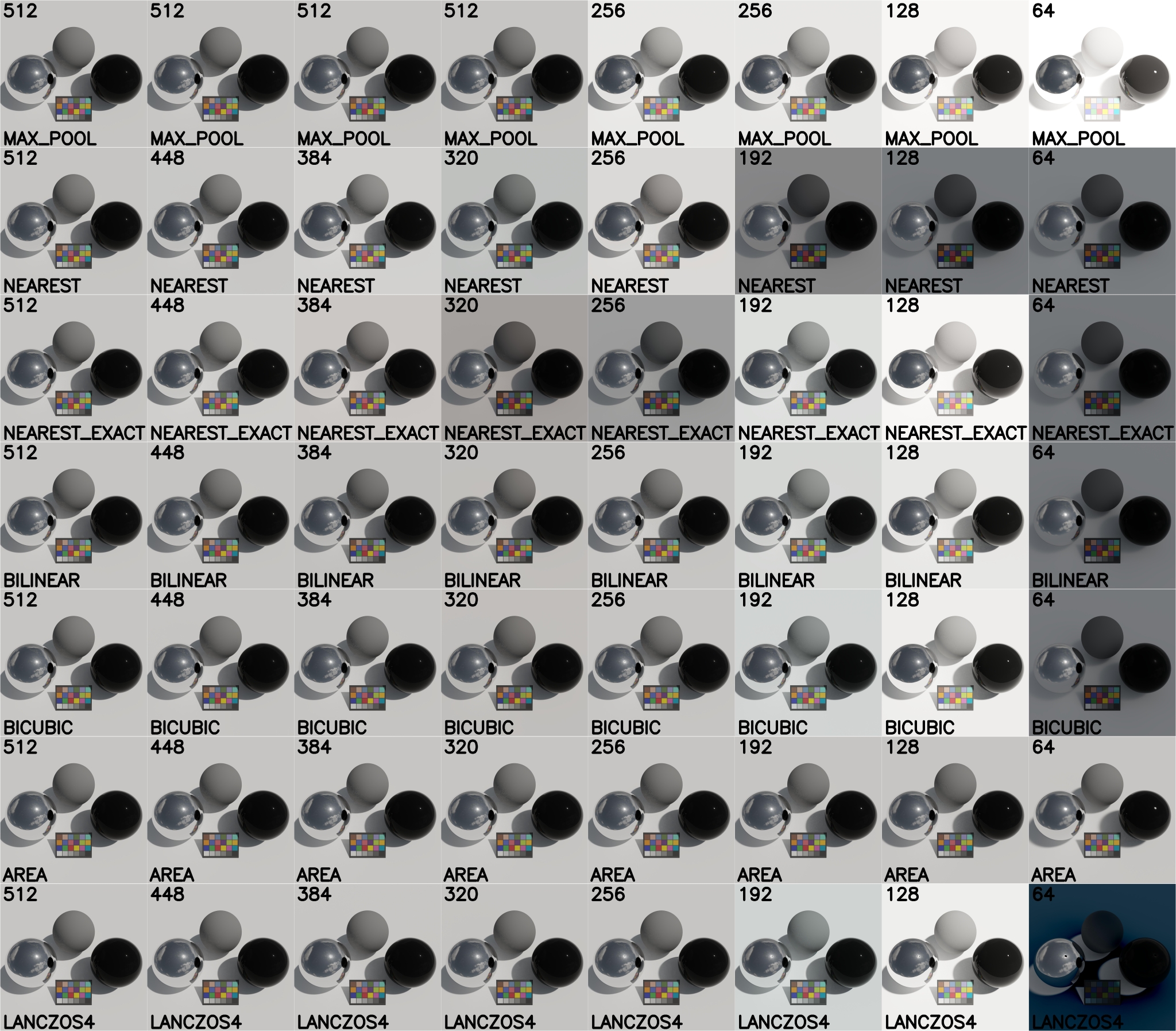}
    \caption{
    IBL renders from downsizing a $512^2$ environment map using various interpolation methods.
    Downsampling with Area-interpolation (AREA) visually best preserves shadows, tones, and light transmission.
    Max pool interpolation is implemented for factors $f(x)=1/2^x$ and grossly over-saturates rendered scenes.
    HDRDB sample from June 7th, 2016 at 12:52 PM \cite{LavalHDRdb}.
    }
    \label{app:fig::scaling_grid}
\end{figure*}

\begin{sloppypar}
We supplement the analytically demonstrated discrepancies with environment-map scaling reported in \cref{fig:plt_scale} with the renderings in \cref{app:fig::scaling_grid}.
As shown, downsampling with area-interpolation (cv2.AREA) quantitatively and visually best preserves ground-truth illumination and tones.
As reflected in the decay of the crisp edges of shadows and intensity of illumination transmitted to translucent surfaces, exposure range generally decays with downsampling.
This is further demonstrated in \cref{app:fig::scaling_histogram} where the histogram demonstrates the loss of exposure range from inter-linear downsampling a sky-angular environment-map.
Though max-pooling preserves exposure range (\cref{fig:plt_scale}; MAX\textunderscore POOL), it results in overexposure, poor tones, and washed-out shadows.
\end{sloppypar}

Care should be taken in selecting an interpolation method for downsampling and, where possible, the native resolution of the physical capture should be used.
It should be further noted that downsampling affects environment-map morphology where downsampling increases the angular diameter of the sun.
This phenomenon can be observed with bilinear-interpolation (BILINEAR) in \cref{app:fig::scaling_grid}, where the solar disk reflected off of the black glass orb steadily increased in diameter.
Given the role of the sun as an IBL scene's principal light source, this directly impacts shadows, tones, and light transmission.

\begin{figure*}[htb]
    \centering
\includegraphics[width=.9\textwidth,height=.9\textheight,keepaspectratio]{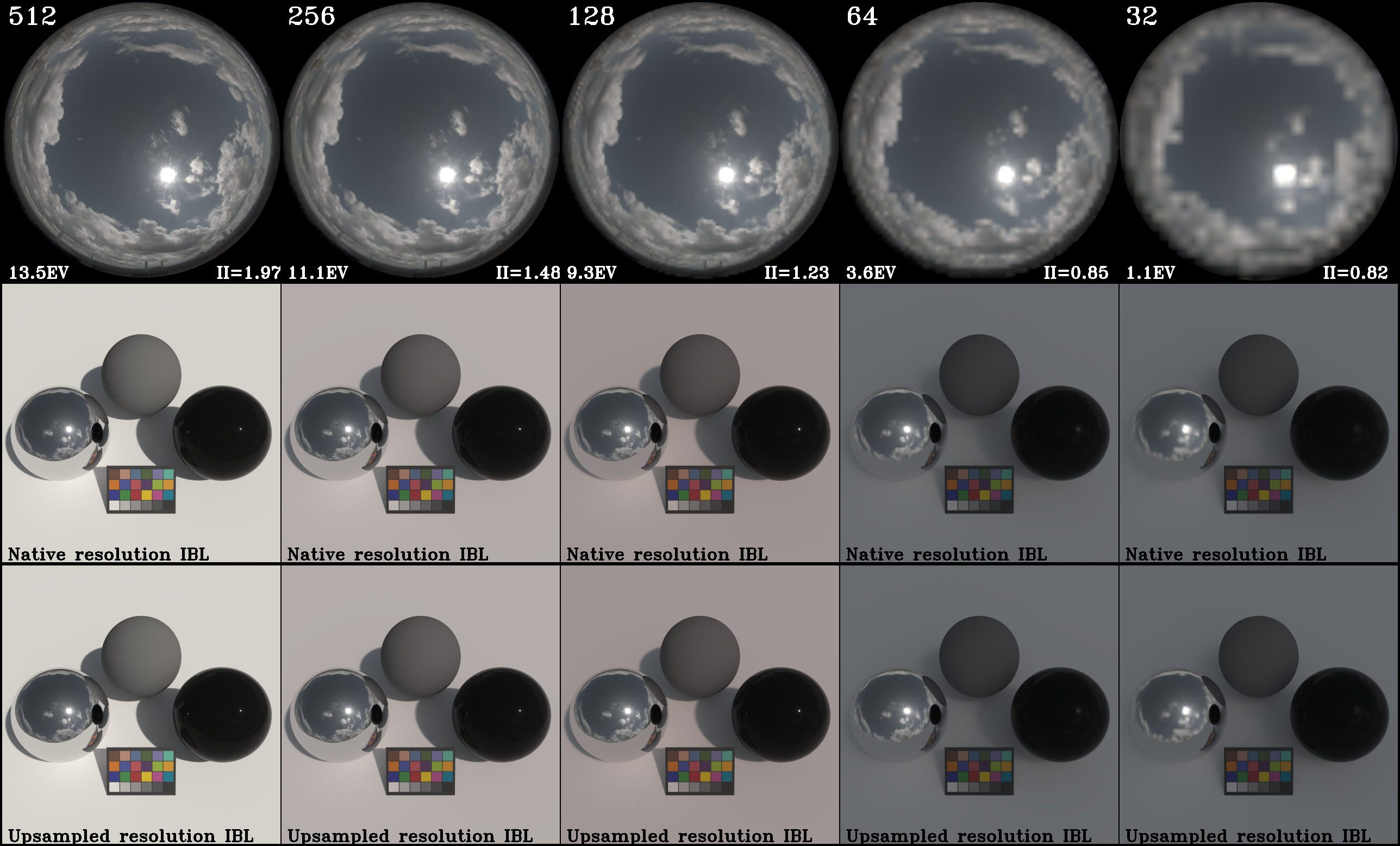}
    \caption{
    IBL renderings with inter-linear upsampled sky-angular environment-maps, demonstrating that upsampling does not recover illumination or exposure range lost in downsampling.
    HDRDB sample from June 7th, 2016 at 1:54PM \cite{LavalHDRdb}.
    }
    \label{app:fig::scaling_renders}
\end{figure*}

In \cref{app:tab::resolution_DR_II} we demonstrate the effect of downsampling by calculating the Exposure Range (EV) and Illumination ($\oiint_I$) at each halving of resolution.
We calculate the upsampled Exposure Range (EV) and Illumination ($\oiint_I$), by upsampling the downsampled environment-maps back to their original native resolution.
Previous works have claimed that environment-maps can be interpolated to more favourable resolutions for IBL rendering \cite{SKYGAN_2022}.
This experiment demonstrates that upsampling is futile as the exposure range and illumination lost in downsampling are not recovered.
As shown in \cref{app:fig::scaling_renders}, upsampling has little or no impact on IBL renderings.

\subsection{Format Conversions}
\label{app:sec::envmaps::format_conversions}

\begin{table}[!ht]
\centering
\caption{Retention of exposure range (EV) in environment-map format conversion using inter-linear interpolation.}
\begin{tabular}{lr|ccc|}
\cline{3-5}
 & \multicolumn{1}{l|}{} & \multicolumn{3}{c|}{To} \\ \cline{3-5}
                           & \multicolumn{1}{l|}{} & \multicolumn{1}{c|}{latlong} & \multicolumn{1}{c|}{sky-latlong} & sky-angular \\ \hline
\multicolumn{1}{|c|}{}     & latlong               & \multicolumn{1}{c|}{\textbf{13.34 (100\%)}}   & \multicolumn{1}{c|}{13.34 (100\%)}       & 12.96 (77\%)      \\ \cline{2-5}
\multicolumn{1}{|c|}{From} & sky-latlong           & \multicolumn{1}{c|}{13.34 (100\%)}   & \multicolumn{1}{c|}{\textbf{13.34 (100\%)}}       & 12.96 (77\%)       \\ \cline{2-5}
\multicolumn{1}{|l|}{}     & sky-angular           & \multicolumn{1}{c|}{13.34 (93\%)}   & \multicolumn{1}{c|}{13.34 (93\%)}       & \multicolumn{1}{c|}{\textbf{13.44 (100\%)}}       \\ \hline
\end{tabular}
\label{app:tab::conversion_EV_loss}
\end{table}

\begin{table}[!ht]
\centering
\caption{Retention of illumination ($\oiint_I$) in environment-map format conversion using inter-linear interpolation.}
\begin{tabular}{lr|ccc|}
\cline{3-5}
 & \multicolumn{1}{l|}{} & \multicolumn{3}{c|}{To} \\ \cline{3-5}
                           & \multicolumn{1}{l|}{} & \multicolumn{1}{c|}{latlong} & \multicolumn{1}{c|}{sky-latlong} & sky-angular \\ \hline
\multicolumn{1}{|c|}{}     & latlong               & \multicolumn{1}{c|}{\textbf{1.95}}    & \multicolumn{1}{c|}{1.95}        & \multicolumn{1}{c|}{1.94}        \\ \cline{2-5}
\multicolumn{1}{|c|}{From} & sky-latlong           & \multicolumn{1}{c|}{1.95}    & \multicolumn{1}{c|}{\textbf{1.95}}        & \multicolumn{1}{c|}{1.94}        \\ \cline{2-5}
\multicolumn{1}{|l|}{}     & sky-angular           & \multicolumn{1}{c|}{1.95}    & \multicolumn{1}{c|}{1.95}        & \multicolumn{1}{c|}{\textbf{1.95}}        \\ \hline
\end{tabular}
\label{app:tab::conversion_II_loss}
\end{table}

\begin{table}[!ht]
\centering
\caption{Retention of exposure range (EV) in environment-map format conversion using inter-linear interpolation and $\mu$-lawLog$_2$ ($T_{\mu \log_2}$) tone-mapping images during format conversion.}

\begin{tabular}{lr|ccc|}
\cline{3-5}
 & \multicolumn{1}{l|}{} & \multicolumn{3}{c|}{To} \\ \cline{3-5}
                           & \multicolumn{1}{l|}{} & \multicolumn{1}{c|}{latlong} & \multicolumn{1}{c|}{sky-latlong} & sky-angular \\ \hline
\multicolumn{1}{|c|}{}     & latlong               & \multicolumn{1}{c|}{\textbf{13.34 (100\%)}}   & \multicolumn{1}{c|}{13.34 (100\%)}       & 12.71 (65\%)      \\ \cline{2-5}
\multicolumn{1}{|c|}{From} & sky-latlong           & \multicolumn{1}{c|}{13.34 (100\%)}   & \multicolumn{1}{c|}{\textbf{13.34 (100\%)}}       & 12.71 (65\%)       \\ \cline{2-5}
\multicolumn{1}{|l|}{}     & sky-angular           & \multicolumn{1}{c|}{13.27 (89\%)}   & \multicolumn{1}{c|}{13.27 (89\%)}       & \multicolumn{1}{c|}{\textbf{13.44 (100\%)}}       \\ \hline
\end{tabular}%
\label{app:tab::conversion_EV_loss_tonemapped}
\end{table}

\begin{table}[!ht]
\centering
\caption{Retention of illumination ($\oiint_I$) in environment-map format conversion using inter-linear interpolation and $\mu$-lawLog$_2$ ($T_{\mu \log_2}$) tone-mapping images during format conversion.}

\begin{tabular}{lr|ccc|}
\cline{3-5}
 & \multicolumn{1}{l|}{} & \multicolumn{3}{c|}{To} \\ \cline{3-5}
                           & \multicolumn{1}{l|}{} & \multicolumn{1}{c|}{latlong} & \multicolumn{1}{c|}{sky-latlong} & sky-angular \\ \hline
\multicolumn{1}{|c|}{}     & latlong               & \multicolumn{1}{c|}{\textbf{1.95}}    & \multicolumn{1}{c|}{1.95}        & \multicolumn{1}{c|}{1.66}        \\ \cline{2-5}
\multicolumn{1}{|c|}{From} & sky-latlong           & \multicolumn{1}{c|}{1.95}    & \multicolumn{1}{c|}{\textbf{1.95}}        & \multicolumn{1}{c|}{1.66}        \\ \cline{2-5}
\multicolumn{1}{|l|}{}     & sky-angular           & \multicolumn{1}{c|}{1.44}    & \multicolumn{1}{c|}{1.44}        & \multicolumn{1}{c|}{\textbf{1.95}}        \\ \hline
\end{tabular}%
\label{app:tab::conversion_II_loss_tonemapped}
\end{table}

As shown in \cref{app:tab::conversion_EV_loss}, converting environment-maps from latlong or sky-latlong to sky-angular format retains 77\%+ of the exposure range (EV).
Note, if expressed as a percentile of raw intensity in linear space, the loss of exposure range is more apparent.
As shown in \cref{app:tab::conversion_II_loss}, converting environment-maps from latlong or sky-latlong to sky-angular formats retains 99.5\%+ of illumination ($\oiint_I$).

We repeat this experiment with $\mu$-lawLog$_2$ ($T_{\mu \log_2}$) tone-mapping during format conversions.
As shown in \cref{app:tab::conversion_EV_loss_tonemapped,app:tab::conversion_II_loss_tonemapped}, converting between environment-map formats in compressed space negatively impacts exposure range (EV) and illumination ($\oiint_I$), with retention falling to 65\% and 85\% respectively.
This impact is exacerbated further with increased HDRI exposure range and tone-mapper aggressiveness.

Format conversions are unavoidable as HDRDB provides FDR HDRI in latlong format, models require HDRI sky-angular format for skydome continuity, and Blender \cite{blender} currently supports only sphere and equirectangular (latlong) environment-maps for IBL renderings.
With this in mind, we demonstrated that format conversions avoid where possible and should be performed in uncompressed HDR space.

\section{Metrics}
\label{app:sec::metrics}

\subsection{Metrics: Hinge Loss}
\label{app:sec::metrics:::hinge}

\begin{figure}[H]
    \centering
    \includegraphics[width=0.5\linewidth,keepaspectratio]{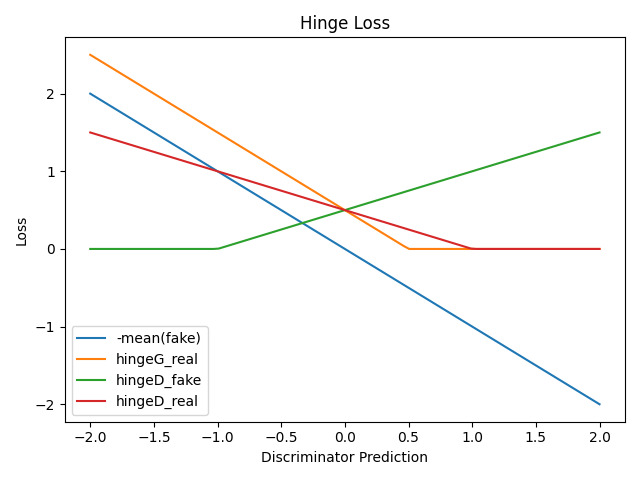}
    \caption{Hinge loss for discriminator and generator}
    \label{app:fig::hinge_loss}
\end{figure}

As defined for the generator in \cref{eq:loss_hinge_G} and for the discriminator in \cref{eq:loss_hinge_D}, we propose a hinge loss
with an offset ($\delta$) to control saturation points.
As illustrated in \cref{app:fig::hinge_loss}, this enables shifting of the pivot point ($1$ for discriminator \textit{hingeD} loss; $0.5$ for generator \textit{hingeG} loss) such that the generator sample ($p(I_{real})$) is encouraged to fool the discriminator rather than confuse it.

\begin{equation}
    \mathcal{L}_{real} = \frac{1}{n}\sum^{n} ReLU(\delta - p(I_{real}))
    \label{eq:loss_hinge_G}
\end{equation}

\begin{equation}
    \begin{split}
        \mathcal{L}_{real} = \frac{1}{n}\sum^{n} ReLU(\delta - p(I_{real})) \\
        \mathcal{L}_{fake} = \frac{1}{n}\sum^{n} ReLU(\delta + p(I_{fake})) \\
        \mathcal{L} = \left(\mathcal{L}_{real} + \mathcal{L}_{fake} \right) / 2
    \end{split}
    \label{eq:loss_hinge_D}
\end{equation}

\subsection{Metrics: Exposure Range and Illumination}
\label{app:sec::metrics:::EV_II}

We plot the sensitivity of our proposed Exposure Range (EV) and Integrated Illumination ($\oiint_I$) metrics to exposure variations in \cref{app:fig::metrics_EV_sensitivity_1}.
EV measures an environment-map's maximum illumination intensity with insensitivity to both resolution and stochasticity in style and positioning.
 $\oiint_I$ measures an environment-map's cumulative illumination energy with respect to resolution.
Though the weighting of each pixels is dependent on the environment-map's solid angles, exact positioning is not required, and the metric is generally insensitive to stochasticity in style and positioning.
In sky-angular format, $\oiint_I$ is invariant to a larger set of augmentations including rotations around the zenith and flips.

\begin{figure}[!htb]
    \centering
      \includegraphics[width=0.5\linewidth,keepaspectratio]{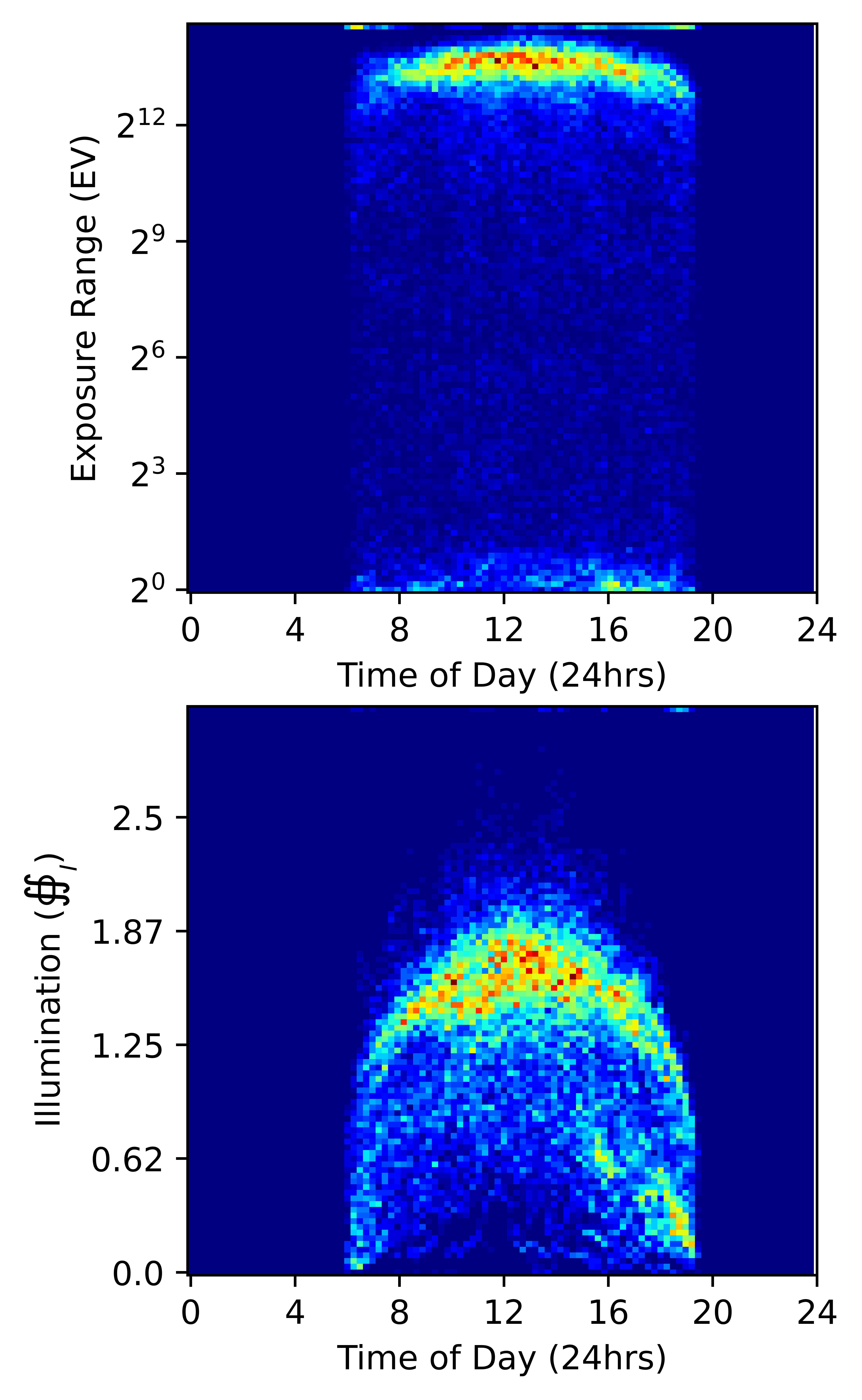}

    \caption{Illustration of Exposure Range (EV) and Illumination ($\oiint_I$) day-cycles in HDRDB \cite{LavalHDRdb}. }
    \label{app:fig::plot_daily_cycles}
\end{figure}

In this work, we report the Exposure Range (EV) and Illumination ($\oiint_I$) as the mean of the sampled subset.
This generalizes a model's ability to match ground truth physically captured HDRI, but belies the model's ability to follow natural variation within a subset of samples.
As shown in \cref{app:fig::plot_daily_cycles}, day-cycles result in a cyclic pattern for Exposure Range (EV) and Illumination ($\oiint_I$).
We do not distinguish samples by (among other) solar-elevation, solar-radiance, or cloud cover, but careful consideration should be given to ensuring balance between subsets.

In quantitative results, we report better/best Exposure Range (EV $\leftarrow$) and Illumination ($\oiint_I \leftarrow$) as the closest in absolute distance to the values calculated from the subset's ground truth physically captured HDRI (indicated by $\leftarrow$, closest is best).
For Exposure Range (EV $\leftarrow$), the absolute distance is calculated using linear space intensity ($i$) as $i = 2^{EV}-1$.

Given natural variability in illumination and stochasticity in textured skies, we do not propose using Exposure Range (EV) and/or Illumination ($\oiint_I$) as losses to guide model learning.
Our experiments show that EV- and $\oiint_I$-losses would act similarly to an $L_1$-loss, overemphasizing the sun and resulting in blurred atmospheric formations (clouds) and over-saturation of the skydome.

\subsection{Metrics: PSNR}
\label{app:sec::metrics:::PSNR}

\begin{figure}[htb]
    \centering
    \includegraphics[width=\linewidth,keepaspectratio]{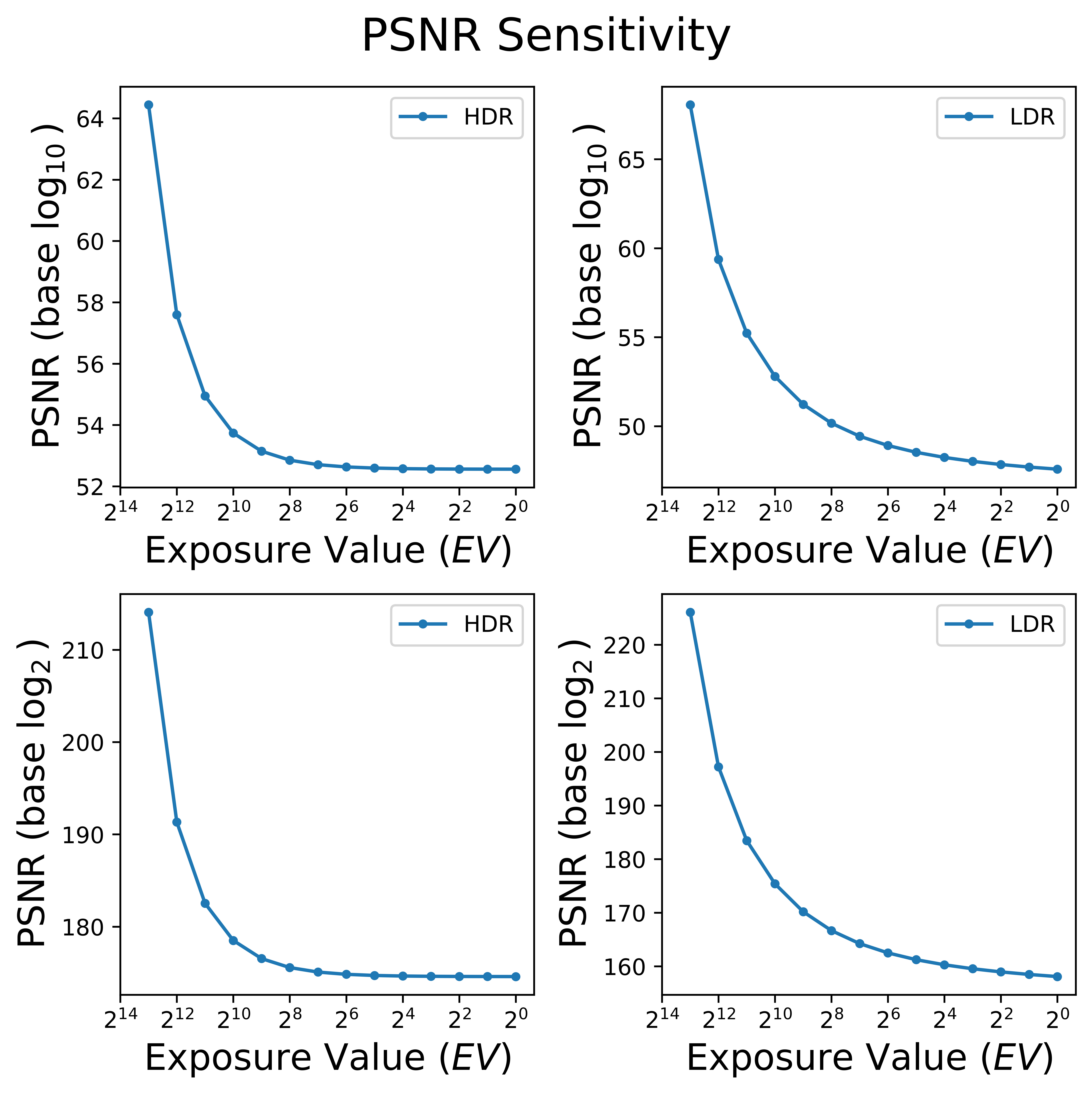}
    \caption{
    PSNR sensitivity to iteratively clipping Exposure Range (EV).
    Selecting a lower logarithmic base improves sensitivity in both HDR and compressed LDR ($T_\gamma$) space.
    Clipping by EV$\ge$ 9 only impacts the four-pixel solar disk.
    HDRDB sample form June 7th, 2016 at 12:52 PM \cite{LavalHDRdb}.
    }
    \label{app:fig::ER_PSNR}
\end{figure}

PSNR is commonly used to evaluate HDR images and, as shown in \cref{app:fig::ER_PSNR,app:fig::metrics_EV_sensitivity_2}, offers sensitivity to variance in exposure range.
Though viable for many HDR reconstruction applications, we note that PSNR is a poor measure of sky-model performance.

Per \cref{eq:PSNR}, PSNR is reliant on Mean Squared Error (MSE; $L_2$) and analytically measures reconstruction exactness.
Given the gross differences in class value distributions (i.e., exposure range of the skydome vs. sun), this results in a quantization heavily influenced by high-exposure values.
As shown in \cref{app:fig::ER_PSNR}, PSNR is sensitive to iteratively clipping exposure range, but when clipping to $EV\ge9$ only the four-pixel solar disk of the $512^2$ environment map is impacted.
Furthermore, atmospheric formations (e.g., clouds) and the solar disk generally cannot be deterministically reproduced or localized, resulting in style and positional stochasticity which is unimpeding of photorealism but penalized by PSNR.

Therefore, PSNR is a poor measure of sky-model performance due to over-representation of solar regions and penalizing unimpeding stochasticity in style and positioning.

\begin{align}
    \text{MSE}(X,Y) &= \frac{1}{MN}\sum_{ij}^{MN}{\left(X_{ij}-Y_{ij}\right)^2} \\
    R &= \text{data range (max-min)}\\
    \text{PSNR}_{\log_{10}}(X,Y) &= 10\log_{10}\frac{R^2}{\text{MSE}(X,Y)}
    \label{eq:PSNR}
\end{align}

\subsection{Other Metrics}
\label{app:sec::metrics:::metrics_other}

\begin{figure*}[htb]
    \centering
    \includegraphics[width=\textwidth,height=0.85\textheight,keepaspectratio]{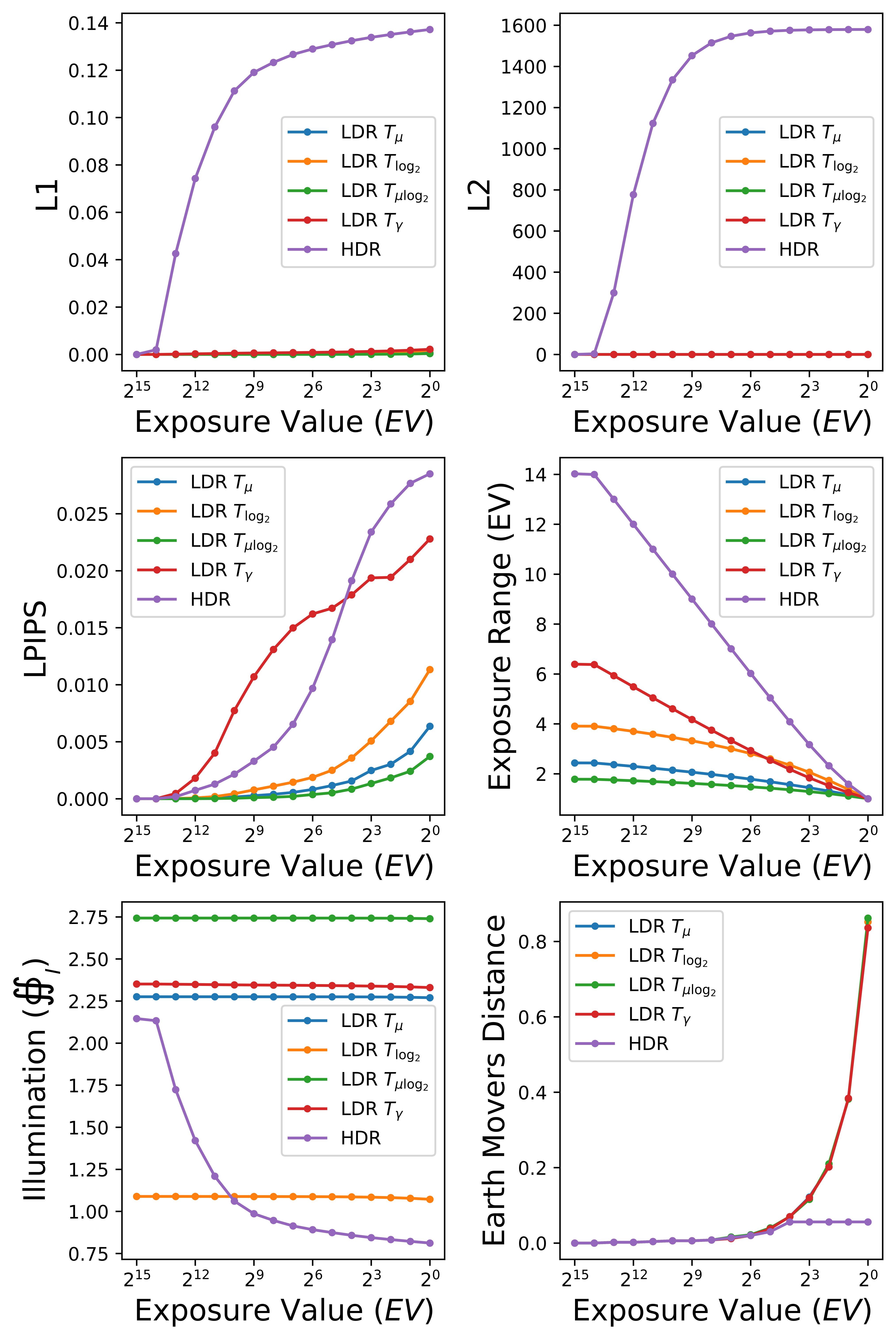}
    \caption{
    Mean Average Error (MAE; $L_1$),
    Mean Squared Error (MSE; $L2$),
    Learned Perceptual Image Patch Similarity (LPIPS),
    Exposure Range (EV; \textbf{ours}),
    Integrated Illumination ($\oiint_I$; \textbf{ours}),
    and Earth Movers Distance (EMD; Wasserstein distance)
    sensitivity to exposure range (EV).
    With the exception of EV, metrics in compressed LDR (unclipped) offer little or no sensitivity to exposure range.
    Most metrics demonstrate some sensitivity in HDR space, but values are not necessarily reliable quantifications of exposure.
    $L_1$, $L_2$, and EMD sensitivity is dependent on environment-map resolution, therefore sensitivity to high-exposure regions (solar disk) diminishes with increasing resolution.
    $L_1$, $L_2$, LPIPS and EMD sensitivity assume an objective of reconstruction exactness and therefore stochasticity in style and positioning introduce noise.
    Both our proposed metrics EV and $\oiint_I$ are insensitive to resolution.
    Furthermore, EV is fully insensitive and $\oiint_I$ is generally insensitive to stochasticity in style and positioning.
    HDRDB sample form June 7th, 2016 at 12:52 PM \cite{LavalHDRdb}.
    }
    \label{app:fig::metrics_EV_sensitivity_1}
\end{figure*}

\begin{figure*}[ht]
    \centering
    \includegraphics[width=\textwidth,keepaspectratio]{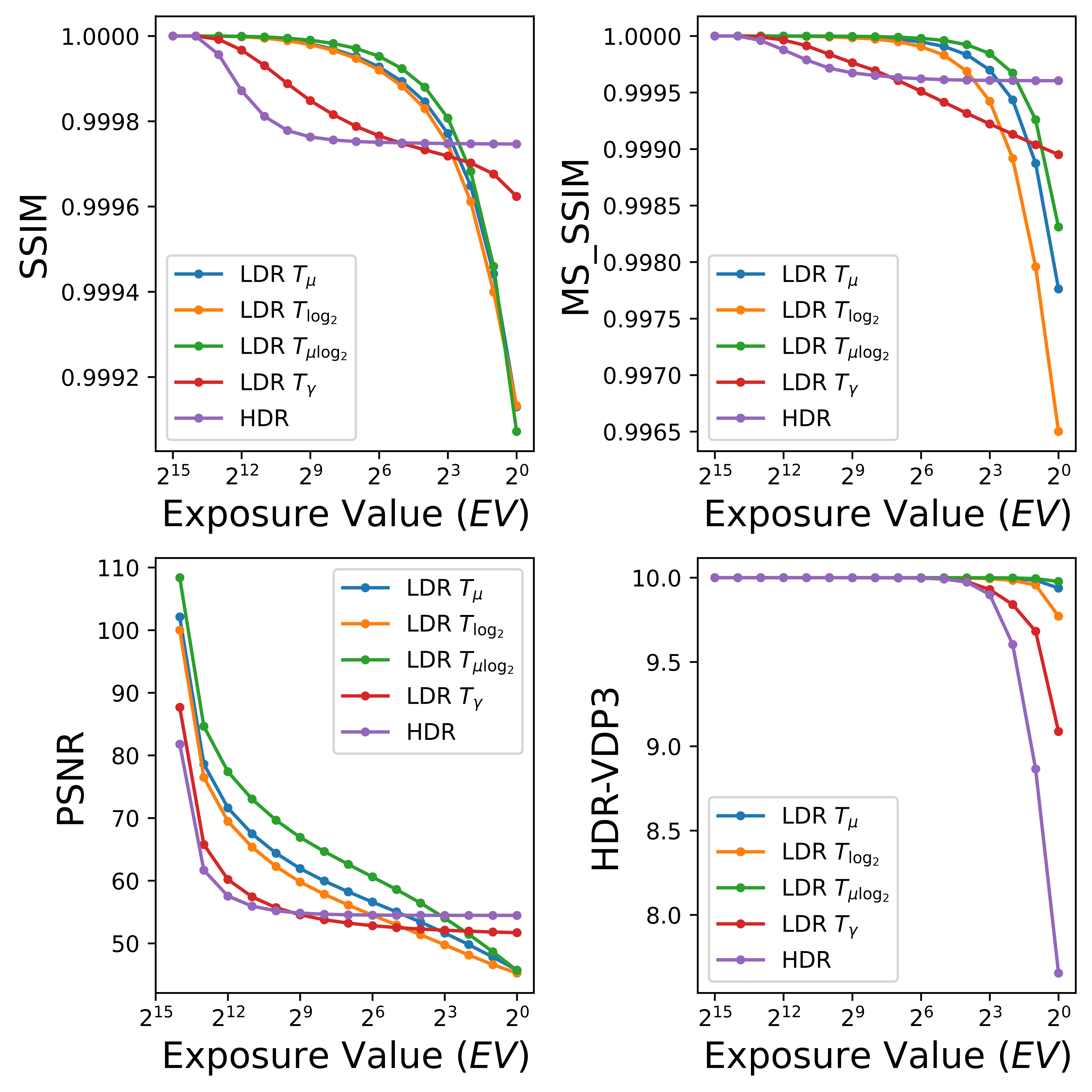}
    \caption{
    SSIM, Multi-Scale SSIM (MS\_SSIM), PSNR (base $\log_{10}$) and HDR-VDP3 sensitivity to exposure range (EV).
    PSNR, given its reliance on $L_2$, is dependent on environment-map resolution and sensitive to stochasticity in style and positioning.
    SSIM, MS\_SSIM and HDR-VDP3 offer no meaningful measure of environment-map exposure range in both compressed (unclipped) LDR and HDR space.
    HDRDB sample form June 7th, 2016 at 12:52 PM \cite{LavalHDRdb}.
    }
    \label{app:fig::metrics_EV_sensitivity_2}
\end{figure*}

In \cref{app:fig::ER_PSNR,app:fig::metrics_EV_sensitivity_1,app:fig::metrics_EV_sensitivity_2} we plot the sensitivity of various metrics to changes in exposure range.
We find that SSIM, Multi-Scale SSIM (MS\_SSIM), and HDR-VDP3 offer no meaningful sensitivity to environment-map exposure range in both compressed (unclipped) LDR and HDR space.
With the exception of EV, we find that metrics in compressed (unclipped) LDR space offer little or no sensitivity to changes in exposure range.
We note that the values from
Mean Average Error (MAE; $L_1$),
Mean Squared Error (MSE; $L2$),
and Earth Movers Distance (EMD; Wasserstein distance)
are dependent on environment-map resolution and therefore have diminishing sensitivity to the high-exposure solar disk which shrinks in angular diameter with increasing resolution.
Furthermore, $L_1$, $L_2$, Learned Perceptual Image Patch Similarity (LPIPS)
\footnote{
    LPIPS is intended for $\left\{I_{3,ij} \subset \mathbb{R} | -1 \leq I_{3,ij} \leq 1 \right\}$ input \cite{LPIPS}.
    To avoid truncation and unbound values, we implement LPIPS with $\left\{I_{3,ij} \subset \mathbb{R} | -1 \leq I_{3,ij} \right\}$ input.
}
and EMD sensitivity assume an objective of reconstruction exactness and therefore stochasticity in style and positioning is penalized.

From literature, Zhang et al.\ evaluate environment-map illumination with $L_2$ segmented to solar regions \cite{Zhang_2017_ICCV}.
Their results penalize variability and are proportional to angular diameter of the solar region (of unknown size) where the solar disk may be as little as one-pixel.

\clearpage
\section{Dataset}
\label{app:sec::dataset}

\begin{figure}[htpb]
    \centering
    \begin{subfigure}{.45\linewidth}
        \centering
        \includegraphics[width=\linewidth]{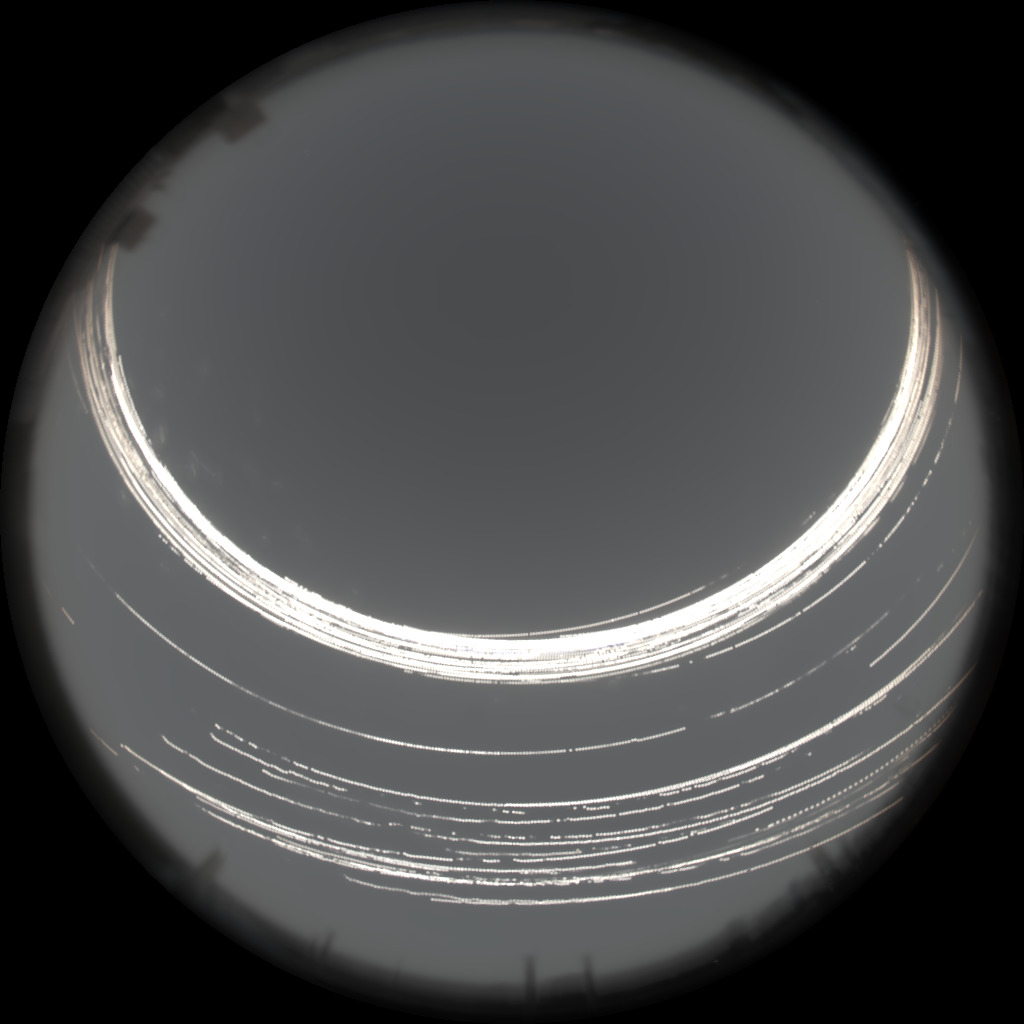}
        \caption{Mean Skydome}
        \label{app:fig::img_mean_skydome}
    \end{subfigure}
    \begin{subfigure}{.45\linewidth}
        \centering
        \includegraphics[width=\linewidth]{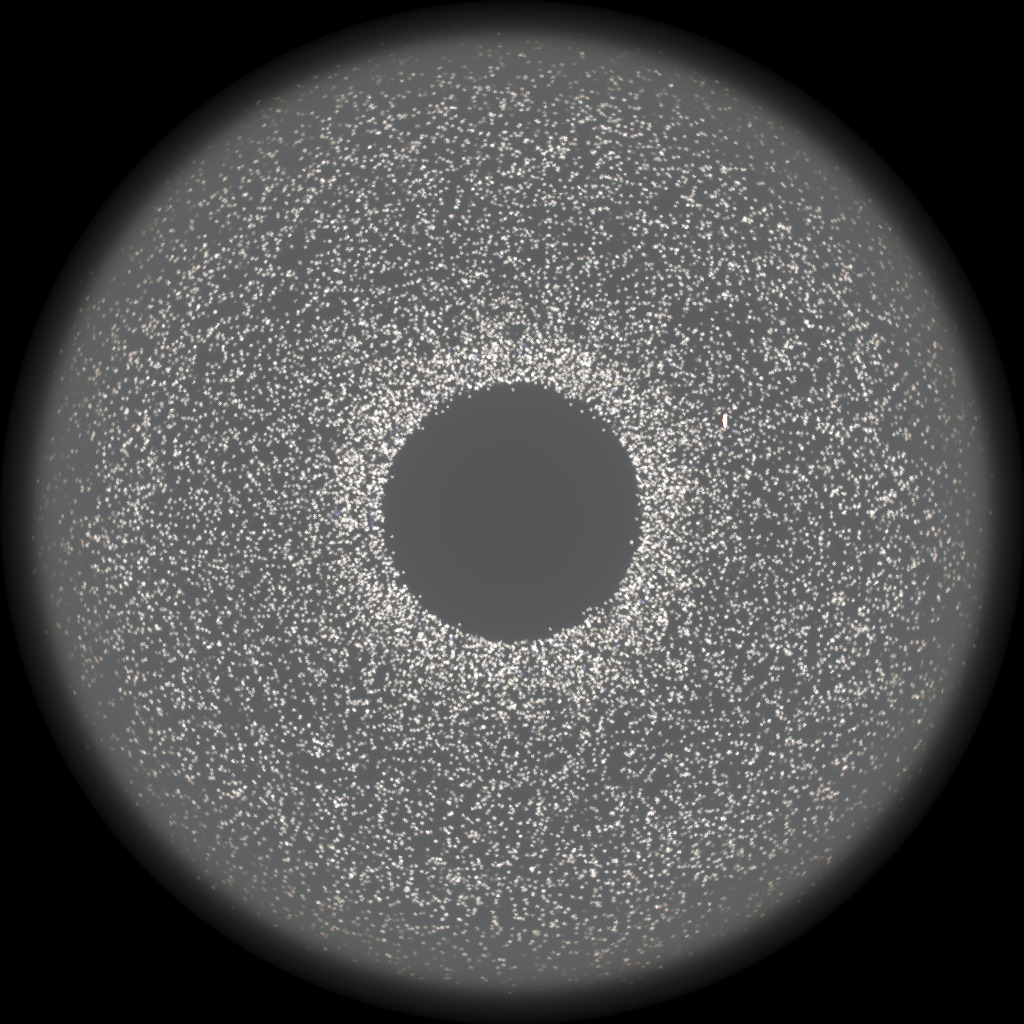}
        \caption{Mean Augmentation}
        \label{app:fig::img_mean_augmented_skydome}
    \end{subfigure}
    \caption{HDRDB average (left) and augmented average (right) skydomes \cite{LavalHDRdb}}
\end{figure}

The Laval HDR Sky database (HDRDB, \cite{LavalHDRdb}) consists of 34K+ HDR images captured in Quebec City, Canada at various intervals between 2014 and 2016 using a capture method synonymous to that proposed by Stumpfel et al.~\cite{STUMPFEL_HDR_Sky_Capture}.

\begin{figure}[htbp]
    \centering
    \includegraphics[width=0.5\linewidth]{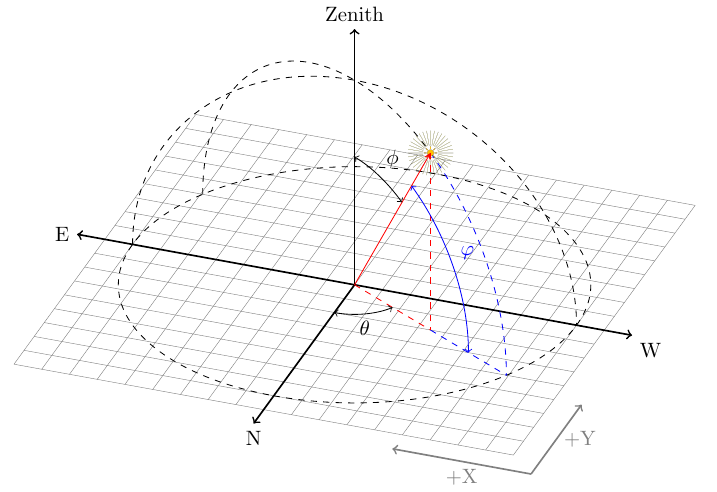}
    \caption{Sky-Angular environment-map}
    \label{app:fig::dia_skyangular}
\end{figure}

\subsection{Subsets}
\label{app:sec::dataset:::subsets}

\begin{figure*}[ht]
    \centering
    \includegraphics[width=.9\textwidth,height=.9\textheight,keepaspectratio]{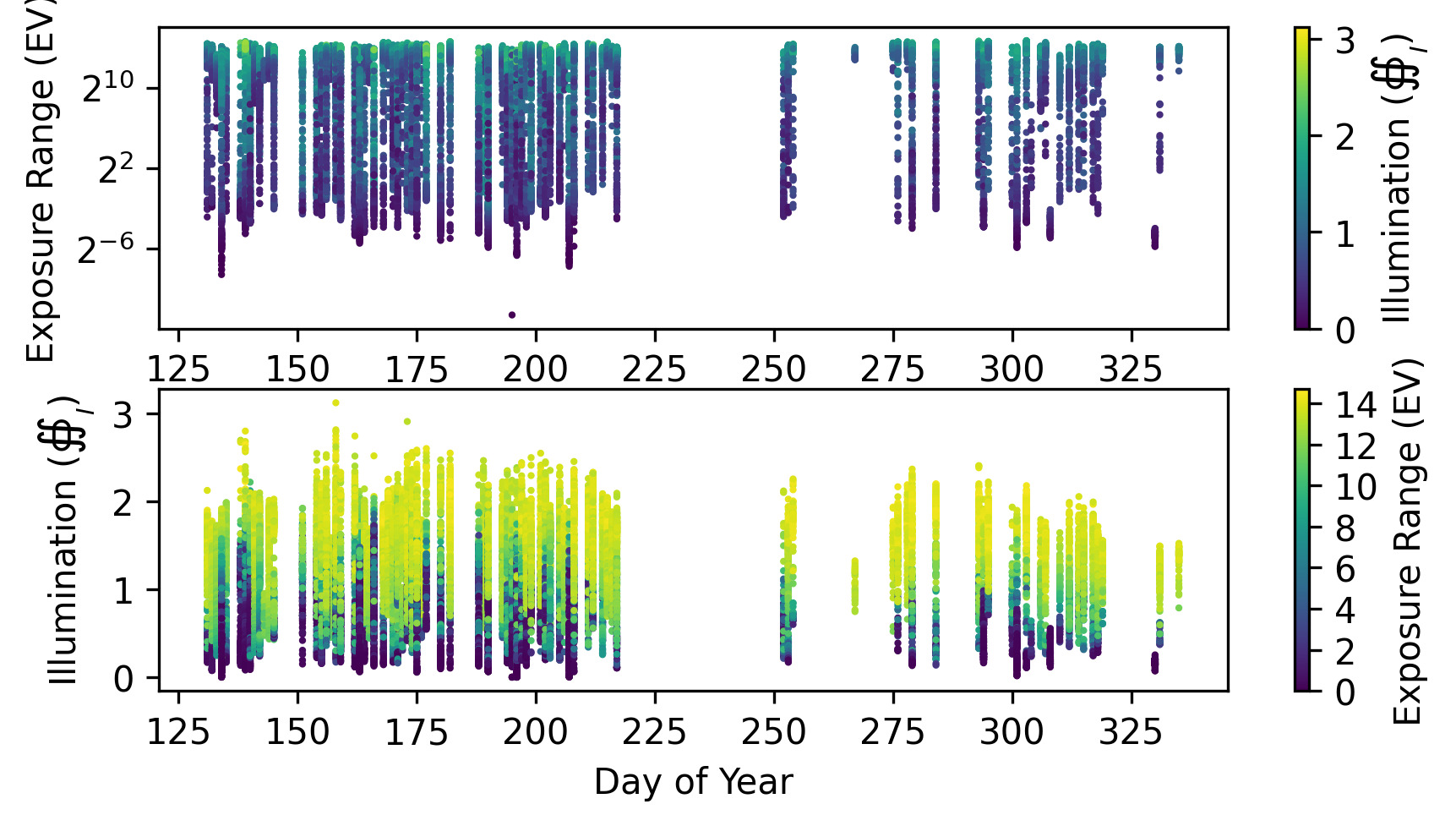}
    \caption{Annual representation of HDRDB exposure range and illumination \cite{LavalHDRdb}}
    \label{app:fig::hdrdb_DOY}
\end{figure*}

\begin{figure*}[ht]
    \centering
    \includegraphics[width=.9\textwidth,height=.9\textheight,keepaspectratio]{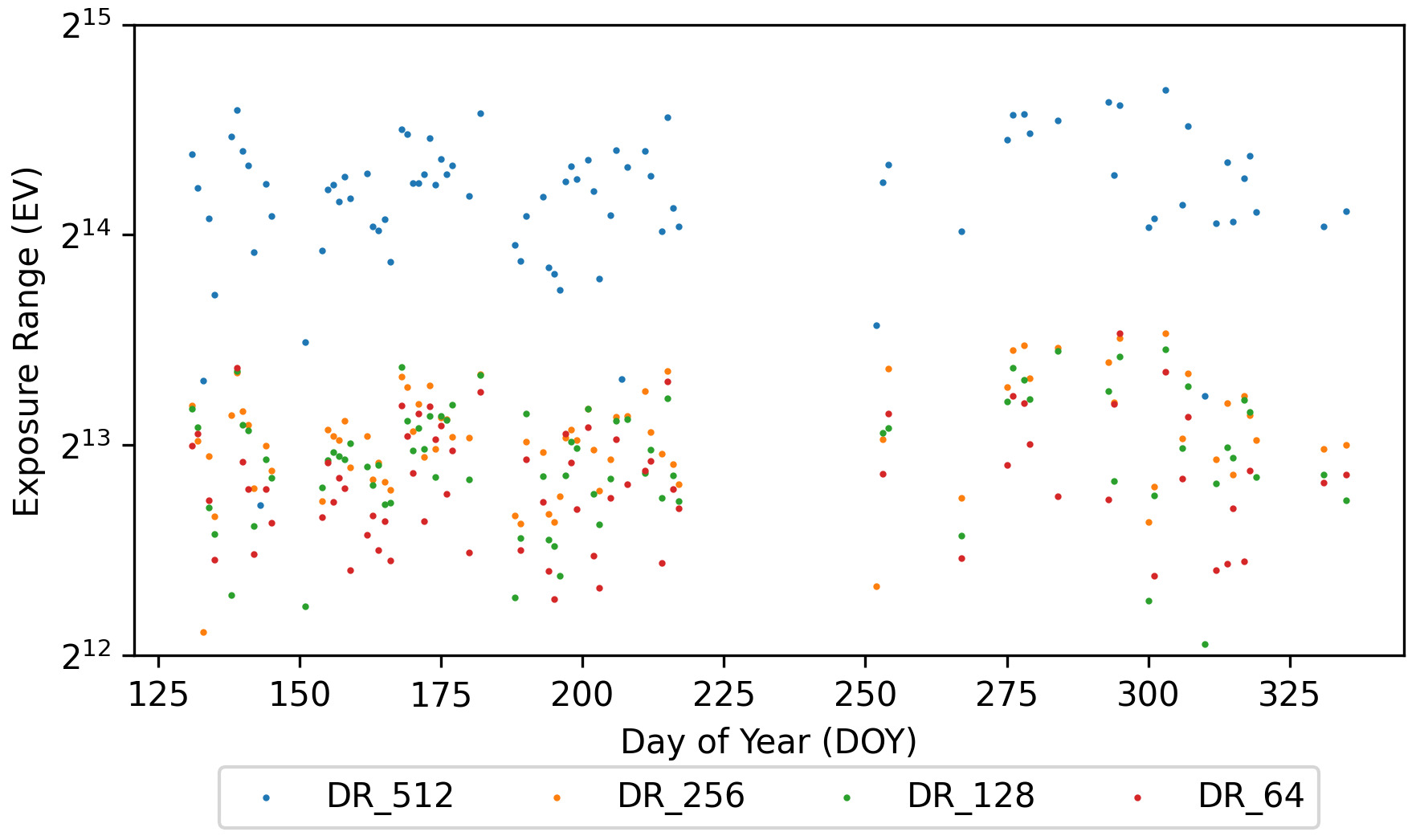}
    \caption{Annual representation of HDRDB exposure range at various resolutions (inter-linear interpolation) \cite{LavalHDRdb}, highlighting
    the loss of exposure range in downsampling.}
    \label{app:fig::hdrdb_DR}
\end{figure*}

From the database — the mean of which is shown in \cref{app:fig::img_mean_skydome} — we augment the dataset with flips and rotations around the zenith.
As shown in \cref{app:fig::img_mean_augmented_skydome}, this increases solar placement coverage and enables the generation of skies outside of HDRDB.
We split the dataset into training, validation, and testing subsets by arbitrarily splitting by date of capture.
This ensures that each subset has a random assortment of images from each year and season of capture, but does guarantee equal distribution of samples in terms of (among other) solar-elevation, solar-radiance, or cloud cover.
HDRDB's distribution of samples for Exposure Range (EV) and Illumination ($\oiint_I$) is illustrated in \cref{app:fig::hdrdb_DOY}.
Loss of HDRDB Exposure Range (EV) in downsampling is illustrated in \cref{app:fig::hdrdb_DR}.

To prevent discontinuity in generated skydomes (see \cref{app:sec::text2light:::seams}), we convert HDRI from latlong to sky-angular format as shown in \cref{app:fig::dia_skyangular}.
For the purpose of collecting visual and quantitative results, border regions of captured and generated environment maps are zeroed per label segmentation.

\subsubsection{Experimental Subset}
Unless otherwise stated, experimental runs of AllSky are completed using an experimental subset consisting of 4,096 training, 2,238 validation, and 5,811 testing physically captured HDRI from HDRDB \cite{LavalHDRdb}.

This subset is manually pruned to remove undesirable images (e.g.\ raindrops, birds, and other lens obstructions) and heuristically pruned to retain only images where the sun is above the horizon (approximated at $10^\circ$ of solar elevation).

\subsubsection{Legacy Subset}
For fair comparison between AllSky, CloudNet and SkyGAN, we train all models against a subset of 25,243 training, 3,697 validation, and 3,519 testing physically captured HDRI from HDRDB \cite{LavalHDRdb}. This subset is manually pruned to remove undesirable images (e.g.\ raindrops, birds, and other lens obstructions).

\subsection{Dataset Assumptions}
\label{app:sec::dataset:::assumptions}
For the purpose of this work, we assume that HDRDB latlong physically captured FDR HDRI were correctly captured, calibrated to linear RGB colour space with BT.709 primaries and converted to latlong format without loss.
We therefore assume no artifacts were introduced and no alterations were made to the exposure range or illumination.

\subsection{Tone-mapping Operators}
\label{app:sec::dataset:::tonemapping}

\begin{figure*}[ht]
        \centering
        \includegraphics[width=\textwidth,keepaspectratio]{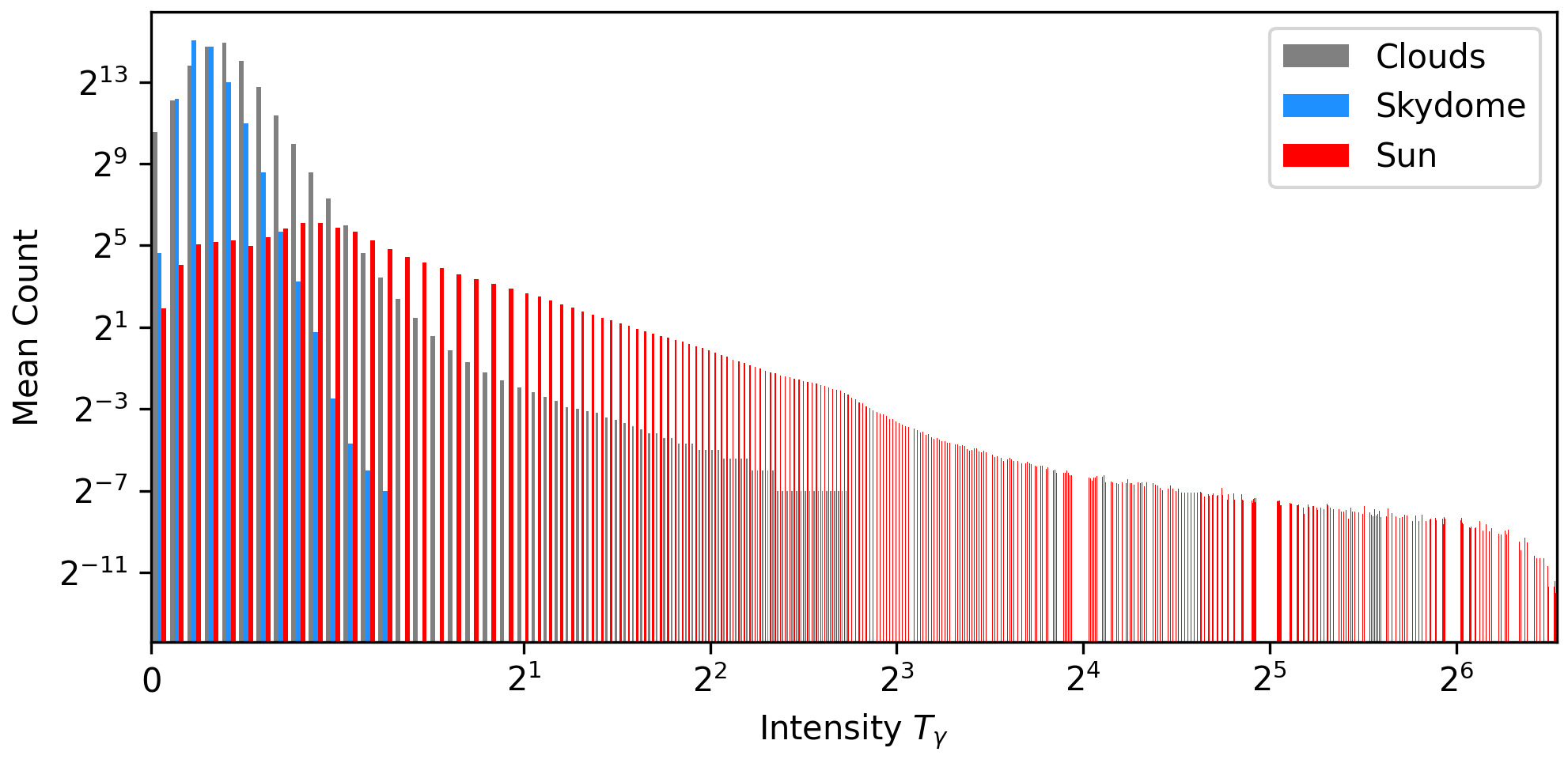}
        \caption{Histogram of Power-Law ($T_\gamma$) tone-mapped HDRDB \cite{LavalHDRdb}}
        \label{app:fig::hdrdb_hist_tm_gamma}
\end{figure*}
\begin{figure*}[ht]
        \centering
        \includegraphics[width=\textwidth,keepaspectratio]{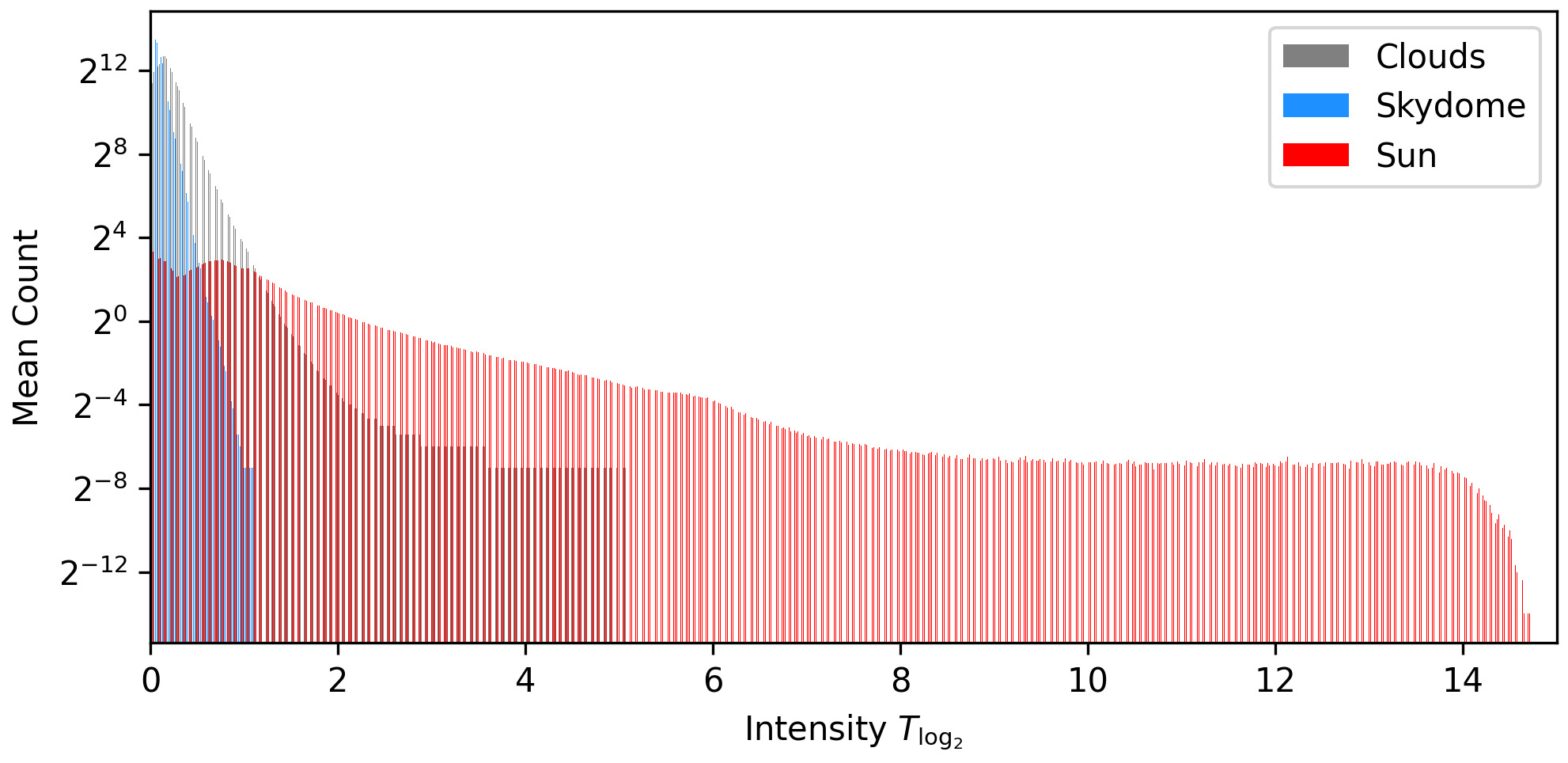}
        \caption{Histogram of logarithmic ($T_{log_2}$) tone-mapped HDRDB \cite{LavalHDRdb}}
        \label{app:fig::hdrdb_hist_tm_log2}
\end{figure*}
\begin{figure*}[ht]
        \centering
        \includegraphics[width=\textwidth,keepaspectratio]{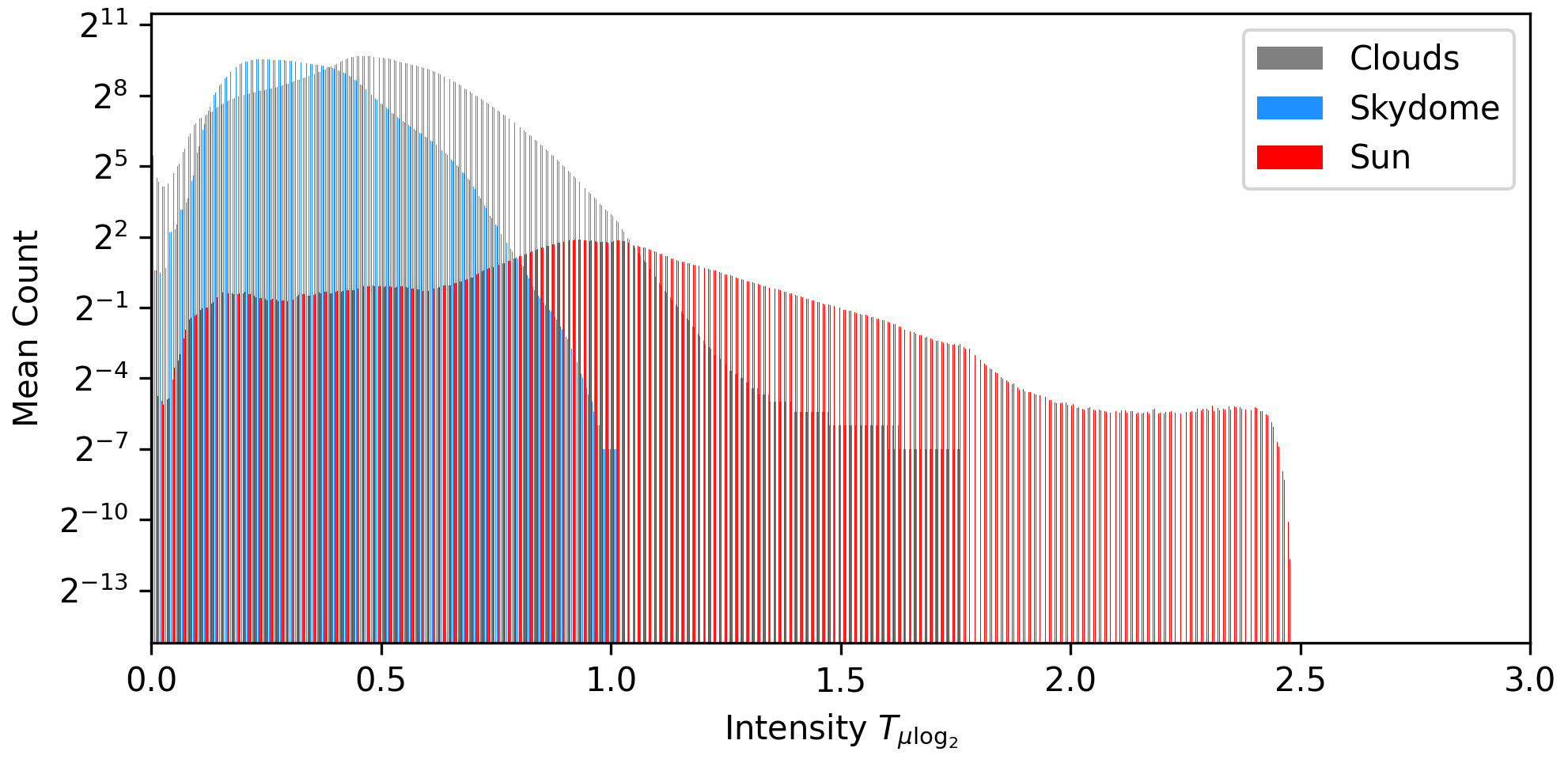}
        \caption{Histogram of $\mu$-lawLog$_2$ ($T_{\mu \log_2}$) tone-mapped HDRDB \cite{LavalHDRdb}}
        \label{app:fig::hdrdb_hist_tm_mixed}
\end{figure*}
\begin{figure*}[ht]
        \centering
        \includegraphics[width=\textwidth,keepaspectratio]{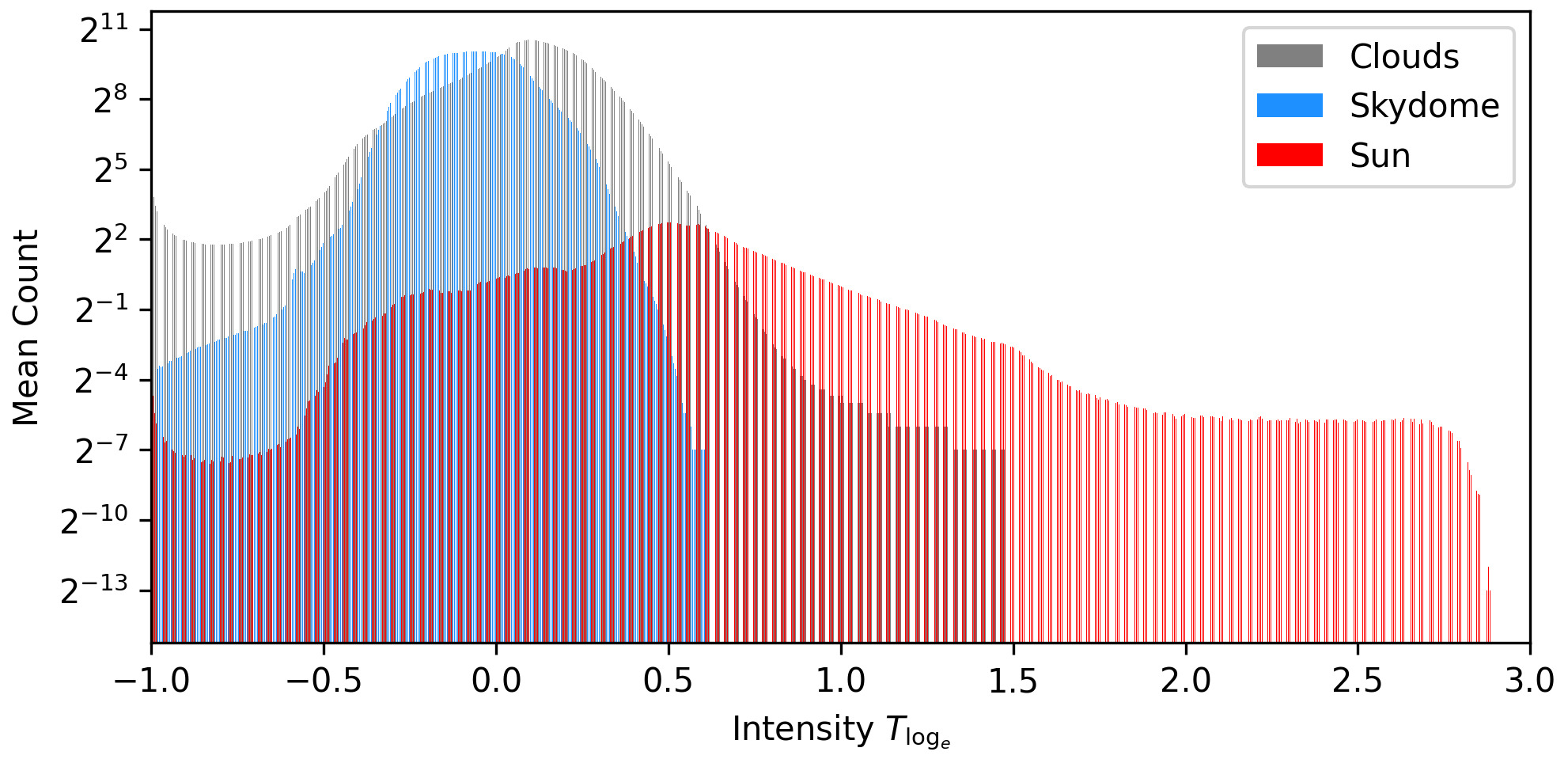}
        \caption{Histogram of natural logarithmic (${T}_{\log_e}$) tone-mapped HDRDB \cite{LavalHDRdb}}
        \label{app:fig::hdrdb_hist_tm_naturallog}
\end{figure*}
\begin{figure*}[ht]
        \centering
        \includegraphics[width=\textwidth,keepaspectratio]{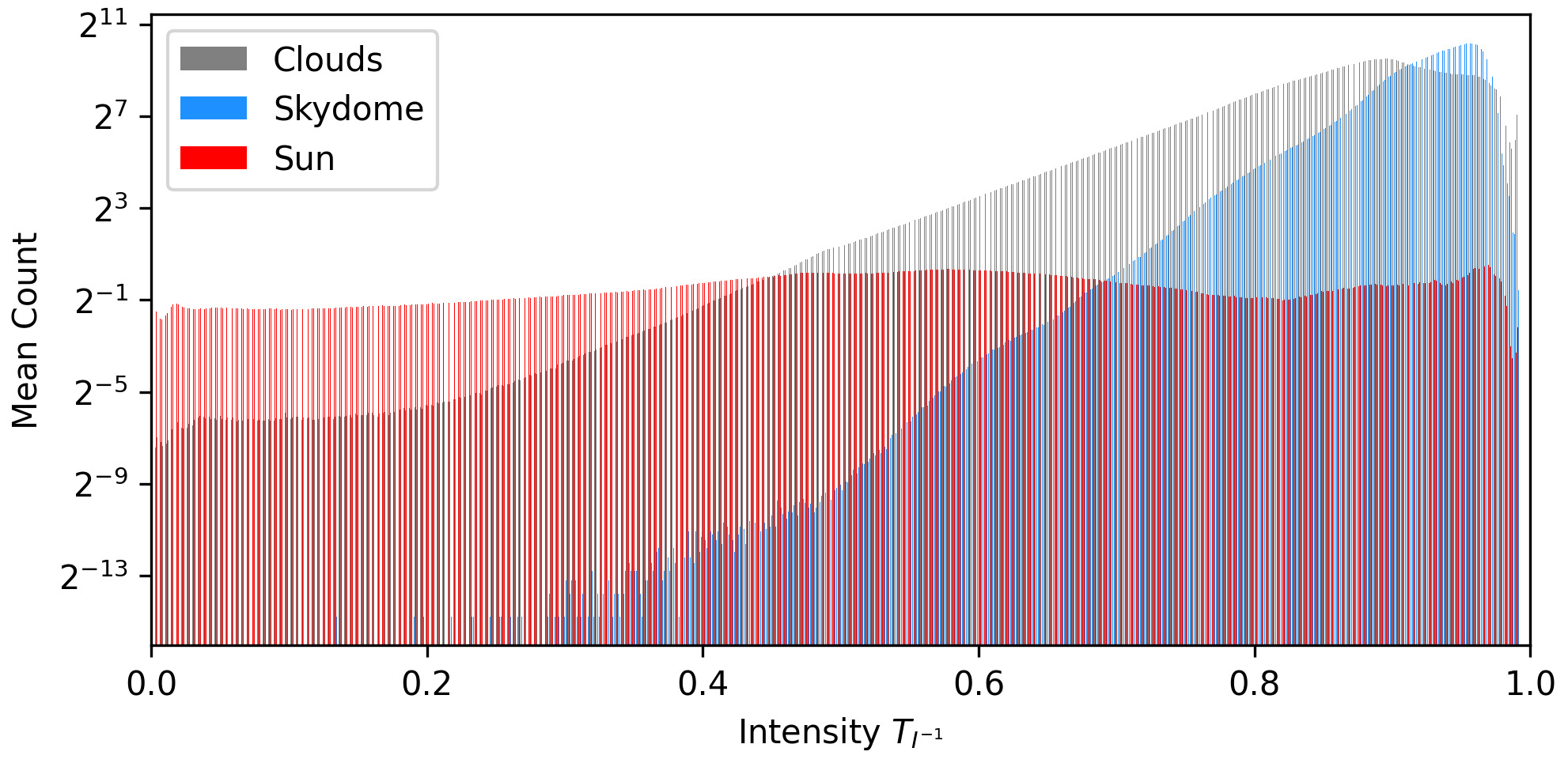}
        \caption{Histogram of inverted ($T_{I^{-1}}$) tone-mapped HDRDB \cite{LavalHDRdb}}
        \label{app:fig::hdrdb_hist_tm_inverted}
\end{figure*}

For visualization of LDR-space intensity for respective tone-mapping operators, we include the mean histogram from tone-mapping HDRDB (all subsets, 34K+ HDR images) for each tone-mapper used in ablation studies \cref{app:fig::hdrdb_hist_tm_gamma,app:fig::hdrdb_hist_tm_log2,app:fig::hdrdb_hist_tm_inverted,app:fig::hdrdb_hist_tm_mixed,app:fig::hdrdb_hist_tm_naturallog}.

\clearpage

\begin{figure*}[htbp]
    \includegraphics[width=\linewidth,keepaspectratio]{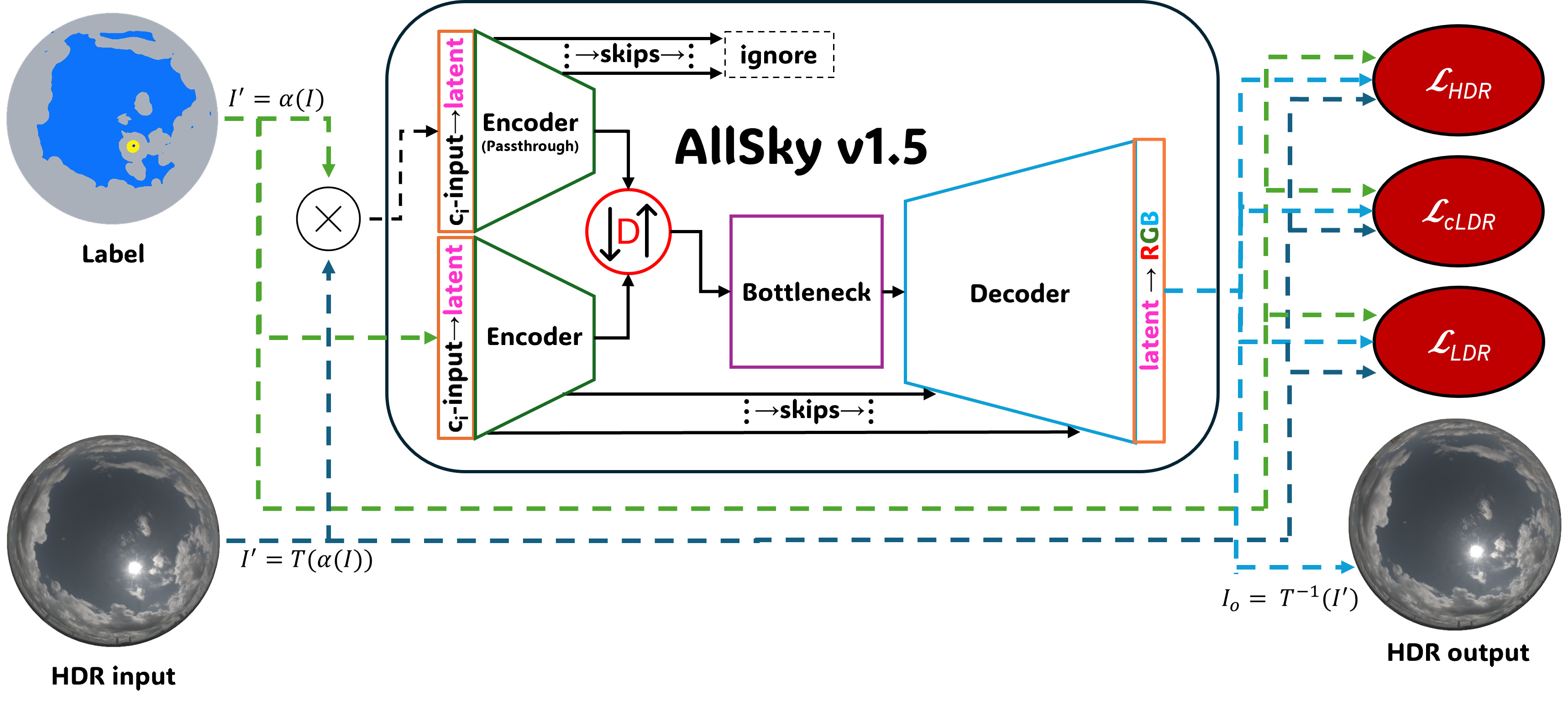}
    \caption{A second decoder can be incorporated into AllSky to enable a `passthrough' pre-training}
    \label{app:fig::AllSky_v1_5}
\end{figure*}

\section{AllSky}
\label{app:sec::AllSky}

\section{Tonemapping Operators}
\label{app:sec::AllSky:::tonemappers}

Tone-mapping operators ($T_m$) compress HDRI to a visible, displayable, or latent colour-space more favourable for DNN training given $I' = {T_{m}}(I)$.
We investigate a range of operators, including:
Power-Law ($T_\gamma$),
logarithmic ($T_{log_n}$),
$\mu$-law ($T_\mu$),
and variations thereof as shown in \cref{eq:tm_gamma,eq:tm_log_n,eq:tm_mu,eq:tm_muLawLog2_appendix} and \cref{fig:plt_tm} (left).
With the growing popularity of mixed-tone-mappers (combinations of tone-mappers) such as Hybrid Log-Gamma for display of 8K broadcasts \cite{HLG}, we propose $\mu$-lawLog$_2$ ($T_{\mu \log_2}$) to better preserve texture-rich cloud and skydome components (generally $EV \lessapprox 1$) and aggressively compress texture-poor high-exposure regions which are generally saturated in LDR imagery.

\begin{align}
\text{Power-Law: }& T_\gamma \left(I\right) = I^{\frac{1}{\gamma}}
    \label{eq:tm_gamma} \\
\text{Logarithmic: }& T_{log_n} \left(I\right) = \log_n(I+1) \label{eq:tm_log_n} \\
\mu\text{-law: }& T_\mu \left(I\right) = \frac{\log_e(1+\mu I)}{\log_e(1.0+\mu)} \label{eq:tm_mu} \\
\mu\text{-lawLog}_2 \text{: }& T_{\mu \log_2} \left(I\right) = \log_2\left[ \frac{\log_e(1+\mu I)}{\log_e(1.0+\mu)} +1\right] \label{eq:tm_muLawLog2_appendix}
\end{align}

\subsection{Alternate Model Architecture: Passthrough}
\label{app:sec::AllSky_passthrough}

Where explored an alternative `passthrough' routine where a generator ($G$) is first pre-trained to reconstruct ($I_{fake}$) the HDRI input ($I_{real}$) per $I_{fake}=G(I_{real})$, then trained with the desired input modality ($I_m$) per $I_{fake}=G(I_m)$.
In theory, this would enable the model to pre-learn photorealistic textures that are retained in training for new input modalities $I_m$.
Alternatively, this could also enable iterative erosion of label segmentation during training.

Unfortunately, AllSky's U-Net architecture as shown in \cref{fig:dia_AllSky} is incompatible with `passthrough' pre-training.
In experimentation, we found the U-Net learns to bypass the bottleneck via the skip connections and thus leads to no downstream training advantage.

AllSky can be modified to enable a `passthrough' pre-training by incorporating a second encoder.
Though this encoder's skip connections are ignored, the encoder provides the bottleneck with latents from HDRI input ($I_{real}$) interleaved into the scattered label by `X'.
Latent inputs to the bottleneck are modulated by \textcolor{red}{D}, which applies dropout and decay.
Once the `passthrough' pre-training is complete, the second decoder can be discarded.

\subsection{Training}
\label{app:sec::AllSky:::training}

We train ablation studies for 400 epochs on 4096 environment-maps at $128^2$ resolution from the experimental subset.
Training parameters are kept constant, with a learning rate of $1e^{-5}$ (generator and discriminator), although we note that at higher resolutions (e.g.\ $512^2$) trainings can benefit from increasing this learning rate to $1e^{-4}$.
Although training times varied, this configuration enabled training and evaluation in just under 24 hours on an NVIDIA Tesla V100 Volta 32 GB GPU.
For equal comparison, the training subset's environment-maps are consistent, and evaluation is completed on the entirety of our test subset.

\begin{sloppypar}
Though conventional supervised (e.g. $L_1$) and unsupervised (e.g. discriminator) training support arbitrary value ranges $\left\{x \subset\mathbb{R}\right\}$, our discriminator is dependent on LPIPS.
To accommodate, we linearly shift tone-mapped HDRI to the operational domain of LPIPS $\left\{x \subset \mathbb{R} | -1 \leq x \lessapprox 1 \right\}$.
Note, in data pre-processing we do not clip compressed LDR space to $\left\{x \subset \mathbb{R} | 0 \leq x \leq 1 \right\}$ (hence $\lessapprox$ above).
Clipping HDRI destroys the exposure range and illumination and, if applied post-generation, can result in values being unconstrained by losses (leading to NaNs and collapse of training).

Constraining the ranges $\left\{x \subset \mathbb{R} | -1 \leq x \leq t \right\}$ and $\left\{x \subset \mathbb{R} | t < x \right\}$ with seperate losses proved to be problematic.
Given stochasticity in illumination and arbitrary tone-mapper selection, content affected by the threshold $t$ could include the sun, cloud formations, the skydome, nothing or everything.
Further experimentation is required to explore the impact of value distributions on training DNNs for HDR, EDR and FDR.
\end{sloppypar}

\section{Supplemental Results}
\subsection{Tone-mapping Operator Ablation}
\label{app:sec::AllSky:::ablation_tone-mappers}

\begin{figure*}[htbp]
    \includegraphics[width=\linewidth,height=0.5\textheight,keepaspectratio]{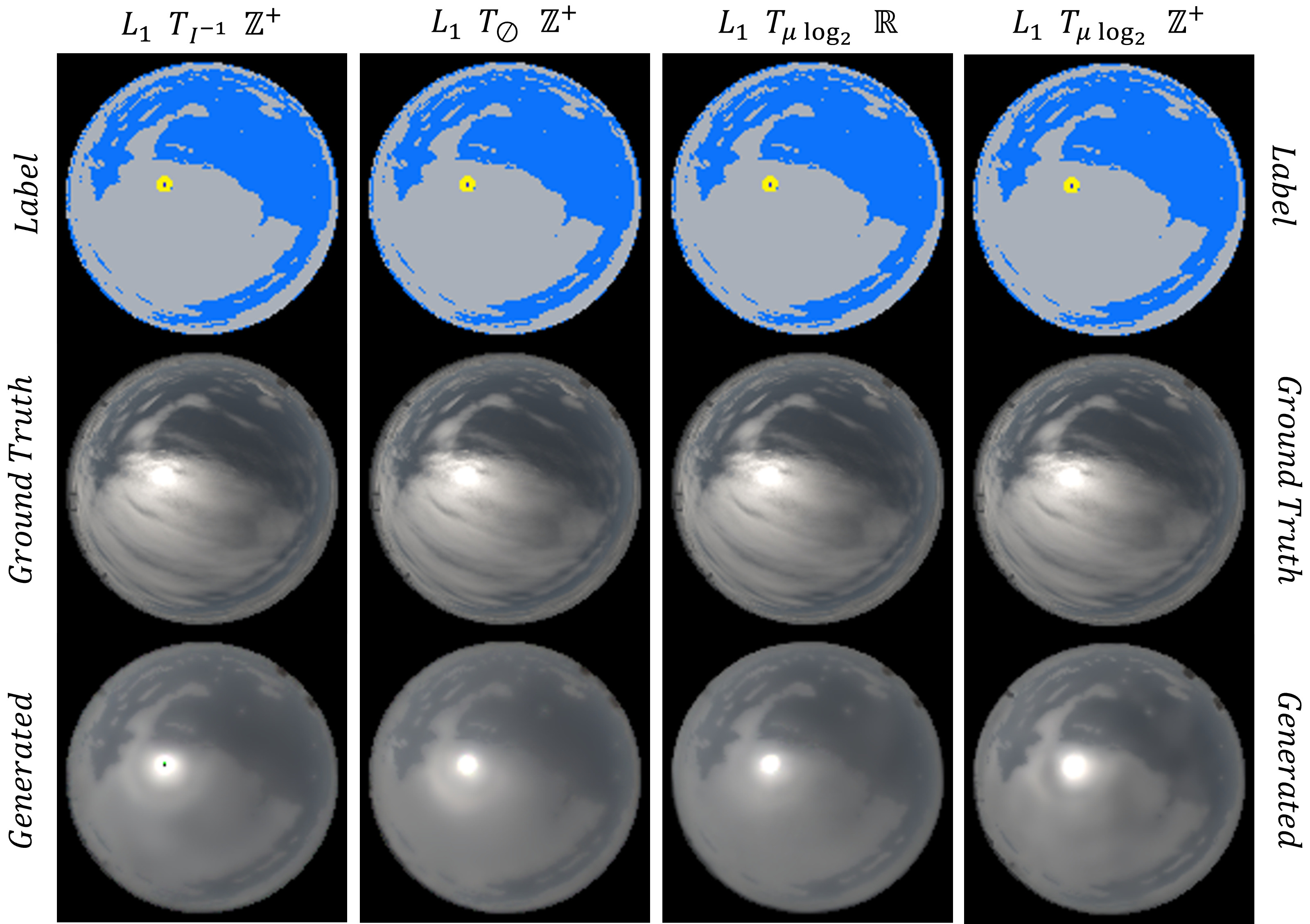}
    \caption{
    Visual results from training AllSky with an $L_1$-loss and varied tonemapping operators.
    An $L_1$-loss produces only smooth featureless cloud formations, but not all tonemapping enables coherent emulation of the sun.
    As shown, $T_{I^-1}$ produce `holes' in place of the solar disk.
    }
    \label{app:fig::grid_AllSky_L1}
\end{figure*}

\begin{figure*}[htbp]
    \includegraphics[width=\linewidth,height=0.4\textheight,keepaspectratio]{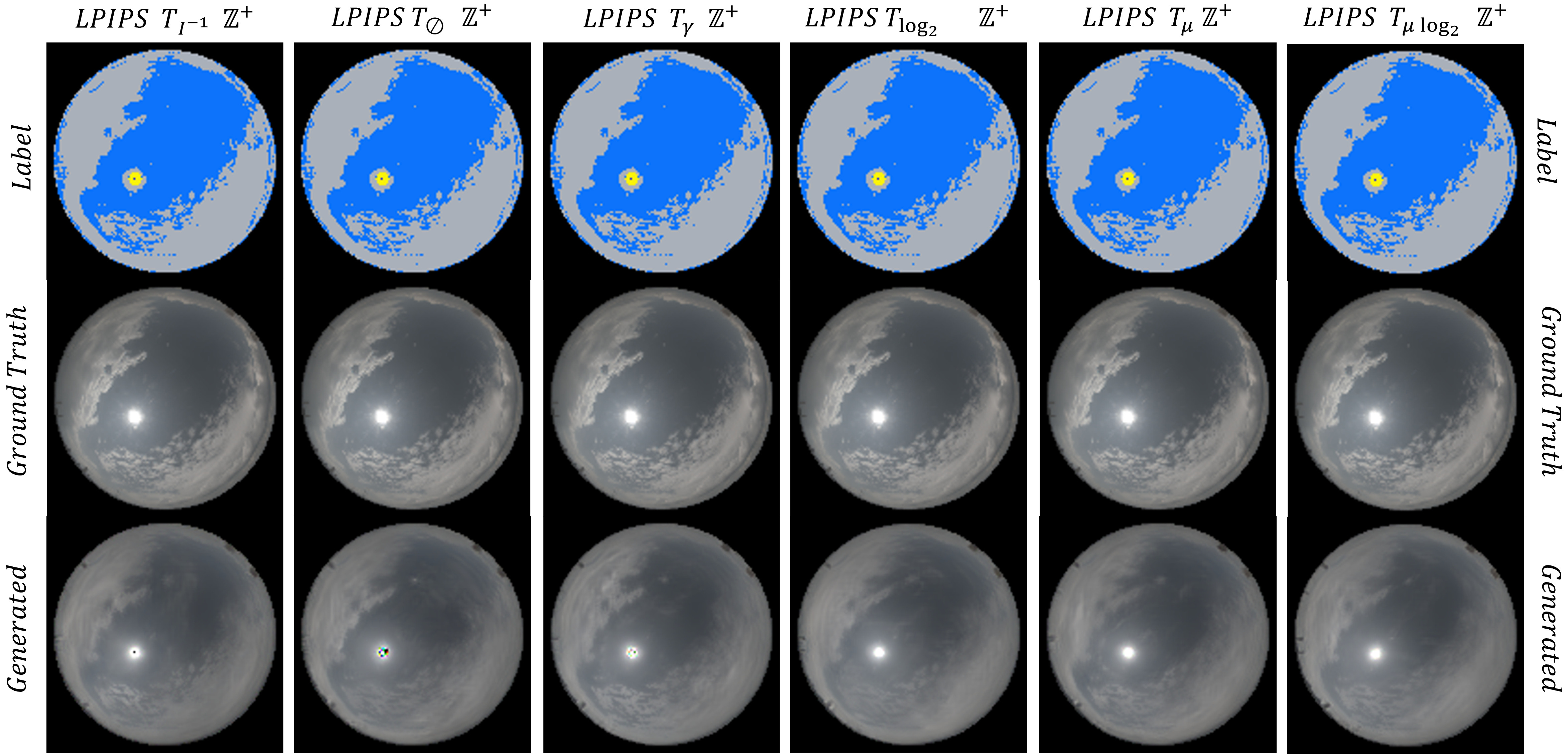}
    \caption{
    Visual results from training AllSky with an $LPIPS$-loss and varied tonemapping operators.
    LPIPS enables photorealistic weathered skies, but aggressive tonemapping is required to coherently emulate the sun.
    As shown, $T_\varnothing$, $T_{I^-1}$, and $T_{\gamma}$ produce `holes' in place of the solar disk.
    }
    \label{app:fig::grid_AllSky_LPIPS}
\end{figure*}

\begin{figure*}[htbp]
    \includegraphics[width=\linewidth,height=0.5\textheight,keepaspectratio]{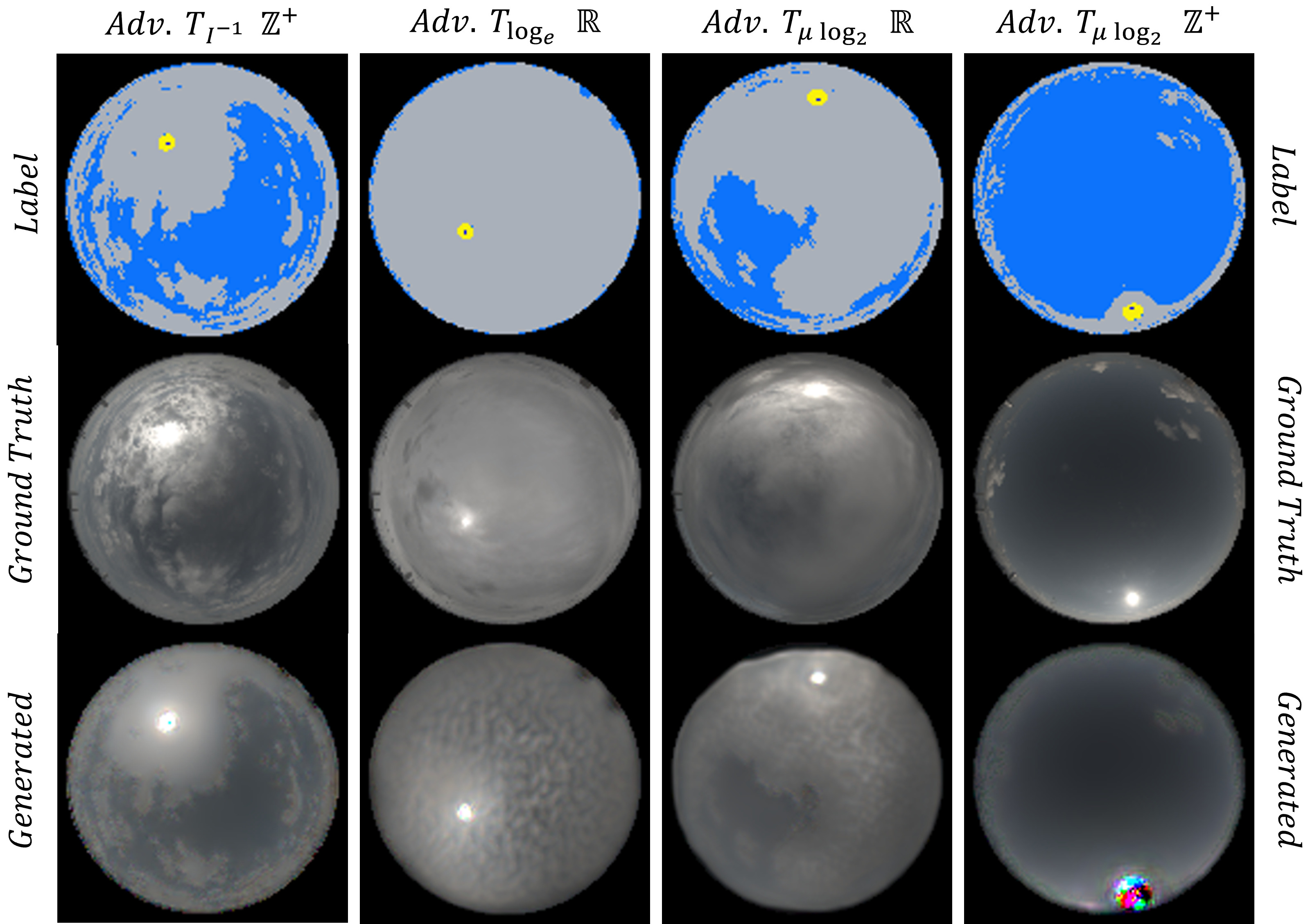}
    \caption{Visual results from training AllSky with an adversarial loss and varied tonemapping operators. Where discrete label ($\mathbb{Z^+}$) produce smooth textureless cloud formations, perlin noise injected into continuous labels ($\mathbb{R}$) appears carried forwards in generated environment maps. }
    \label{app:fig::grid_AllSky_adv}
\end{figure*}

We completed an ablation study to determine the impact of tone-mapping operator selection on model training and performance.
We ablate
no tonemapping ($T_\varnothing$),
Power-Law ($T_\gamma$),
logarithmic ($T_{log_n}$),
$\mu$-law ($T_\mu$),
$\mu$-lawLog$_2$ ($T_{\mu \log_2}$),
Inverted ($T_{I^{-1}}$),
and Natural Logarithmic (${T}_{\log_e}$) tonemapping.
We repeat this ablation for AllSky trained with discrete ($\mathbb{Z}^{+}$) and continuous ($\mathbb{R}$) labels for training with $L_1$, LPIPS and adversarial losses in \cref{tab:FixUpUnet_ablation_L1,tab:FixUpUnet_ablation_LPIPS,tab:FixUpUnet_ablation_DISC}.
Visuals from these ablations are included in \cref{app:fig::grid_AllSky_L1,app:fig::grid_AllSky_LPIPS,app:fig::grid_AllSky_adv}.

\begin{figure*}[htbp]
    \includegraphics[width=\linewidth,height=0.5\textheight,keepaspectratio]{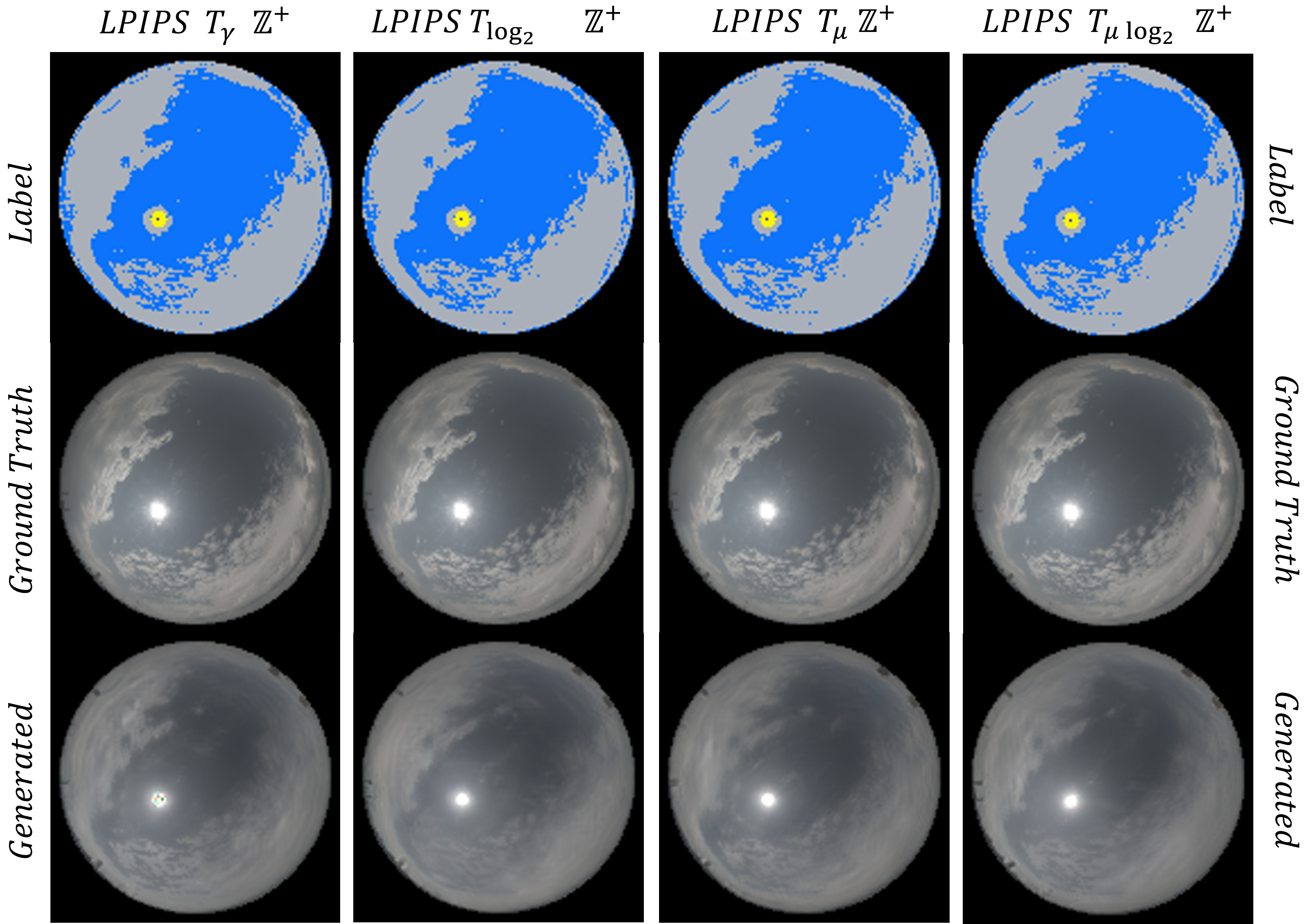}
    \caption{
    Visual results from training AllSky with an $LPIPS$-loss and varied tonemapping operators.
    LPIPS enables photorealistic weathered skies, but aggressive tonemapping is required to coherently emulate the sun.
    As shown, $T_{\gamma}$ produces `holes' in place of the solar disk.
    }
    \label{app:fig::grid_AllSky_LPIPS_256}
\end{figure*}

The results demonstrate that tonemappers can positively or negatively impact exposure range and illumination depending on loss selection.
With an $L_1$-loss, tonemapped non-linear space encourages restrained exposure range and illumination through lower high-intensity illumination error in comparison to linear-space ($T_\varnothing$).
This restraint is alleviated with an LPIPS-loss, allowing for stable illumination and for tonemapping to provide improved visual quality (LPIPS, FID).
Aggressive tonemapping is shown to improve visual quality, mitigating high-intensity pixels which are not sufficiently compressed for generation by the model without saturation or erroneous regions.
This is illustrated in \cref{app:fig::grid_AllSky_LPIPS,app:fig::grid_AllSky_LPIPS_256} through ‘holes’ appearing in place of the solar disk with $T_{I^-1}$, $T_{\varnothing}$ and $T_{\gamma}$.

The ablation further demonstrates that adversarial-losses exhibit poor illumination and visual quality, reflected by frequent numeric overflows in HDR reconstruction metrics.
Per overshooting of target illumination (EV and $\oiint_I$) and poor qualitative results (LPIPS, FID) illustrated in \cref{app:fig::grid_AllSky_adv}, we believe that the discriminator learns to differentiate inputs by honing in on the sun, resulting in saturation of the solar region in LDR-space and numeric overflow from exponentially-large error in HDR-space.
Exacerbated by the class imbalance, the sun provides a reliable means of differentiating real from generated environment maps even under tonemapped conditions.


Through the adversarial loss quantitatively favours continuous ($\mathbb{R}$) labels, visual quality (LPIPS, FID) and illumination (EV, $\oiint_I$) are poor in all circumstances.
As shown in \cref{app:fig::grid_AllSky_adv}, this improvement is dubious and likely the result of perlin noise propagating from the cloud masks of continuous labels.
With both $L_1$ and LPIPS losses, training favours discrete ($\mathbb{Z}^{+}$) labels, with favourable visual quality (LPIPS, FID) and illumination (EV, $\oiint_I$) produced with our proposed $\mu$-lawLog$_2$ ($T_{\mu \log_2}$) tonemapping operator.

\subsection{Supplemental Results: Cloud Mask Erosion}
\label{app:sec::AllSky:::ablation_cloudMaskErosion}

We perform an ablation study to determine the tractability provided by cloud masks by iteratively eroding the morphology of cloud segmentation.\footnote{
  Our crude (C) and hand-drawn (H) cloud masks are near equivalents to iterative erosion by $k=1$ and $k=15$.
  The morphology of crude cloud masks is unaltered (no erosion) prior to discretization.
  Hand-drawn cloud masks follow a manually-defined pipeline which alters morphology through multiple iterations of erosion and dilation prior to discretization.
}
As shown in \cref{tab:FixUpUnet_ablation_LPIPS_cloudMaskErosion}, illumination (EV, $\oiint_I$) is sustained, but visual performance (LPIPS, FID) is incrementally lost as cloud masks are eroded from crude (C) to hand-drawn (H).
We repeat the evaluation of each trained model (indicated by $\hookrightarrow$) with hand-drawn (H) labels to determine whether models trained with detailed cloud masks sustain performance with more-human-like eroded cloud masks post-training.
The results show a drop in visual performance (LPIPS, FID) proportional to the difference in kernel size between training and evaluation cloud mask erosion.
This shows that models trained for favourable visual performance with crude (C) cloud masks do not produce the same performance when used by a human operator.

\begin{figure*}[htbp]
    \centering
    \includegraphics[angle=90,height=0.90\textheight,width=\textwidth,keepaspectratio]{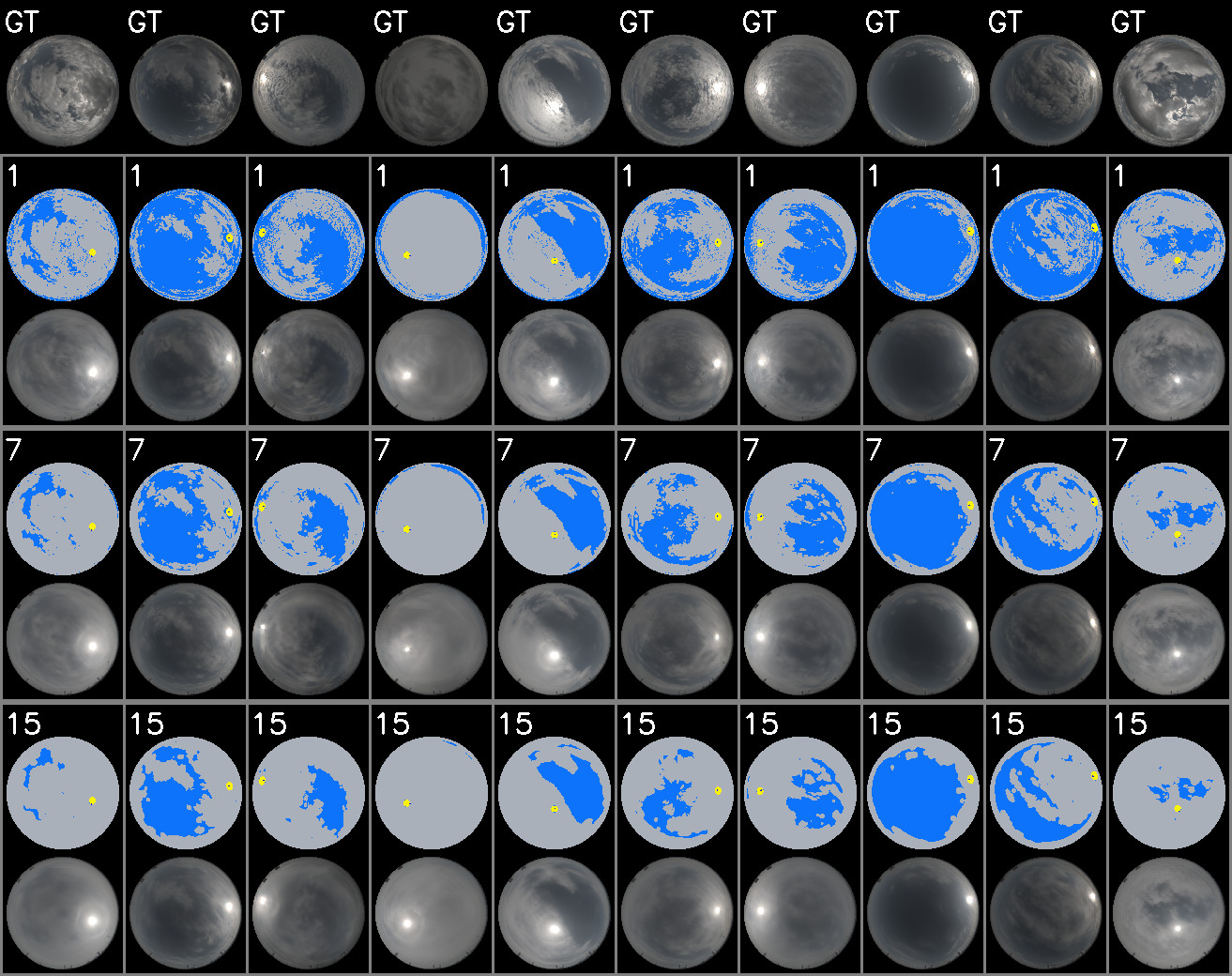}
    \caption{
        Environment-maps from training AllSky with $\mathbb{Z}^{+}$ labels, $\mu$-lawLog$_2$ ($T_{\mu \log_2}$) tone-mapping, and varied cloud-mask kernel size (1, 7, and 15).
        In comparison to the ground truth (GT), the clouds' of generated imagery progressively loose texture detail as the size of the cloud-mask erosion kernel increases.
        At a kernel size of 15, cloud textures are generalizable an amorphous gray.
    }
    \label{app:fig::AllSky_ablation_labels_128}
\end{figure*}

\begin{figure*}[htbp]
    \centering
    \includegraphics[height=0.9\textheight,width=\textwidth,keepaspectratio]{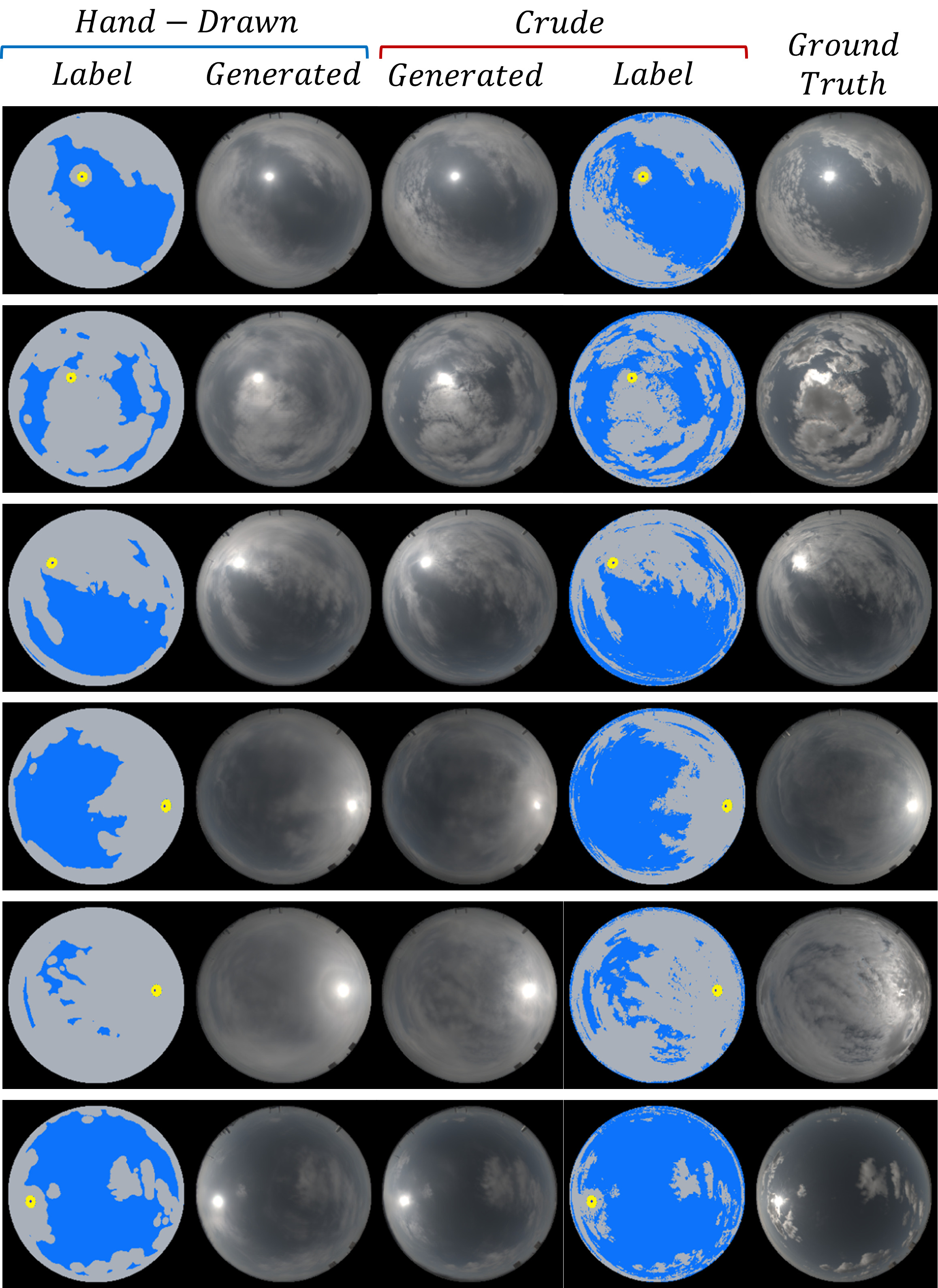}
    \caption{
        Environment-maps generated by AllSky with crude (C) clouds masks and hand-drawn (H) cloud masks.
        AllSky was trained with crude (C) $\mathbb{Z}^{+}$ labels and $\mu$-lawLog$_2$ ($T_{\mu \log_2}$) tone-mapping.
       Environment maps generated with hand-drawn (H) clouds mask exhibit significantly lower visual quality than those generated from crude clouds masks.
    }
    \label{app:fig::AllSky_ablation_labels_256}
\end{figure*}

We provide a visual comparison of select in \cref{app:fig::AllSky_ablation_labels_128,app:fig::AllSky_ablation_labels_256}
focusing on those with the lowest FID scores.
The results show that crude (highly detailed) labels produce better visual quality results (FID), which degrades quickly as kernel size increases.
Little variation in the exposure range (EV) or illumination ($\oiint_I$) is observed.

\begin{table*}[ht]
\centering
  \caption{
  $L_1$-loss ablation of AllSky for crude (C) discrete ($\mathbb{Z}^{+}$) and continuous ($\mathbb{R}$) labels and varied tone-mapping operators ($T_m$).
  The results demonstrate that tonemappers negatively impact the exposure range and illumination of AllSky trained with an $L_1$-loss.
  \textbf{Bold} and \textcolor{red}{red} values indicate performance limits (\textbf{best}/\textcolor{red}{worse}).
  }
  \label{tab:FixUpUnet_ablation_L1}
  \begin{tabular}{@{}r|cc|cc|cc|cc@{}}
    \multicolumn{1}{c}{ } & \multicolumn{2}{c|} {\textbf{LDR}}
    & \multicolumn{2}{c|} {\textbf{HDR}}
    & \multicolumn{2}{c|}{\textbf{$T_{\gamma}$-cLDR}}
    & \multicolumn{2}{c} {\textbf{HDR}} \\
    \cline{2-9}
    \textbf{$L_1$ Ablation} & $L_1$ $\downarrow$ & $L_2$ $\downarrow$ & $L_1$ $\downarrow$ & $L_2$ $\downarrow$ & LPIPS $\downarrow$ & FID $\downarrow$ & EV $\leftarrow$ & $\oiint_I \leftarrow$ \\
    \toprule
    Ground Truth $128^2$              & -  & -  & -  & -   & -     & -    & 5.88 & 1.26 \\
    \hdashline 
    \begin{tabular}{p{.3cm}p{1cm}p{0.3cm}} $\mathbb{Z}^{+}$ & $T_{\varnothing}$ & C \end{tabular}
    & \underline{0.19} & \underline{1964} & 0.09 & \underline{\textbf{491}} & \underline{0.18} & \underline{48} & \underline{6.78} & 0.96 \\
    \begin{tabular}{p{.3cm}p{1cm}p{0.3cm}} $\mathbb{R}$ & $T_{\varnothing}$ & C     \end{tabular}
    & \textcolor{red}{0.23} & \textcolor{red}{2227} & \textcolor{red}{0.11} & 557 & \textcolor{red}{0.21} & \textcolor{red}{58} & 7.44 & \underline{\textbf{1.04}} \\

    \hdashline 
    \begin{tabular}{p{.3cm}p{1cm}p{0.3cm}} $\mathbb{Z}^{+}$ & $T_{\gamma}$ & C \end{tabular}
    & \underline{0.06} & \underline{0.109} & 0.09 & 608 & \underline{0.17} & 45 & 4.51 & 0.71 \\
    \begin{tabular}{p{.3cm}p{1cm}p{0.3cm}} $\mathbb{R}$ & $T_{\gamma}$ & C     \end{tabular}
    & 0.07 & 0.111 & 0.09 & \underline{606} & 0.18 & \underline{40} & \underline{4.95} & \underline{0.74} \\

    \hdashline 
    \begin{tabular}{p{.3cm}p{1cm}p{0.3cm}} $\mathbb{Z}^{+}$ & $T_{\log_2}$ & C \end{tabular}
    & 0.06 & \underline{0.024} & 0.09 & 602 & \underline{0.17} & 39 & 5.01 & 0.75 \\
    \begin{tabular}{p{.3cm}p{1cm}p{0.3cm}} $\mathbb{R}$ & $T_{\log_2}$ & C     \end{tabular}
    & 0.06 & 0.026 & 0.09 & \underline{600} & 0.18 & \underline{36} & \underline{5.41} & \underline{0.76} \\

    \hdashline 
    \begin{tabular}{p{.3cm}p{1cm}p{0.3cm}} $\mathbb{Z}^{+}$ & $T_{\mu}$ & C \end{tabular}
    & 0.06  & \underline{0.022} & 0.09 & \underline{585} & 0.17 & 38 & 5.56 & 0.76 \\
    \begin{tabular}{p{.3cm}p{1cm}p{0.3cm}} $\mathbb{R}$ & $T_{\mu}$ & C     \end{tabular}
    & 0.06 & 0.026 & 0.09 & 591 & 0.17 & \underline{37} & \underline{\textbf{5.62}} & \underline{0.77} \\

    \hdashline 
    \begin{tabular}{p{.3cm}p{1cm}p{0.3cm}} $\mathbb{Z}^{+}$ & $T_{\mu\log_2}$ & C \end{tabular}
    & \underline{0.08} & \underline{0.023} & 0.09 & \underline{607} & \underline{0.17} & \underline{\textbf{35}} & \underline{4.55} & \underline{0.73} \\
    \begin{tabular}{p{.3cm}p{1cm}p{0.3cm}} $\mathbb{R}$ & $T_{\mu\log_2}$ & C     \end{tabular}
    & 0.09 & 0.028 & 0.09 & 610 & 0.18 & 44 & 3.97 & 0.71 \\

    \hdashline 
    \begin{tabular}{p{.3cm}p{1cm}p{0.3cm}} $\mathbb{Z}^{+}$ & $T_{\log_e}$ & C \end{tabular}
    & 0.05 & \underline{0.008} & 0.09 & \underline{609} & 0.18 & \underline{46} & \underline{4.33} & \underline{0.74} \\
    \begin{tabular}{p{.3cm}p{1cm}p{0.3cm}} $\mathbb{R}$ & $T_{\log_e}$ & C     \end{tabular}
    & 0.05 & 0.009 & 0.09 & 611 & 0.18 & 48 & \textcolor{red}{3.69} & \textcolor{red}{0.68} \\

    \hdashline 
    \begin{tabular}{p{.3cm}p{1cm}p{0.3cm}} $\mathbb{Z}^{+}$ & $T_{I^{-1}}$ & C \end{tabular}
    & \underline{\textbf{0.03}} & \underline{\textbf{0.005}} & 0.10 & \textcolor{red}{2454} & \underline{0.18} & \underline{43} & 3.95 & 0.73 \\
    \begin{tabular}{p{.3cm}p{1cm}p{0.3cm}} $\mathbb{R}$ & $T_{I^{-1}}$ & C     \end{tabular}
    & 0.04 & 0.006 & 0.10 & \underline{669} & 0.19 & 47 & \underline{4.42} & \underline{0.81} \\

    \bottomrule
\end{tabular}
\end{table*}

\begin{table*}[ht]
\centering
  \caption{
  LPIPS-loss ablation of AllSky for hand-drawn (H) and crude (C) discrete ($\mathbb{Z}^{+}$) and continuous ($\mathbb{R}$) labels and varied tone-mapping operators ($T_m$).
  The results demonstrate tonemapper selection directly impacts visual quality (LPIPS,FID), and an LPIPS-loss improves exposure range (EV) and illumination ($\oiint_I$) for all tone-mapping operators.
  Though illumination is retained when increasing resolution to $256^2$, visual quality is lost with Handrawn (H) labels.
  \textbf{Bold} and \textcolor{red}{red} values indicate performance limits (\textbf{best}/\textcolor{red}{worse}).
  }
  \label{tab:FixUpUnet_ablation_LPIPS}
  \begin{tabular}{@{}r|cc|cc|cc|cc@{}}
    \multicolumn{1}{c}{ } & \multicolumn{2}{c|} {\textbf{LDR}}
    & \multicolumn{2}{c|} {\textbf{HDR}}
    & \multicolumn{2}{c|}{\textbf{$T_{\gamma}$-cLDR}}
    & \multicolumn{2}{c} {\textbf{HDR}} \\
    \cline{2-9}
    \textbf{LPIPS Ablation} & $L_1$ $\downarrow$ & $L_2$ $\downarrow$ & $L_1$ $\downarrow$ & $L_2$ $\downarrow$ & LPIPS $\downarrow$ & FID $\downarrow$ & EV $\leftarrow$ & $\oiint_I \leftarrow$ \\
    \toprule

    Ground Truth $128^2$              & -  & -  & -  & -   & -     & -    & 5.88 & 1.26 \\
    \hdashline 
    \begin{tabular}{p{.3cm}p{1cm}p{0.3cm}} $\mathbb{Z}^{+}$ & $T_{\varnothing}$ & C \end{tabular}
    & \textcolor{red}{0.21} & \textcolor{red}{2401} & \underline{0.09} & 599 & \textcolor{red}{0.18} & \textcolor{red}{60} & \underline{5.84} & 0.76 \\
    \begin{tabular}{p{.3cm}p{1cm}p{0.3cm}} $\mathbb{R}$ & $T_{\varnothing}$ & C     \end{tabular}
    & \textcolor{red}{0.21} & \underline{2385} & 0.10 & \underline{596} & \textcolor{red}{0.18} & \underline{47} & 6.14 & \underline{0.78} \\

    \hdashline 
    \begin{tabular}{p{.3cm}p{1cm}p{0.3cm}} $\mathbb{Z}^{+}$ & $T_{\gamma}$ & C \end{tabular}
    & 0.07 & \underline{0.106} & 0.10 & \underline{570} & 0.16 & 20 & \underline{6.42} & \underline{0.89} \\
    \begin{tabular}{p{.3cm}p{1cm}p{0.3cm}} $\mathbb{R}$ & $T_{\gamma}$ & C     \end{tabular}
    & 0.07 & 0.109 & 0.10 & 575 & 0.16 & 20 & 6.60 & 0.87 \\

    \hdashline 
    \begin{tabular}{p{.3cm}p{1cm}p{0.3cm}} $\mathbb{Z}^{+}$ & $T_{\log_2}$ & C \end{tabular}
    & \underline{0.06} & 0.027 & 0.09 & 561 & 0.16 & \underline{18} & 6.32 & \underline{0.85} \\
    \begin{tabular}{p{.3cm}p{1cm}p{0.3cm}} $\mathbb{R}$ & $T_{\log_2}$ & C     \end{tabular}
    & 0.07 & \underline{0.026} & 0.09 & \underline{558} & 0.16 & 19 & \underline{6.04} & 0.84 \\

    \hdashline 
    \begin{tabular}{p{.3cm}p{1cm}p{0.3cm}} $\mathbb{Z}^{+}$ & $T_{\mu}$ & C \end{tabular}
    & 0.06 & \underline{0.025} & 0.09 & \underline{554} & \underline{\textbf{0.15}} & \underline{\textbf{17}} & \underline{6.26} & 0.84 \\
    \begin{tabular}{p{.3cm}p{1cm}p{0.3cm}} $\mathbb{R}$ & $T_{\mu}$ & C     \end{tabular}
    & 0.06 & 0.026 & 0.09 & 561 & 0.16 & 19 & 6.39 & \underline{\textbf{0.93}} \\

    \hdashline 
    \begin{tabular}{p{.3cm}p{1cm}p{0.3cm}} $\mathbb{Z}^{+}$ & $T_{\mu\log_2}$ & C \end{tabular}
    & \underline{0.09} & \underline{0.025} & 0.09 & \underline{552} & \underline{\textbf{0.15}} & \underline{\textbf{17}} & \underline{\textbf{5.88}} & 0.91 \\
    \begin{tabular}{p{.3cm}p{1cm}p{0.3cm}} $\mathbb{R}$ & $T_{\mu\log_2}$ & C     \end{tabular}
    & 0.10 & 0.030 & 0.09 & 564 & 0.16 & 18 & 5.97 & 0.91 \\

    \hdashline 
    \begin{tabular}{p{.3cm}p{1cm}p{0.3cm}} $\mathbb{Z}^{+}$ & $T_{\log_e}$ & C \end{tabular}
    & 0.05 & \underline{0.008} & 0.09 & 563 & \underline{\textbf{0.15}} & \underline{19} & \underline{5.92} & 0.84 \\
    \begin{tabular}{p{.3cm}p{1cm}p{0.3cm}} $\mathbb{R}$ & $T_{\log_e}$ & C     \end{tabular}
    & 0.05 & 0.010 & 0.09 & \underline{\textbf{524}} & 0.16 & 21 & 6.08 & \underline{0.87} \\

    \hdashline 
    \begin{tabular}{p{.3cm}p{1cm}p{0.3cm}} $\mathbb{Z}^{+}$ & $T_{I^{-1}}$ & C \end{tabular}
    & \underline{\textbf{0.03}} & \textbf{0.005} & \underline{0.09} & \underline{649} & \underline{0.16} & \underline{27} & \textcolor{red}{4.33} & \textcolor{red}{0.74} \\
    \begin{tabular}{p{.3cm}p{1cm}p{0.3cm}} $\mathbb{R}$ & $T_{I^{-1}}$ & C     \end{tabular}
    & 0.04 & \textbf{0.005} & \textcolor{red}{0.11} & \textcolor{red}{81084} & 0.17 & 34 & \underline{4.43} & \underline{0.82} \\
    \midrule
    Ground Truth $256^2$              & -  & -  & -  & -   & -     & -    & 8.54 & 1.21 \\
    \hdashline
    \begin{tabular}{p{.3cm}p{1cm}p{0.3cm}} $\mathbb{Z}^{+}$ & $T_{\gamma}$ & C     \end{tabular}
    & \underline{0.07} & \underline{0.052} & \underline{0.07} & 271 & \underline{0.17} & \underline{19} & \textcolor{red}{9.20} & 1.15 \\
    \begin{tabular}{p{.3cm}p{1cm}p{0.3cm}} $\mathbb{Z}^{+}$ & $T_{\gamma}$ & H     \end{tabular}
    & 0.08 & \textcolor{red}{0.056} & 0.08 & \underline{\textbf{258}} & 0.21 & 37 & \underline{9.11} & \underline{\textbf{1.18}}  \\

    \hdashline
    \begin{tabular}{p{.3cm}p{1cm}p{0.3cm}} $\mathbb{Z}^{+}$ & $T_{\log_2}$ & C     \end{tabular}
    & \underline{\textbf{0.06}} & \underline{0.020} & \underline{0.07} & \underline{284} & \underline{0.17} & \underline{19} & 9.00 & \underline{1.12} \\
    \begin{tabular}{p{.3cm}p{1cm}p{0.3cm}} $\mathbb{Z}^{+}$ & $T_{\log_2}$ & H     \end{tabular}
    & 0.07 & 0.023 & 0.08 & 331 & 0.20 & 36 & \underline{8.62} & \textcolor{red}{1.03} \\

    \hdashline
    \begin{tabular}{p{.3cm}p{1cm}p{0.3cm}} $\mathbb{Z}^{+}$ & $T_{\mu}$ & C     \end{tabular}
    & \underline{\textbf{0.06}} & \underline{\textbf{0.019}} & \underline{0.07} & 320 & \underline{0.17} & \underline{\textbf{18}} & \underline{8.39} & 0.97 \\
    \begin{tabular}{p{.3cm}p{1cm}p{0.3cm}} $\mathbb{Z}^{+}$ & $T_{\mu}$ & H     \end{tabular}
    & 0.07 & 0.023 & 0.08 & \underline{301} & 0.20 & \textcolor{red}{37} & 9.03 & \underline{1.14} \\

    \hdashline
    \begin{tabular}{p{.3cm}p{1cm}p{0.3cm}} $\mathbb{Z}^{+}$ & $T_{\mu\log_2}$ & C     \end{tabular}
    & \underline{0.08} & \underline{0.024} & \underline{0.08} & \underline{424} & \underline{0.17} & \underline{19} & \underline{\textbf{8.59}} & 1.09 \\
    \begin{tabular}{p{.3cm}p{1cm}p{0.3cm}} $\mathbb{Z}^{+}$ & $T_{\mu\log_2}$ & H     \end{tabular}
    & \textcolor{red}{0.10} & 0.031 & \textcolor{red}{0.09} & \textcolor{red}{676} & 0.20 & 33 & 9.03 & \underline{1.31} \\
 \bottomrule
\end{tabular}
\end{table*}

\begin{table*}[ht]
\centering
  \caption{
    Adversarial-loss ablation of AllSky for crude (C) discrete ($\mathbb{Z}^{+}$) and continuous ($\mathbb{R}$) labels and varied tone-mapping operators ($T_m$).
    The results demonstrate adversarial training favours discrete ($\mathbb{R}$) labels, but overall produces poor visual quality. This is reflected through frequent numerical overflow of HDR-space reconstruction metrics ($L_1$,$L_2$, EV, $\oiint_I$) from models saturating the solar corona and disc.
    As demonstrated by
    \textbf{Bold} and \textcolor{red}{red} values indicate performance limits (\textbf{best}/\textcolor{red}{worse})
  }
  \label{tab:FixUpUnet_ablation_DISC}
  \begin{tabular}{@{}r|cc|cc|cc|cc@{}}
      \multicolumn{1}{c}{ } & \multicolumn{2}{c|} {\textbf{LDR}}
    & \multicolumn{2}{c|} {\textbf{HDR}}
    & \multicolumn{2}{c|}{\textbf{$T_{\gamma}$-cLDR}}
    & \multicolumn{2}{c} {\textbf{HDR}} \\
    \cline{2-9}
    \textbf{Adversarial Ablation} & $L_1$ $\downarrow$ & $L_2$ $\downarrow$ & $L_1$ $\downarrow$ & $L_2$ $\downarrow$ & LPIPS $\downarrow$ & FID $\downarrow$ & EV $\leftarrow$ & $\oiint_I \leftarrow$ \\
    \toprule
    Ground Truth $128^2$              & -  & -  & -  & -   & -     & -    & 5.88 & 1.26 \\
    \hdashline 
    \begin{tabular}{p{.3cm}p{1cm}p{0.3cm}} $\mathbb{Z}^{+}$ & $T_{\varnothing}$ & C \end{tabular}
    & \underline{0.20} & \underline{2426} & \underline{\textbf{0.10}} & \underline{\textbf{606}} & \underline{0.22} & \underline{82} & \underline{\textbf{5.98}} & \underline{0.79} \\
    \begin{tabular}{p{.3cm}p{1cm}p{0.3cm}} $\mathbb{R}$ & $T_{\varnothing}$ & C     \end{tabular}
    & \textcolor{red}{8.85} & \textcolor{red}{361897} & 3.16 & 62017 & 0.31 & 118 & 14.76 & 43.23 \\

    \hdashline 
    \begin{tabular}{p{.3cm}p{1cm}p{0.3cm}} $\mathbb{Z}^{+}$ & $T_{\gamma}$ & C \end{tabular}
    & 8.64 & 192800 & 333367 & $2e^{15}$ & 0.33 & 151 & 31.27 & $2.14e^6$ \\
    \begin{tabular}{p{.3cm}p{1cm}p{0.3cm}} $\mathbb{R}$ & $T_{\gamma}$ & C     \end{tabular}
    & \underline{1.90} & \underline{5016} & \underline{546} & \underline{$4e^9$} & \underline{0.29} & \underline{91} & \underline{22.77} & \underline{9832.12} \\

    \hdashline 
    \begin{tabular}{p{.3cm}p{1cm}p{0.3cm}} $\mathbb{Z}^{+}$ & $T_{\log_2}$ & C \end{tabular}
    & 3.11 & 48534 & \textcolor{red}{$1.8e^{308}$} & \textcolor{red}{$1.8e^{308}$} & 0.31 & \underline{91} & 116.21 & $9.10e^{33}$ \\
    \begin{tabular}{p{.3cm}p{1cm}p{0.3cm}} $\mathbb{R}$ & $T_{\log_2}$ & C     \end{tabular}
    & \underline{0.58} & \underline{362} & \textcolor{red}{$1.8e^{308}$} & \textcolor{red}{$1.8e^{308}$} & 0.31 & \textcolor{red}{134} & \underline{100.62} & \underline{$8.6e^{32}$} \\

    \hdashline 
    \begin{tabular}{p{.3cm}p{1cm}p{0.3cm}} $\mathbb{Z}^{+}$ & $T_{\mu}$ & C \end{tabular}
    & 3.41 & 15045 & \textcolor{red}{$1.8e^{308}$} & \textcolor{red}{$1.8e^{308}$} & \underline{0.31} & 117 & 111.40 & $6.40e^{33}$ \\
    \begin{tabular}{p{.3cm}p{1cm}p{0.3cm}} $\mathbb{R}$ & $T_{\mu}$ & C     \end{tabular}
    & \underline{1.30} & \underline{8484} & \textcolor{red}{$1.8e^{308}$} & \textcolor{red}{$1.8e^{308}$} & 0.32 & \underline{106} & \underline{111.37} & \underline{$5.37e^{33}$} \\

    \hdashline 
    \begin{tabular}{p{.3cm}p{1cm}p{0.3cm}} $\mathbb{Z}^{+}$ & $T_{\mu\log_2}$ & C \end{tabular}
    & 4.79 & 95313 & $6.6e^{31}$ & \textcolor{red}{$1.8e^{308}$} & 0.30 & 132 & 112.48 & $4.38e^{32}$ \\
    \begin{tabular}{p{.3cm}p{1cm}p{0.3cm}} $\mathbb{R}$ & $T_{\mu\log_2}$ & C     \end{tabular}
    & \underline{0.10} & \underline{0.030} & \underline{\textbf{0.10}} & \underline{612} & \underline{\textbf{0.20}} & \underline{\textbf{49}} & \underline{2.94} & \underline{0.67} \\

    \hdashline 
    \begin{tabular}{p{.3cm}p{1cm}p{0.3cm}} $\mathbb{Z}^{+}$ & $T_{\log_e}$ & C \end{tabular}
    & 5.04 & 161257 & \textcolor{red}{$1.8e^{308}$} & \textcolor{red}{$1.8e^{308}$} & \textcolor{red}{0.37} & 129 & \textcolor{red}{121.68} & \textcolor{red}{$1.8e^{34}$} \\
    \begin{tabular}{p{.3cm}p{1cm}p{0.3cm}} $\mathbb{R}$ & $T_{\log_e}$ & C     \end{tabular}
    & \underline{0.06} & \underline{0.013} & \underline{\textbf{0.10}} & \underline{612} & \underline{0.21} & \underline{56} & \underline{3.24} & \underline{0.71} \\

    \hdashline 
    \begin{tabular}{p{.3cm}p{1cm}p{0.3cm}} $\mathbb{Z}^{+}$ & $T_{I^{-1}}$ & C \end{tabular}
    & \underline{\textbf{0.04}} & \underline{\textbf{0.006}} & \underline{\textbf{0.10}} & \underline{620} & \underline{0.21} & \underline{58} & 3.03 & 0.73 \\
    \begin{tabular}{p{.3cm}p{1cm}p{0.3cm}} $\mathbb{R}$ & $T_{I^{-1}}$ & C     \end{tabular}
    & 0.06 & 0.019 & 0.15 & 5867 & 0.32 & 130 & \underline{5.40} & \underline{\textbf{0.88}} \\

    \bottomrule
\end{tabular}
\end{table*}

\begin{table*}[ht]
\centering
  \caption{
  LPIPS-loss ablation of AllSky for discrete ($\mathbb{Z}^{+}$) labels with cloud masks eroded from `crude' ($k\text{=}1$) to `hand-drawn' ($k\text{=}15$).
  The results demonstrate sustained illumination but incrementally lost photorealism as cloud masks are eroded, demonstrating edge details is deterministic and provides models with a tractable means of conditioning the generation of cloud formations.
  We repeat the evaluation of each model (indicated by $\hookrightarrow$) with hand-drawn (H) labels, demonstrating further loss of visual performance in all instances.
  \textbf{Bold} and \textcolor{red}{red} values indicate performance limits (\textbf{best}/\textcolor{red}{worse}).
  }
  \label{tab:FixUpUnet_ablation_LPIPS_cloudMaskErosion}
  \begin{tabular}{@{}r|cc|cc|cc|cc@{}}
      \multicolumn{1}{c}{ } & \multicolumn{2}{c|} {\textbf{LDR}}
    & \multicolumn{2}{c|} {\textbf{HDR}}
    & \multicolumn{2}{c|}{\textbf{$T_{\gamma}$-cLDR}}
    & \multicolumn{2}{c} {\textbf{HDR}} \\
    \cline{2-9}
    \textbf{Cloud Mask Ablation} & $L_1$ $\downarrow$ & $L_2$ $\downarrow$ & $L_1$ $\downarrow$ & $L_2$ $\downarrow$ & LPIPS $\downarrow$ & FID $\downarrow$ & EV $\leftarrow$ & $\oiint_I \leftarrow$ \\
    \toprule

    Ground Truth $128^2$              & -  & -  & -  & -   & -     & -    & 5.88 & 1.26 \\
    \hdashline
    \begin{tabular}{p{.3cm}p{1cm}p{0.3cm}} $\mathbb{Z}^{+}$ & $T_{\mu\log_2}$ & C \end{tabular}
    & \textbf{0.086} & \textbf{0.024} & \textbf{0.088} & 542 & \textbf{0.15} & \textbf{16} & 5.72 & 0.86 \\

    \hdashline
    \begin{tabular}{p{.3cm}p{1cm}p{0.3cm}} $\mathbb{Z}^{+}$ & $T_{\mu\log_2}$ & 1     \end{tabular}
    & \underline{\textbf{0.086}} & \underline{\textbf{0.024}} & \underline{\textbf{0.088}} & \underline{\textbf{524}} & \underline{\textbf{0.15}} & \underline{\textbf{16}} & \underline{6.09} & \underline{0.90} \\
    \begin{tabular}{p{.3cm}p{1cm}p{0.3cm}}  & $\hookrightarrow$ & \textit{H}     \end{tabular}
    & 0.116 & 0.0366 & 0.094 & 552 & 0.20 & 33 & 5.21 & 0.80 \\

    \hdashline
    \begin{tabular}{p{.3cm}p{1cm}p{0.3cm}} $\mathbb{Z}^{+}$ & $T_{\mu\log_2}$ & 3 \end{tabular}
    & \underline{0.090} & \underline{0.026} & \underline{0.089} & 542 & \underline{\textbf{0.15}} & \underline{18} & \underline{5.81} & 0.87 \\
    \begin{tabular}{p{.3cm}p{1cm}p{0.3cm}}  & $\hookrightarrow$ & \textit{H}     \end{tabular}
    & 0.105 & 0.032 & 0.092 & \underline{529} & 0.18 & 24 & 5.51 & 0.87 \\

    \hdashline
    \begin{tabular}{p{.3cm}p{1cm}p{0.3cm}} $\mathbb{Z}^{+}$ & $T_{\mu\log_2}$ & 5     \end{tabular}
    & \underline{0.093} & \underline{0.027} & \underline{0.090} & \underline{555} & \underline{0.16} & \underline{18} & \underline{\textbf{5.86}} & \underline{0.88} \\
    \begin{tabular}{p{.3cm}p{1cm}p{0.3cm}}  & $\hookrightarrow$ & \textit{H}     \end{tabular}
    & 0.100 & 0.030 & 0.091 & 584 & 0.18 & 22 & 5.50 & 0.84 \\

    \hdashline
    \begin{tabular}{p{.3cm}p{1cm}p{0.3cm}} $\mathbb{Z}^{+}$ & $T_{\mu\log_2}$ & 7 \end{tabular}
    & \underline{0.094} & \underline{0.028} & 0.091 & 561 & \underline{0.17} & \underline{19} & \underline{\textcolor{red}{5.40}} & \underline{\textcolor{red}{0.83}} \\
    \begin{tabular}{p{.3cm}p{1cm}p{0.3cm}}  & $\hookrightarrow$ & \textit{H}     \end{tabular}
    & 0.098 & 0.030 & 0.091 & 561 & 0.18 & 20 & 5.18 & 0.81 \\

    \hdashline
    \begin{tabular}{p{.3cm}p{1cm}p{0.3cm}} $\mathbb{Z}^{+}$ & $T_{\mu\log_2}$ & 9     \end{tabular}
    & 0.101 & 0.032 & 0.092 & 561 & \underline{0.17} & \underline{20} & 5.57 & \textcolor{red}{0.83} \\
    \begin{tabular}{p{.3cm}p{1cm}p{0.3cm}}  & $\hookrightarrow$ & \textit{H}     \end{tabular}
    & 0.101 & 0.032 & \underline{0.091} & \underline{555} & 0.18 & 21 & \underline{5.62} & \underline{0.84} \\

    \hdashline
    \begin{tabular}{p{.3cm}p{1cm}p{0.3cm}} $\mathbb{Z}^{+}$ & $T_{\mu\log_2}$ & 11 \end{tabular}
    & 0.103 & 0.033 & 0.092 & \textcolor{red}{558} & 0.17 & \underline{21} & 5.47 & \underline{0.85} \\
    \begin{tabular}{p{.3cm}p{1cm}p{0.3cm}}  & $\hookrightarrow$ & \textit{H}     \end{tabular}
    & \underline{0.100} & \underline{0.032} & \underline{0.091} & \underline{557} & 0.17 & 22 & \underline{5.50} & 0.84 \\

    \hdashline
    \begin{tabular}{p{.3cm}p{1cm}p{0.3cm}} $\mathbb{Z}^{+}$ & $T_{\mu\log_2}$ & 15     \end{tabular}
    & \textcolor{red}{0.105} & \textcolor{red}{0.034} & \textcolor{red}{0.093} & 544 & \textcolor{red}{0.18} & \textcolor{red}{22} & 6.08 & \textbf{0.92} \\

    \hdashline
    \begin{tabular}{p{.3cm}p{1cm}p{0.3cm}} $\mathbb{Z}^{+}$ & $T_{\mu\log_2}$ & H \end{tabular}
    & 0.101 & 0.032 & 0.092 & 554 & \textcolor{red}{0.18} & 21 & \textbf{5.86} & 0.86 \\

    \midrule

    Ground Truth $256^2$              & -  & -  & -  & -   & -     & -    & 8.54 & 1.21 \\
    \hdashline
    \begin{tabular}{p{.3cm}p{1cm}p{0.3cm}} $\mathbb{Z}^{+}$ & $T_{\mu\log_2}$ & C     \end{tabular}
    & 0.08 & 0.024 & 0.08 & 424 & 0.17 & 19 & 8.59 & 1.09 \\
    \begin{tabular}{p{.3cm}p{1cm}p{0.3cm}}  & $\hookrightarrow$ & \textit{H}     \end{tabular}
    & 0.12 & 0.040 & 0.089 & 382 & 0.22 & 39 & 8.86 & 1.18 \\

    \hdashline
    \begin{tabular}{p{.3cm}p{1cm}p{0.3cm}} $\mathbb{Z}^{+}$ & $T_{\mu\log_2}$ & H     \end{tabular}
    & 0.10 & 0.031 & 0.089 & 676 & 0.20 & 33 & 9.03 & 1.31 \\
 \bottomrule
\end{tabular}
\end{table*}

\begin{figure*}[htbp]
    \centering
    \includegraphics[angle=90,height=0.95\textheight,width=\textwidth,keepaspectratio]{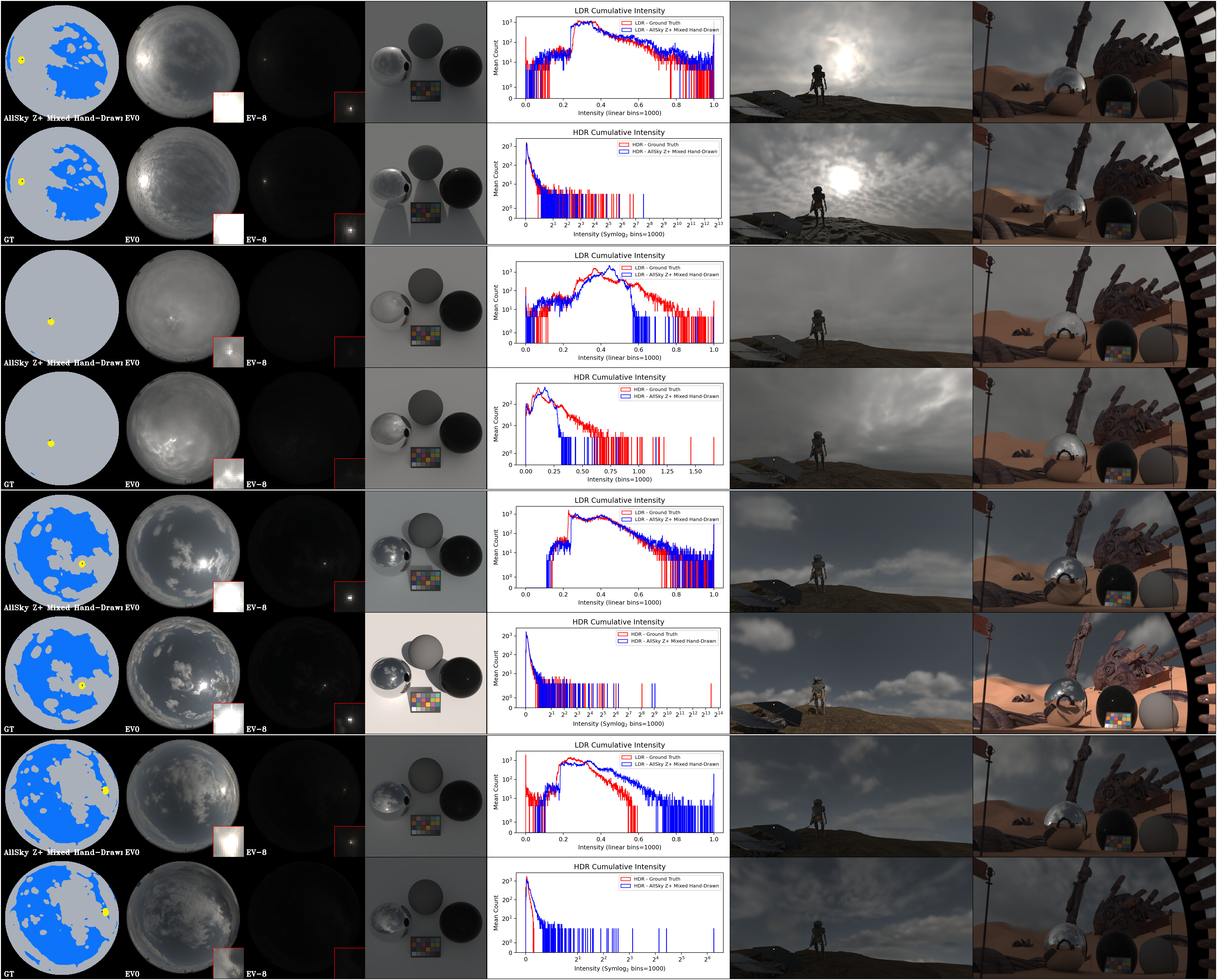}
    \caption{
        $256^2$ Environment-maps generated by AllSky with hand-drawn (H) cloud masks.
    }
    \label{app:fig::AllSky_256_handdrawn}
\end{figure*}
\begin{figure*}[htbp]
    \centering
    \includegraphics[angle=90,height=0.95\textheight,width=\textwidth,keepaspectratio]{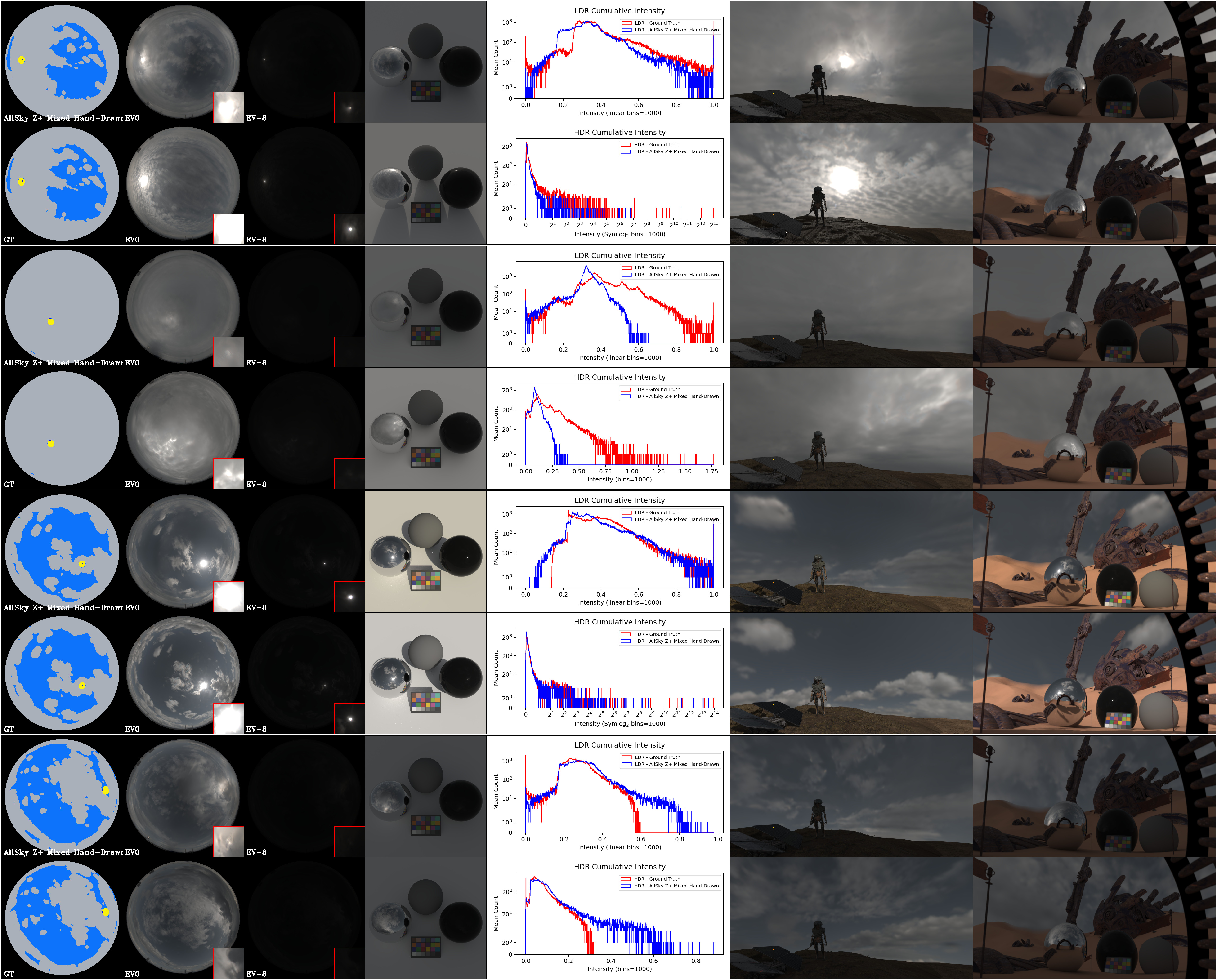}
    \caption{
        $512^2$ Environment-maps generated by AllSky with hand-drawn (H) cloud masks.
    }
    \label{app:fig::AllSky_512_handdrawn}
\end{figure*}
\begin{figure*}[htbp]
    \centering
    \includegraphics[angle=90,height=0.95\textheight,width=\textwidth,keepaspectratio]{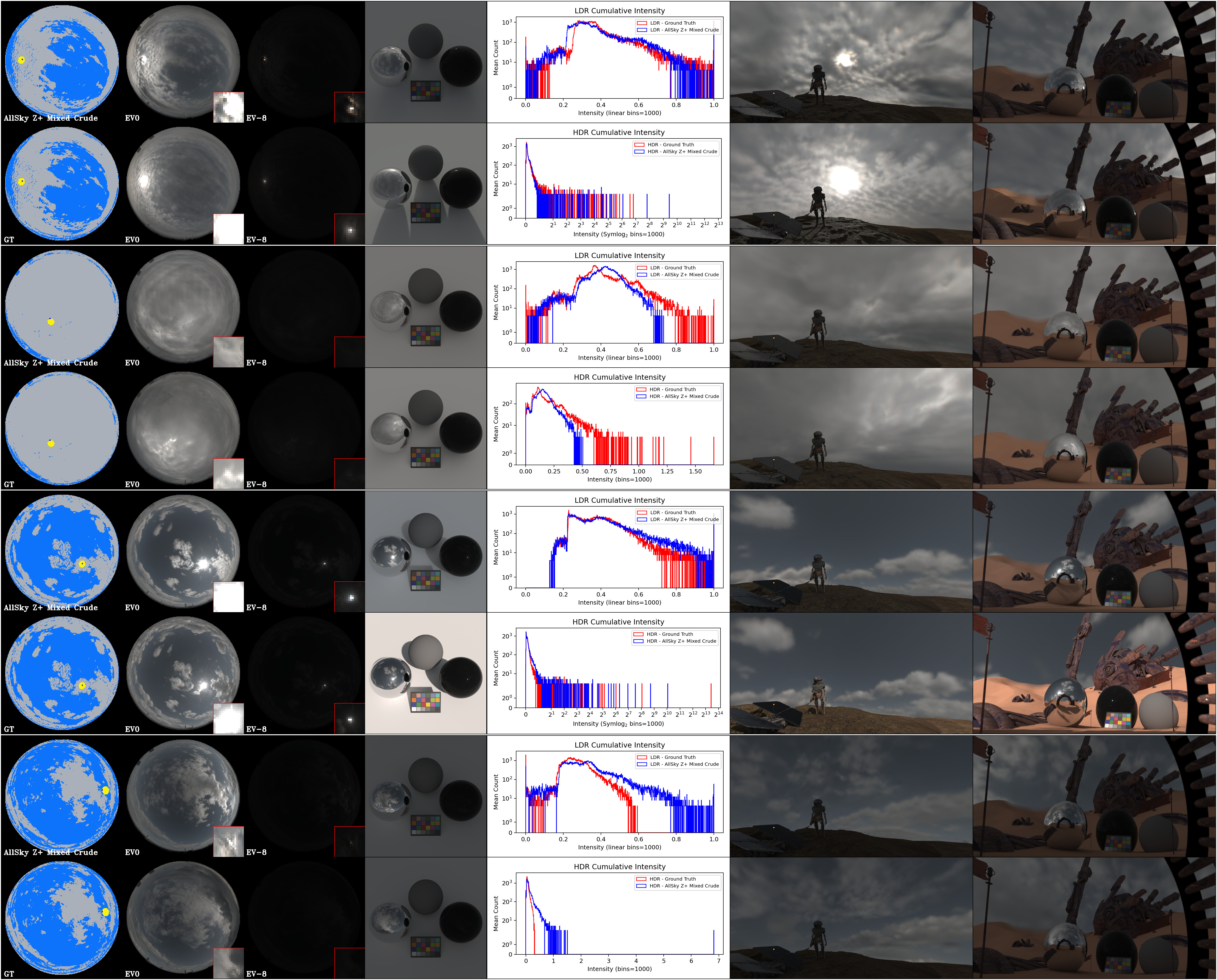}
    \caption{
        $256^2$ Environment-maps generated by AllSky with Crude (C) cloud masks.
    }
    \label{app:fig::AllSky_256_crude}
\end{figure*}
\begin{figure*}[htbp]
    \centering
    \includegraphics[angle=90,height=0.95\textheight,width=\textwidth,keepaspectratio]{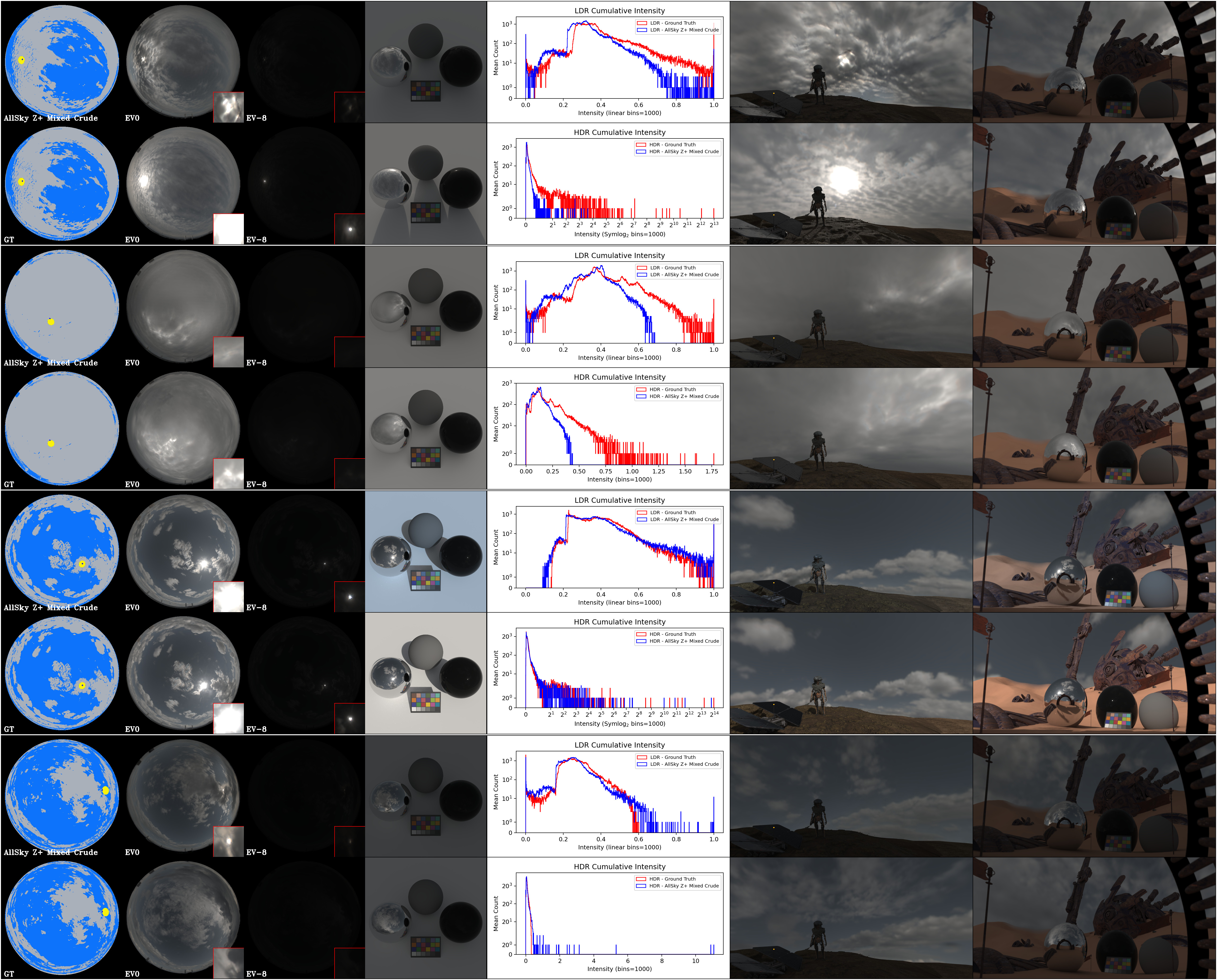}
    \caption{
        $512^2$ Environment-maps generated by AllSky with Crude (C) cloud masks.
    }
    \label{app:fig::AllSky_512_crude}
\end{figure*}

\clearpage

\section{CloudNet}
\label{app:sec::CloudNet}

\begin{figure*}[ht]
    \includegraphics[width=\textwidth,keepaspectratio]{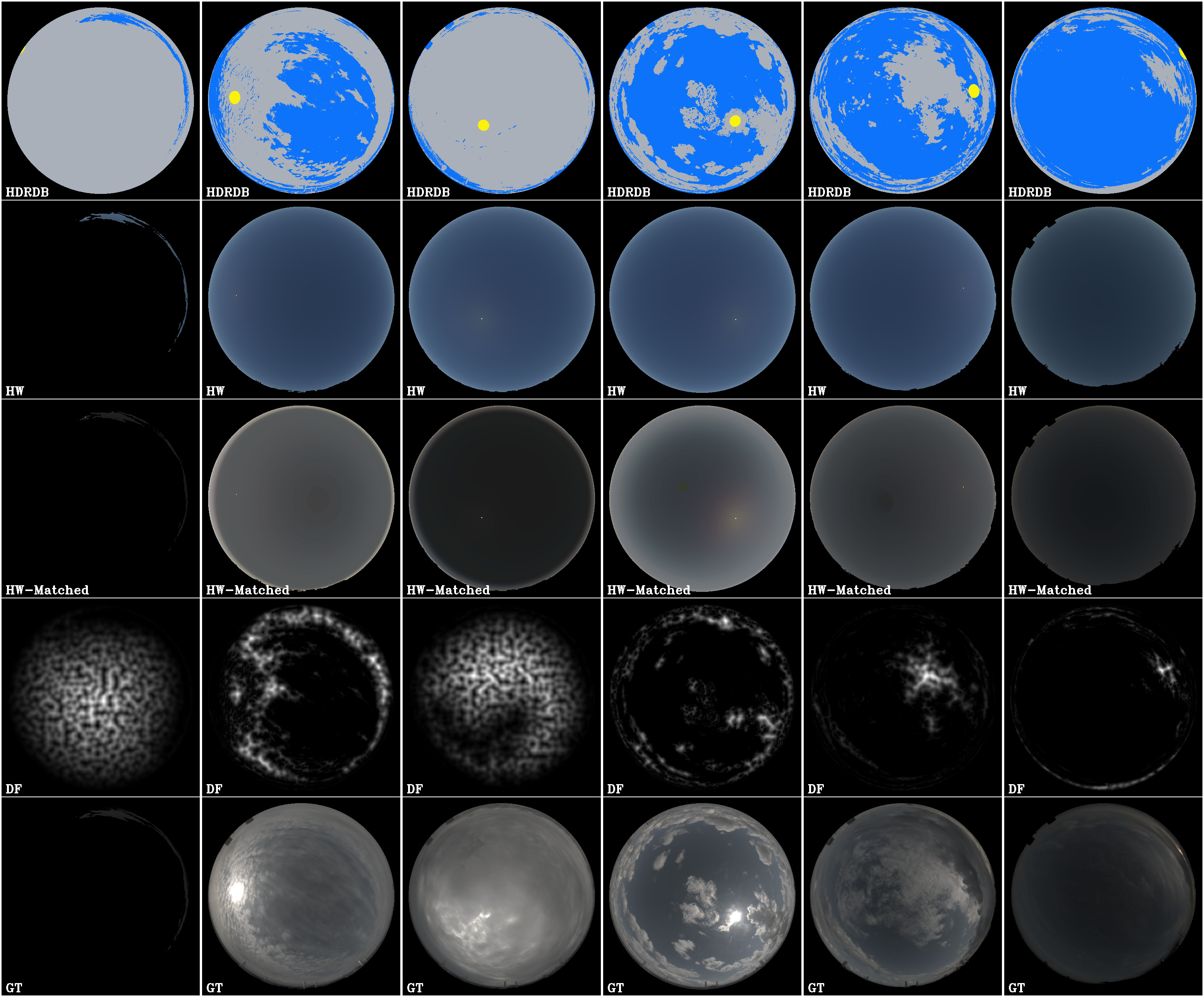}
    \caption{
    CloudNet clear sky priori and our segmentation of HDRDB with a uniform sun label (HDRDB).
    CloudNet' labels consists of a histogram match Ho\v{s}ek-Wilkie clear-skies (HW-Matched) and cloud mask of distance-field modulated perlin noise (DF).
    HW-Matched is created from Ho\v{s}ek-Wilkie clear-skies (HW) matched to ground truth (GT) to HDRI.
    }
    \label{app:fig::CloudNet_labels}
\end{figure*}

\begin{figure*}[htbp]
    \centering
    \includegraphics[width=\textwidth,height=0.95\textheight,keepaspectratio]{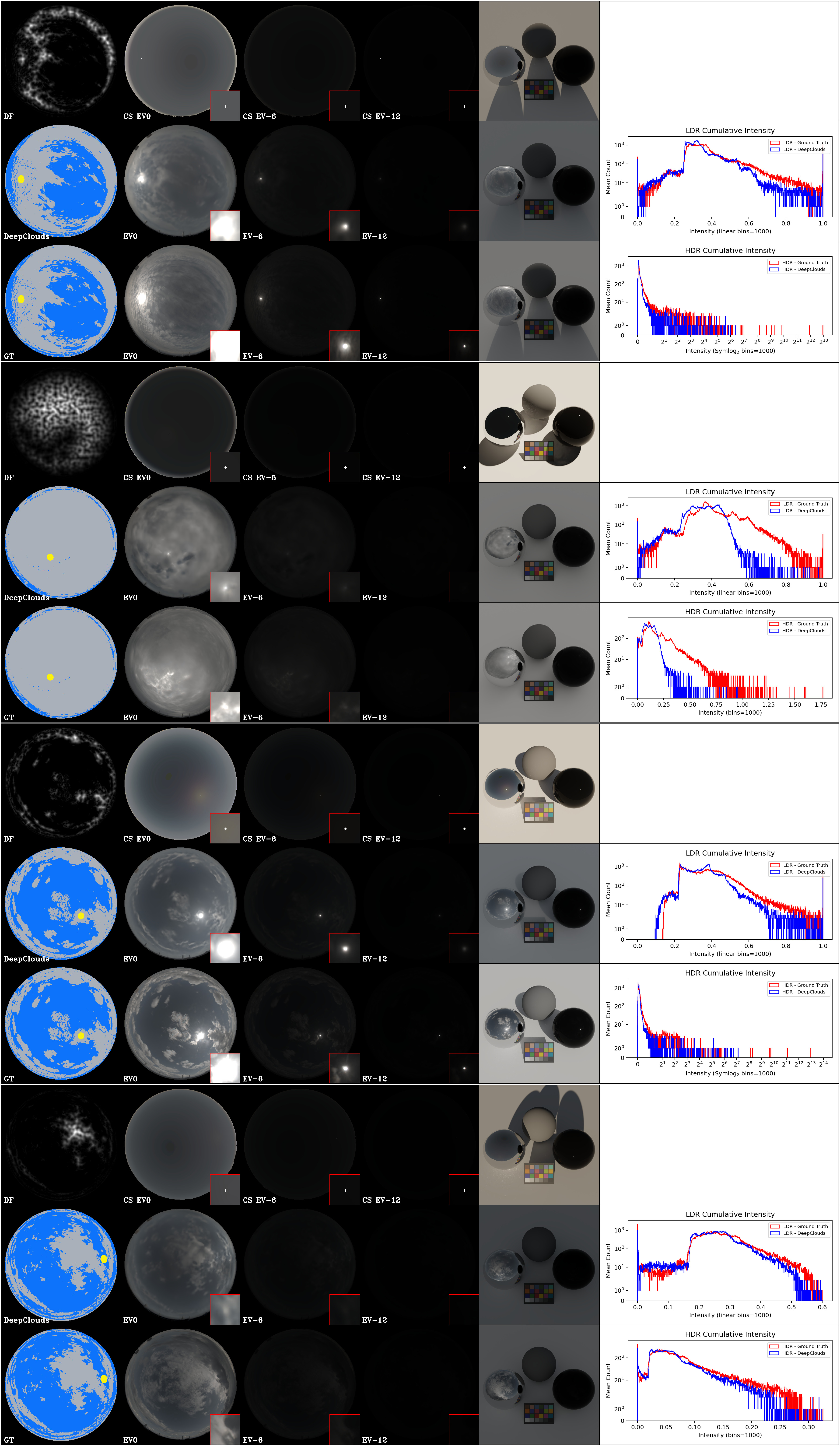}
    \caption{Visual results from CloudNet w/ clear sky priori w/o passthrough parametric sun}
    \label{app:fig::CloudNet}
\end{figure*}

\begin{figure*}[htbp]
    \centering
    \includegraphics[width=\textwidth,height=0.95\textheight,keepaspectratio]{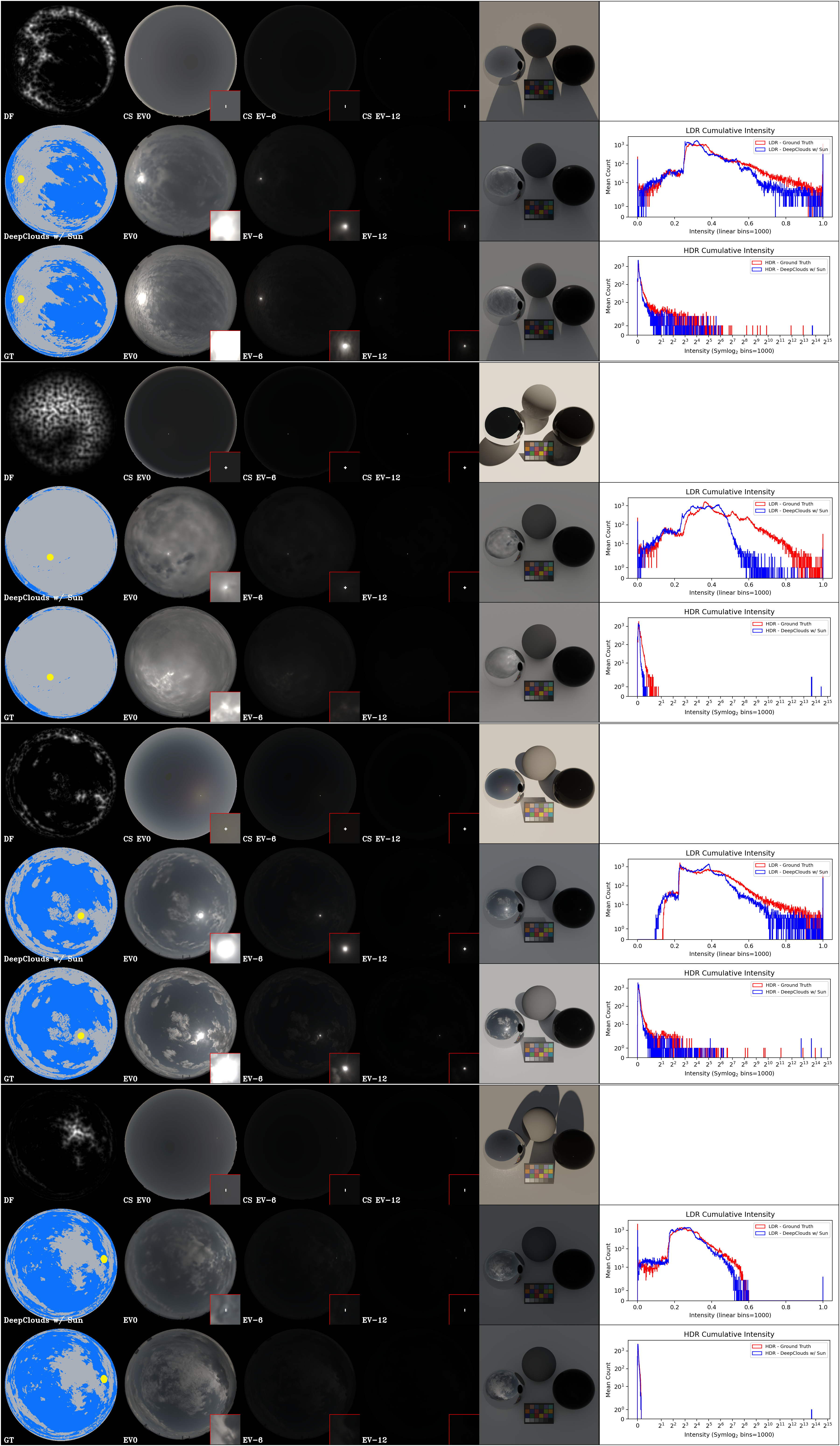}
    \caption{Visual results from CloudNet w/ clear sky priori w/ passthrough parametric sun w/o passthrough parametric sun}
    \label{app:fig::CloudNet_wSun}
\end{figure*}

\begin{figure*}[htbp]
    \centering
    \includegraphics[width=\textwidth,height=0.95\textheight,keepaspectratio]{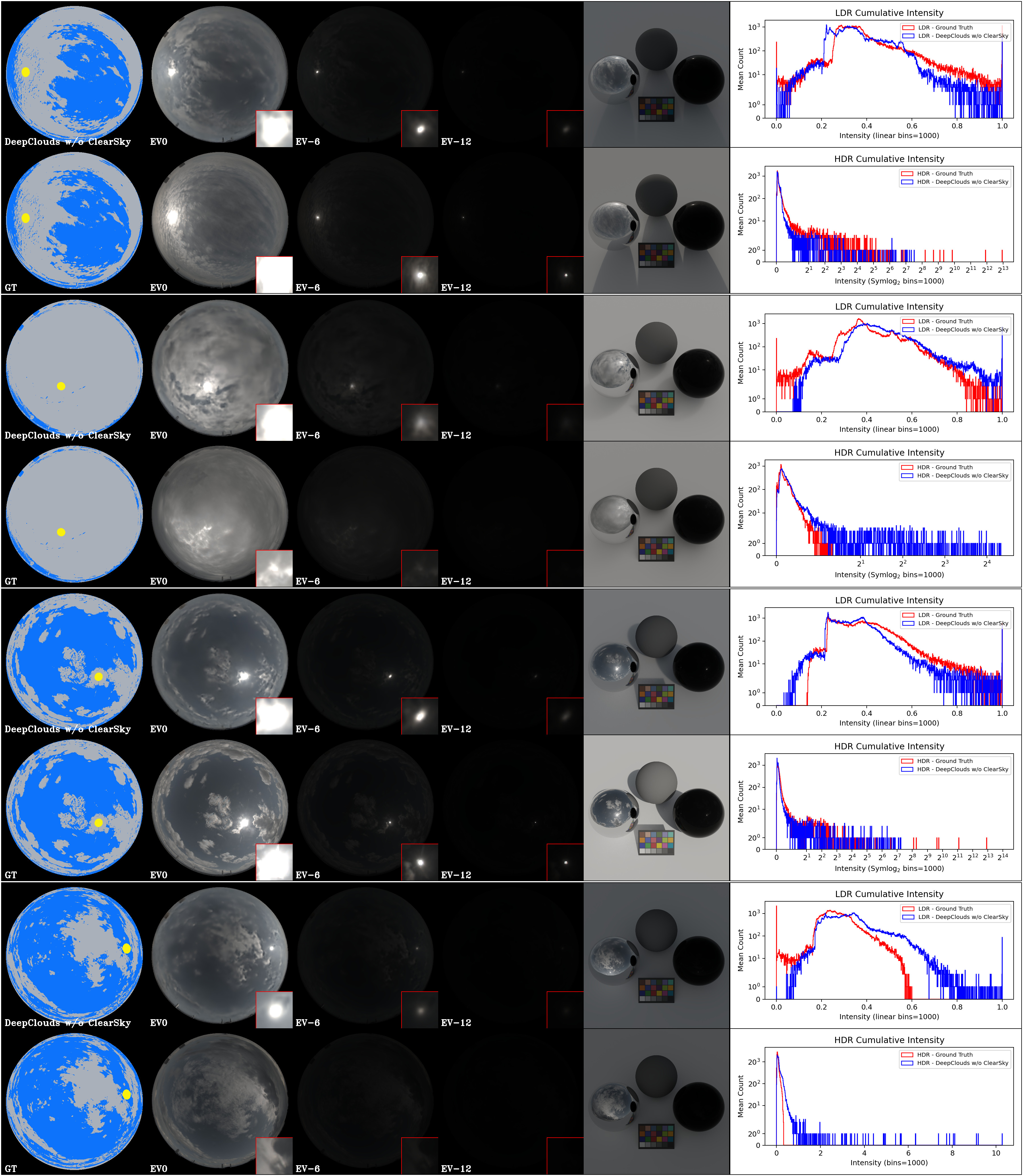}
    \caption{Visual results from CloudNet w/o clear sky priori}
    \label{app:fig::CloudNet_woClearSky}
\end{figure*}

\begin{figure*}[htbp]
    \centering
    \includegraphics[width=\textwidth,height=0.95\textheight,keepaspectratio]{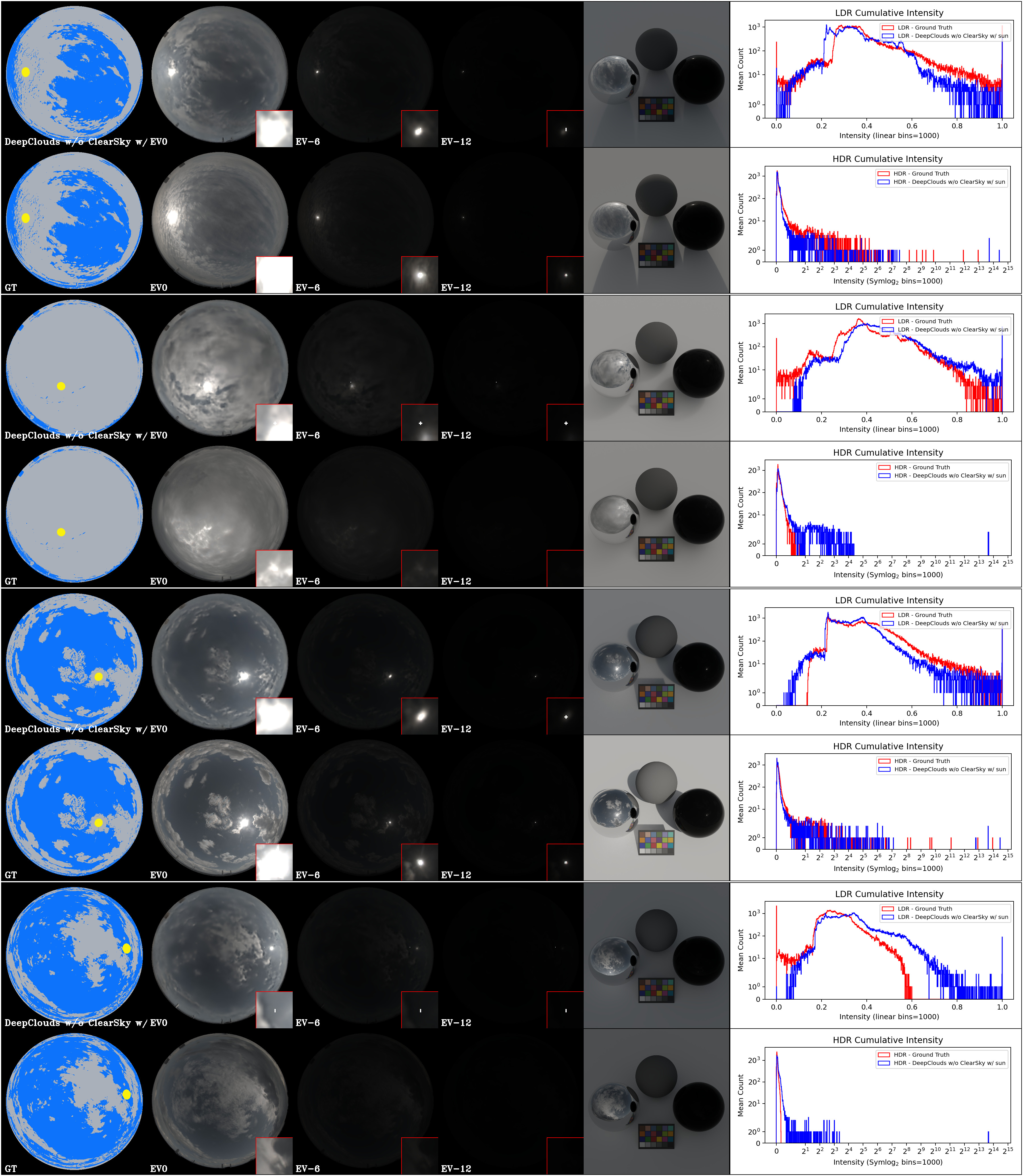}
    \caption{Visual results from CloudNet w/o clear sky priori w/ passthrough parametric sun}
    \label{app:fig::CloudNet_woClearSky_wSun}
\end{figure*}

We train CloudNet per the authors' specification, adapting our HDRDB dataloader to provide the specified cloud masks (similar to our `crude' cloud masks).
As illustrated in \cref{app:fig::CloudNet_labels}, we re-create the Ho\v{s}ek-Wilkie clear-skies (HW) and distance field (DF) priori per our segmentation of HDRDB.
The authors' source code was used create the histogram matched clear-skies (HW-matched) from Ho\v{s}ek-Wilkie clear-skies (HW) and ground truth (GT) images.
Illustrations for each variant of CloudNet are provided in \cref{app:fig::CloudNet,app:fig::CloudNet_wSun,app:fig::CloudNet_woClearSky,app:fig::CloudNet_woClearSky_wSun}.
CloudNet with labels (w/labels) was trained per our segmentation of HDRDB with a uniform sun label.

\subsection{Evaluation}
In \cref{tab:quantitative_results}, increased $L_1$, $L_2$, and $\oiint_I$ is the result of minor solar misalignment and the sun-pass-through mechanism, where $I[|S|>10]' = I[|S|>10] + S[|S|>10]$ for clear sky image $S$.
As a result, \textit{CloudNet}'s generated $4.5EV$ sun is summed with the Ho\v{s}ek-Wilkie sun to produce a cumulative $14EV$.

We further note the Ho\v{s}ek-Wilkie clear-skies input (HW) includes the parametric sun, but it is subsequently not reproduced by CloudNet and thus optionally re-added post-generation.
This is the result of the model learning to ignore the parametric sun as it is not present in ground truth imagery and therefore inconsequential to the training loss.

\clearpage

\begin{figure*}[ht]
    \centering
    \includegraphics[width=\textwidth,height=0.95\textheight,keepaspectratio]{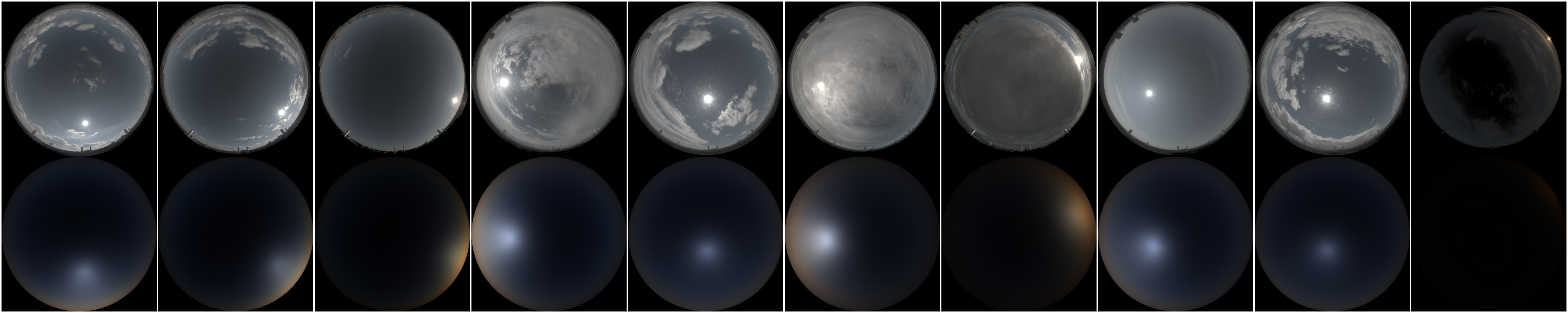}
    \caption{SkyGAN PSM clear sky priori}
    \label{app:fig::SkyGAN_PSM_clearSkies}
\end{figure*}

\section{SkyGAN}
\label{app:sec::SkyGAN}

\begin{figure*}[ht]
    \centering
    \includegraphics[width=\textwidth,height=0.95\textheight,keepaspectratio]{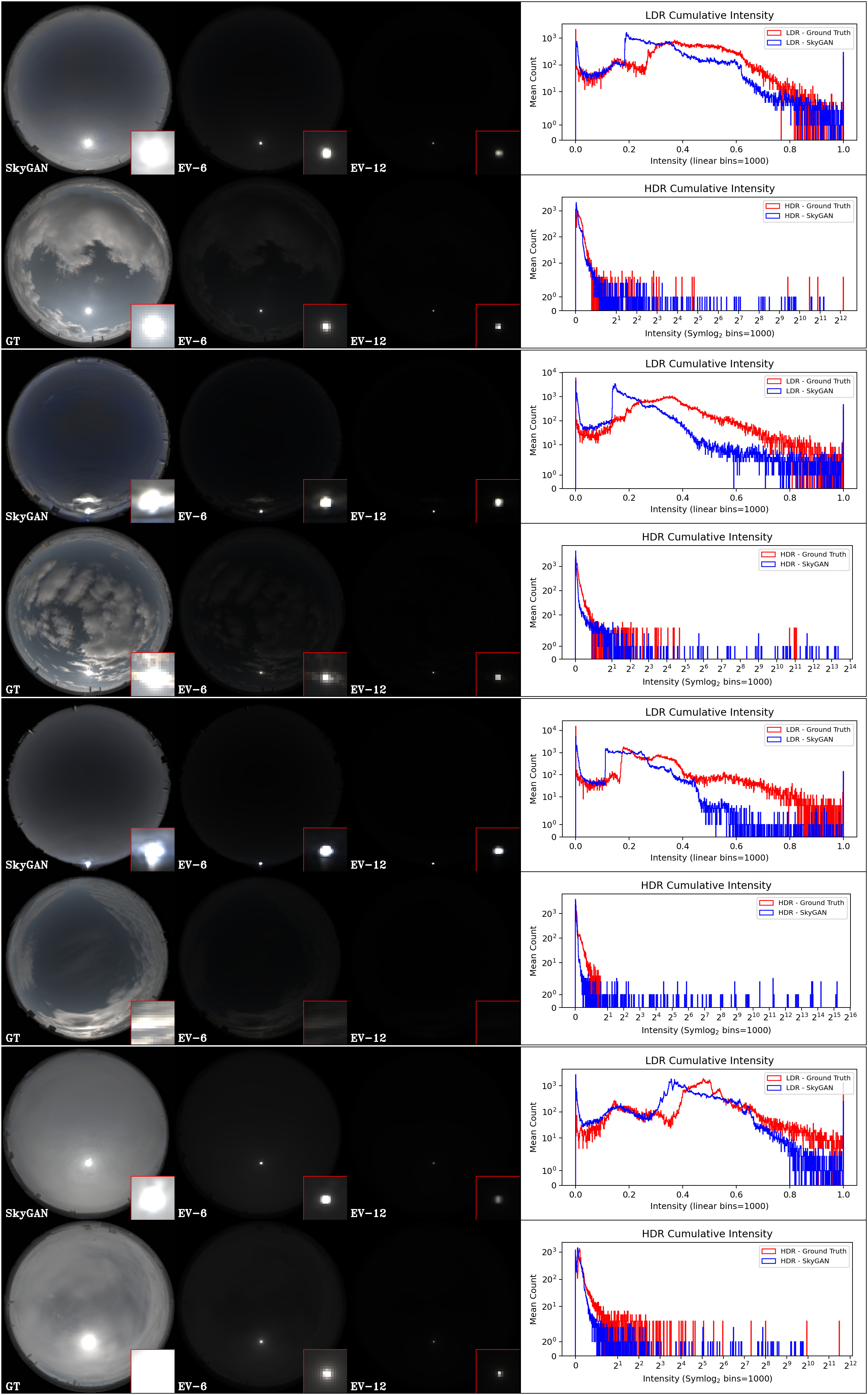}
    \caption{Environment maps from SkyGAN}
    \label{app:fig::SkyGAN}
\end{figure*}

\begin{figure*}[ht]
    \centering
    \includegraphics[width=\textwidth,height=0.95\textheight,keepaspectratio]{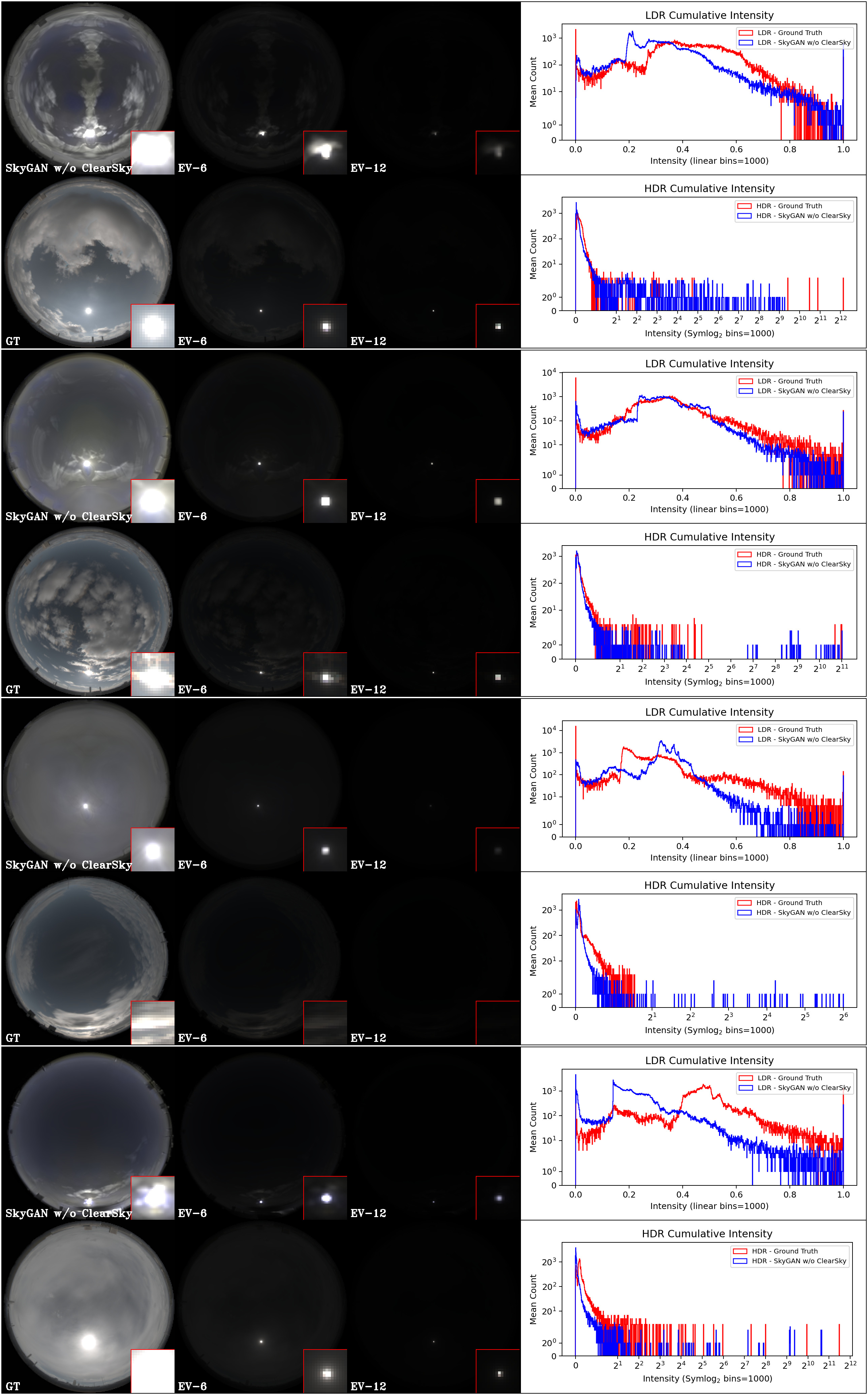}
    \caption{Environment maps from SkyGAN without clear sky priori}
    \label{app:fig::SkyGAN_noClearSky}
\end{figure*}

\begin{figure*}[ht]
    \centering
    \includegraphics[width=\textwidth,height=0.95\textheight,keepaspectratio]{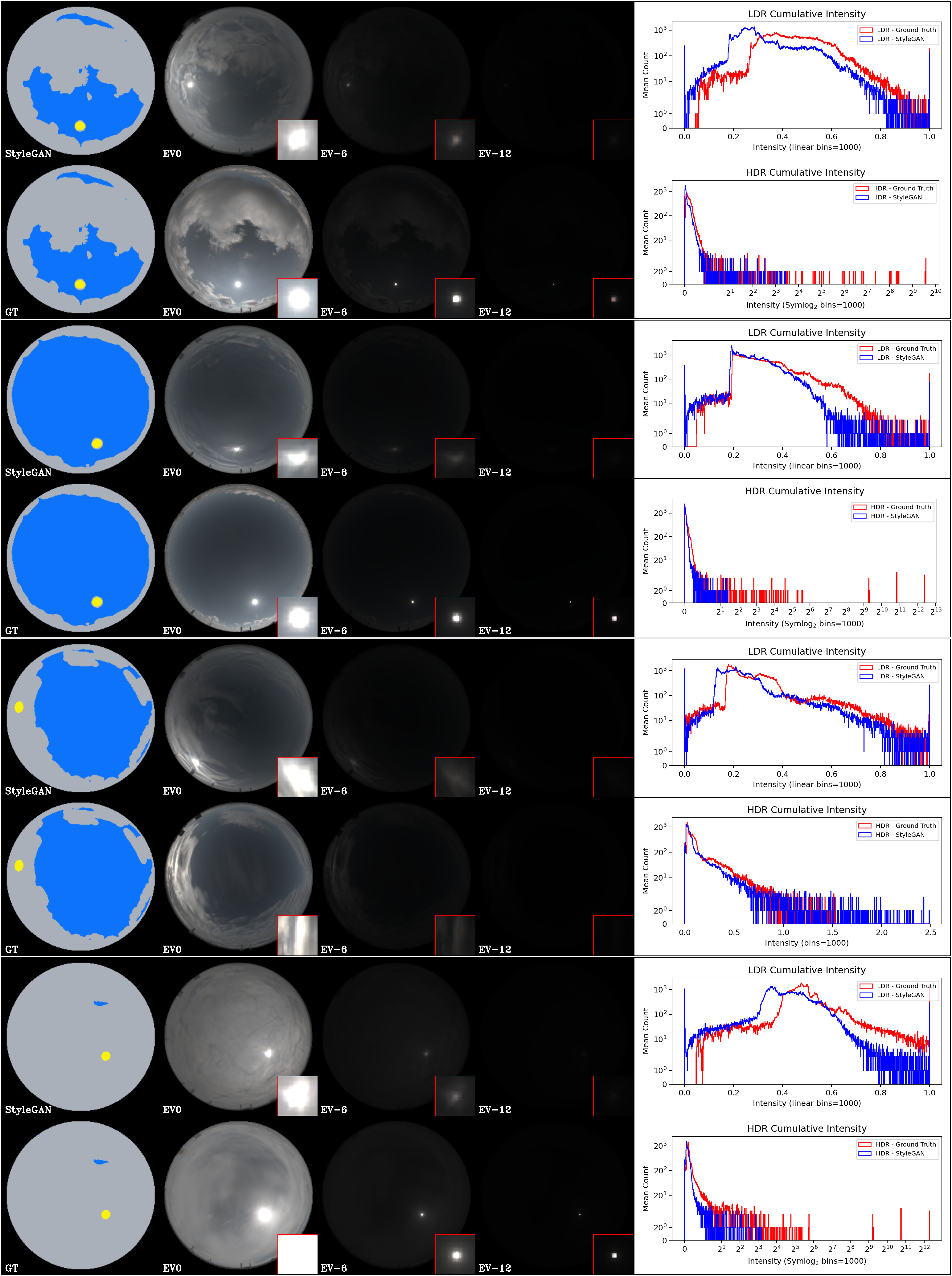}
    \caption{Environment maps from StyleGAN}
    \label{app:fig::StyleGAN}
\end{figure*}

\begin{figure*}[ht]
    \centering
    \includegraphics[width=\textwidth,height=0.95\textheight,keepaspectratio]{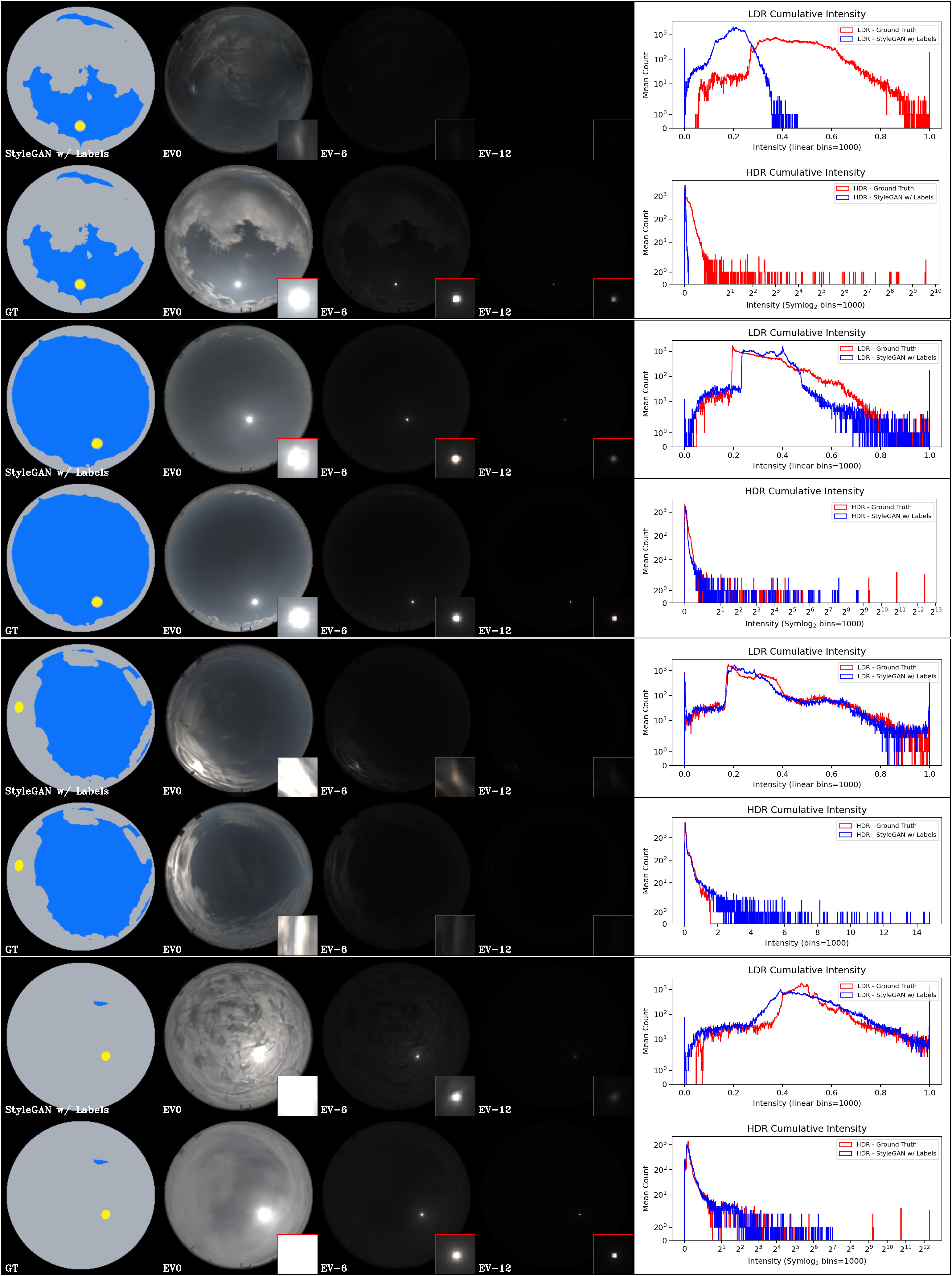}
    \caption{Environment maps from StyleGAN with labels}
    \label{app:fig::StyleGAN_wLabels}
\end{figure*}

\begin{figure*}[ht]
    \centering
    \includegraphics[width=\textwidth,height=0.95\textheight,keepaspectratio]{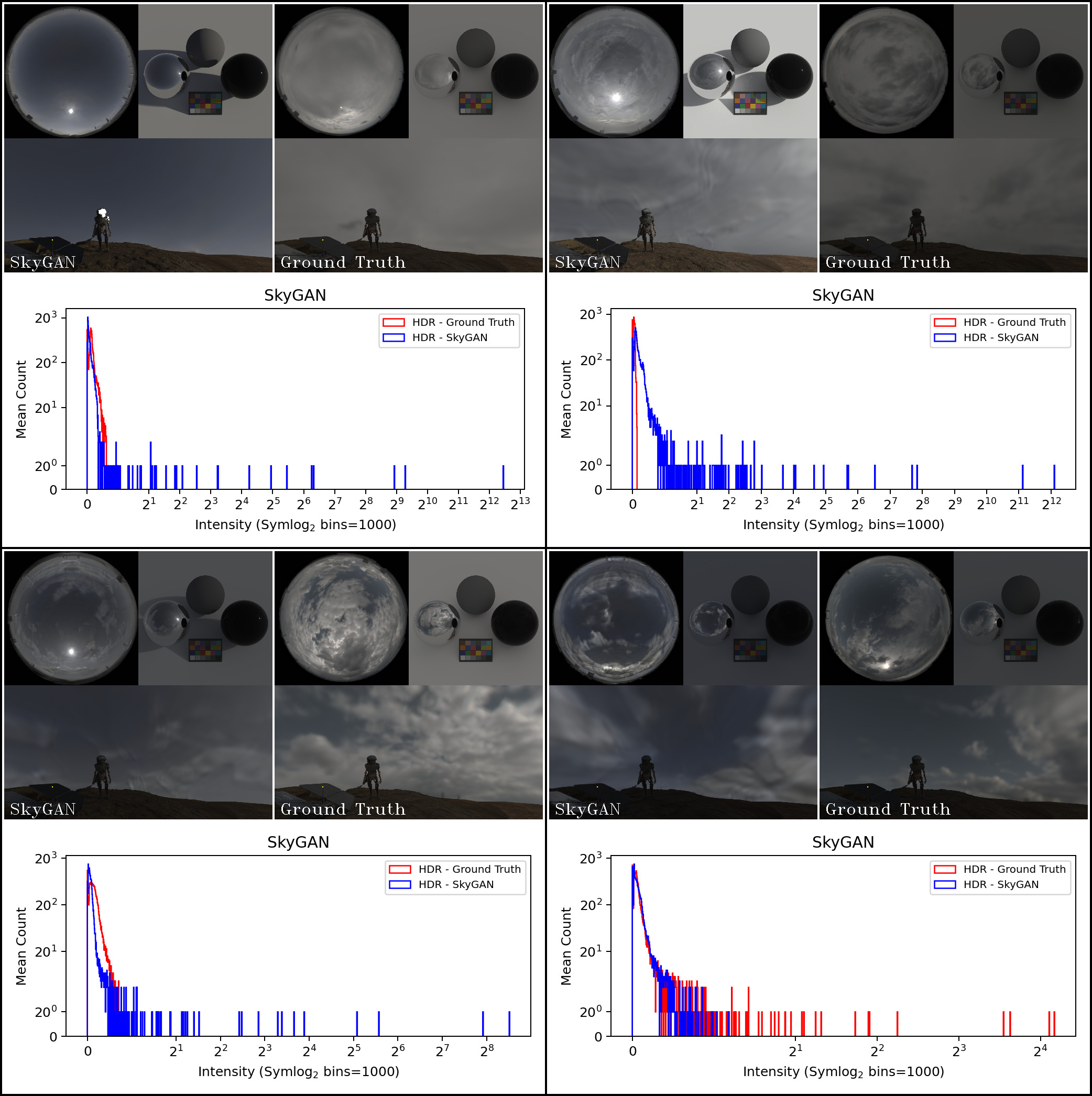}
    \caption{IBL renders of environment maps from SkyGAN}
    \label{app:fig::SkyGAN_renders}
\end{figure*}

\begin{figure*}[ht]
    \centering
    \includegraphics[width=\textwidth,height=0.95\textheight,keepaspectratio]{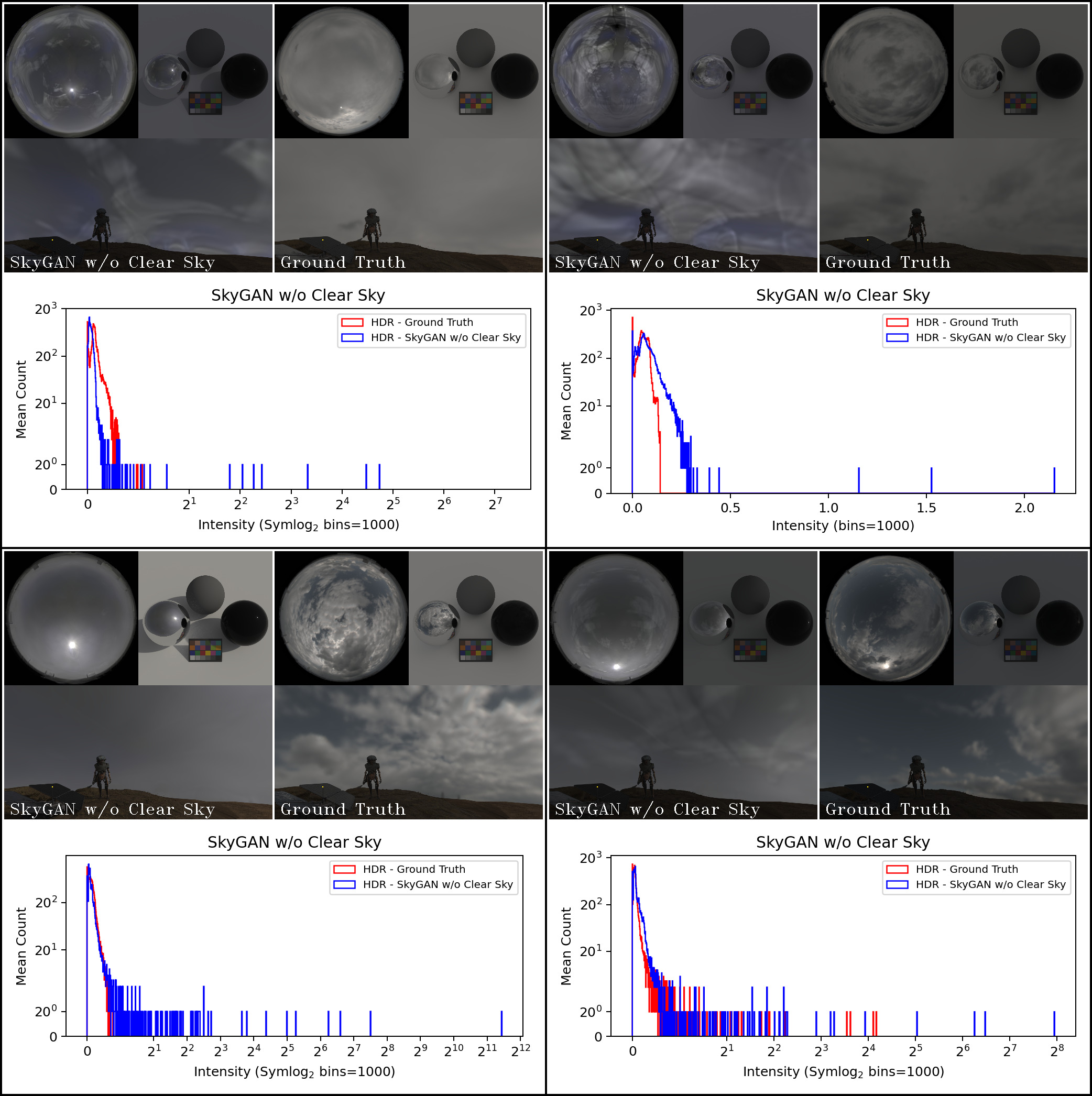}
    \caption{IBL renders of environment maps from SkyGAN w/o clear sky priori}
    \label{app:fig::SkyGAN_woClearSkies_renders}
\end{figure*}

\begin{figure*}[ht]
    \centering
    \includegraphics[width=\textwidth,height=0.95\textheight,keepaspectratio]{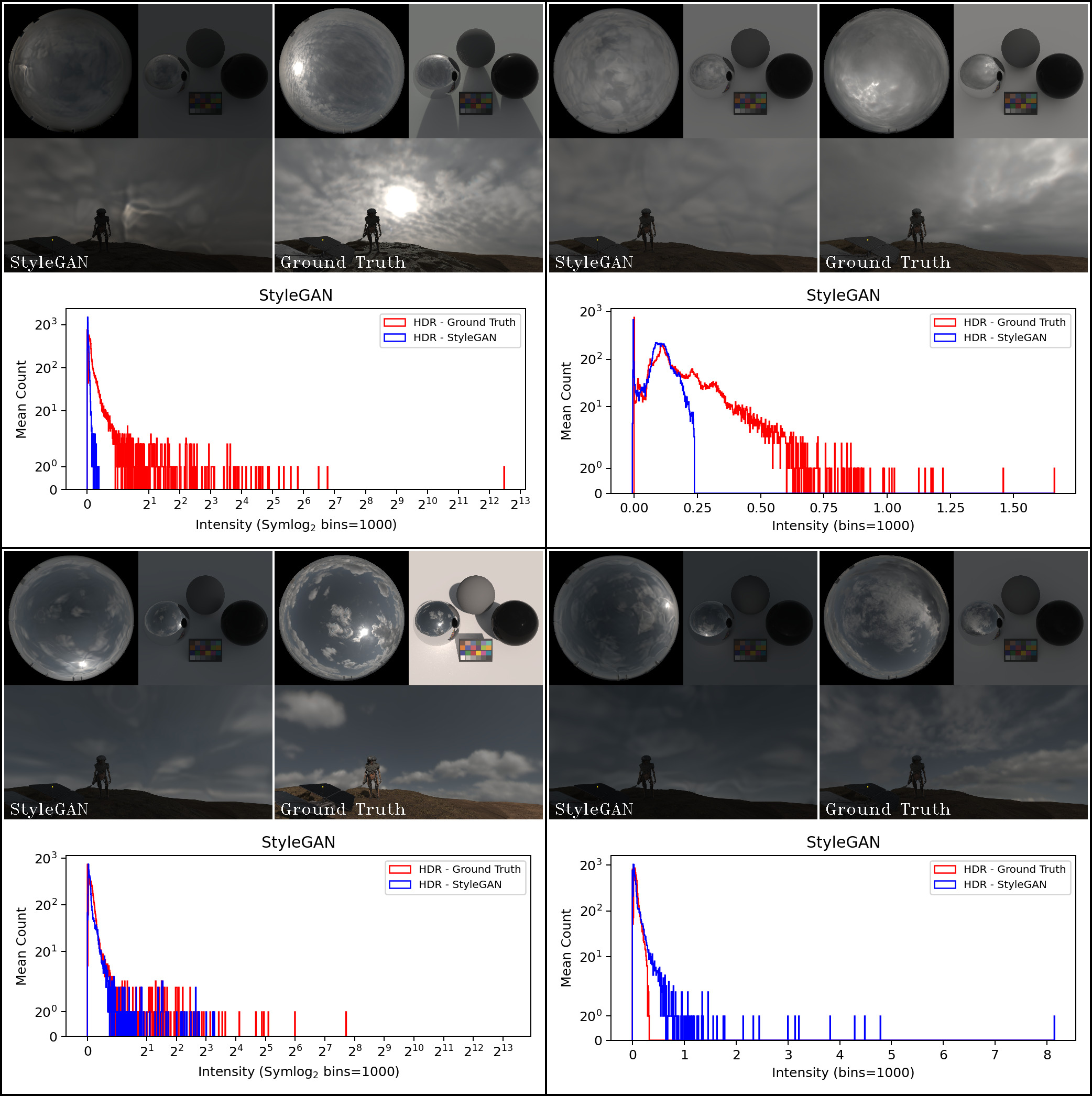}
    \caption{IBL renders of environment maps from StyleGAN w/o labels}
    \label{app:fig::StyleGAN_renders}
\end{figure*}

\begin{figure*}[ht]
    \centering
    \includegraphics[width=\textwidth,height=0.95\textheight,keepaspectratio]{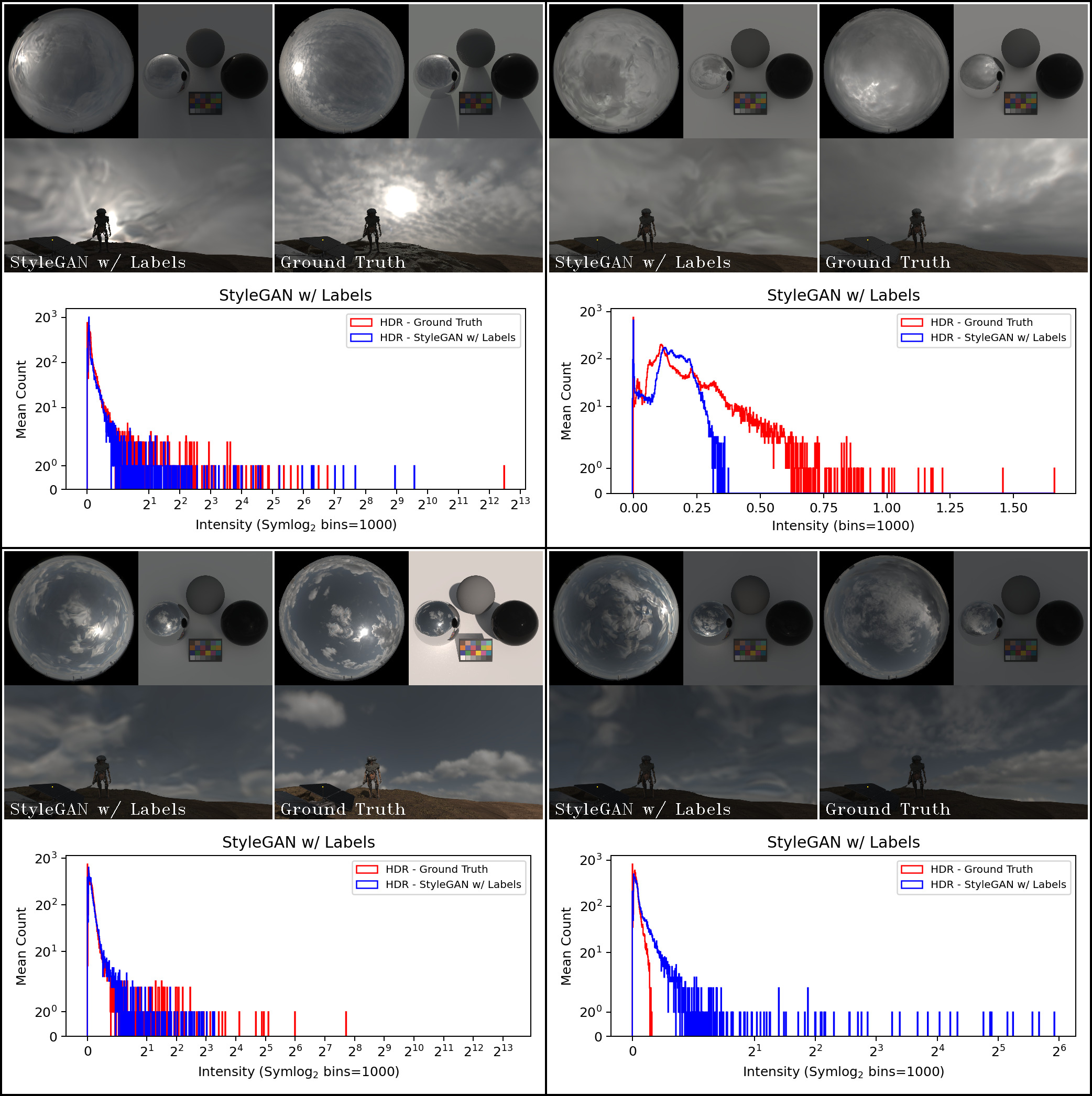}
    \caption{IBL renders of environment maps from StyleGAN with labels}
    \label{app:fig::StyleGAN_wLabels_renders}
\end{figure*}

We train SkyGAN per the authors' specification, adapting our HDRDB dataloader to provide the specified labels.
As illustrated in \cref{app:fig::SkyGAN_PSM_clearSkies}, we re-create the Prague Sky Model clear-skies (PSM) per solar-positioning from our segmentation of HDRDB.
Illustrations from each variant of SkyGAN are included in \cref{app:fig::SkyGAN,app:fig::SkyGAN_noClearSky,app:fig::StyleGAN,app:fig::StyleGAN_wLabels}.
For better visualization of cloud formations, additional IBL renders for each variant are included in \cref{app:fig::SkyGAN_renders,app:fig::SkyGAN_woClearSkies_renders,app:fig::StyleGAN_renders,app:fig::StyleGAN_wLabels_renders}.

Note, though illustrated alongside a ground truth HDRI from HDRDB, the relationship to SkyGAN's generated imagery is exclusive to solar positioning.
This relationship is lost with the removal of the clear-sky priori or HDRDB label.
StyleGAN w/labels was trained per our segmentation of HDRDB with a uniform sun label.



\clearpage

\begin{table*}[htb]
\centering
  \caption{
  Quantitative analysis of Text2Light for LDR and HDR imagery. Exposure matching is denoted by (*). LDR+Boost is the parametric boosting of the un-clipped LDR output to EDR (skipping iTMO).
  The results show that exposure matching Text2Light's HDR SRiTMO Boosted output is essential, with un-exposure-matched imagery under-performing LDR output.
  The results further demonstrate that bypassing iTMO can improve results across most metrics.
  }
  \label{app:tab::text2light_supplemental}
  \begin{tabular}{@{}r|cc|cc|ccc|cc@{}}
    \multicolumn{1}{c}{ } & \multicolumn{2}{c|} {\textbf{LDR}}
    & \multicolumn{2}{c|} {\textbf{HDR}}
    & \multicolumn{3}{c|}{\textbf{$T_{\gamma}$-cLDR}}
    & \multicolumn{2}{c} {\textbf{HDR}} \\
    \cline{2-10}
    Text2Light & $L_1$ $\downarrow$ & $L_2$ $\downarrow$ & $L_1$ $\downarrow$ & $L_2$ $\downarrow$ & LPIPS $\downarrow$ & CLIP-IQA $\downarrow$ & UQI$\uparrow$ & EV & $\oiint_I$ \\
    \midrule
    GT                & -    & -    & -    & -    & -    & 0.36    & -     & 12.85 & 3.54 \\
    LDR  & 0.13 & \textbf{0.04} & -    & -    & \textbf{0.41} &\textbf{ 0.37}    & 0.032 & -     & 3.96 \\
    HDR SRiTMO Boosted  & \textcolor{red}{0.29} & \textcolor{red}{0.15} & 0.27 & \textbf{816}  & \textcolor{red}{0.51} & \textcolor{red}{0.34}    & \textcolor{red}{0.006} & 8.4   & \textcolor{red}{0.16} \\
    HDR SRiTMO Boosted* & 0.13 & \textbf{0.04} & \textcolor{red}{0.32} & \textcolor{red}{1380} & 0.42 & 0.41    & 0.027 & \textbf{12.92} & 3.73 \\
    LDR+Boost* & \textbf{0.11} & \textbf{0.04} & \textbf{0.25} & 827  & 0.42 & 0.47    & \textbf{0.037} & \textcolor{red}{7.96}  & \textbf{3.69} \\

  \bottomrule
\end{tabular}
\end{table*}

\section{Text2Light}
\label{app:sec::text2light}

Evaluation of Text2Light's \textit{outdoor} functionality was completed via unaltered implementation the author's published code, checkpoints and textual CLIP prompts without upscaling ($\beta{=}1$) \cite{text2light}.
Generated LDR and HDR imagery were matched to ground truth images from the Laval Outdoor Dataset (LOD) \cite{YANNICK_2019_SKYNET}. \footnote{We evaluate Text2Light through comparison to imagery matched from the Laval Outdoor Dataset (LOD). Due to insufficient sample size, we do not report FID.}

\subsection{Seams}
\label{app:sec::text2light:::seams}

\begin{figure*}
    \centering
    \includegraphics[width=\textwidth,height=.955\textheight,keepaspectratio]{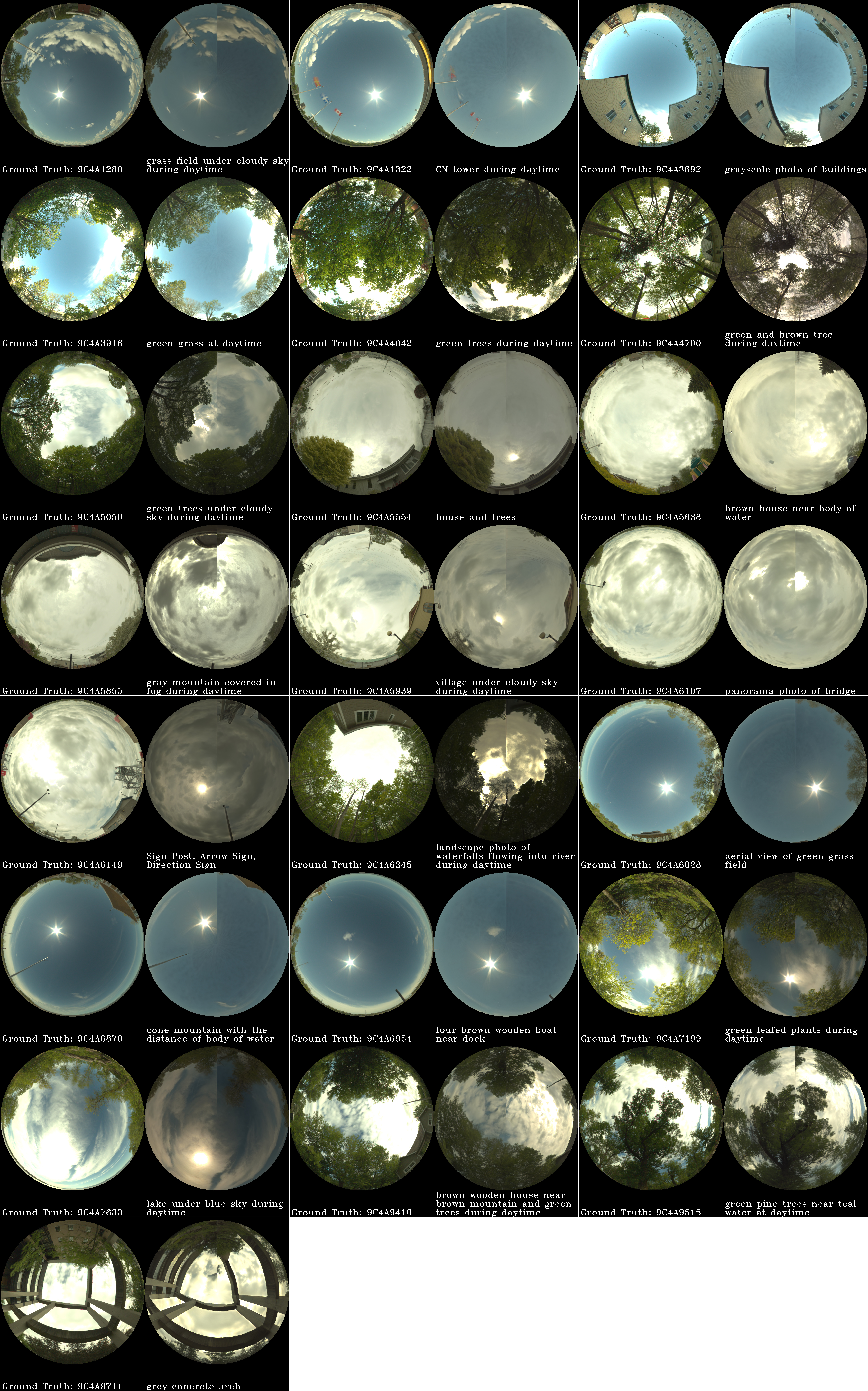}
    \caption{Matches, Sky-Angular format}
    \label{app:fig::text2light_skyAngular}
\end{figure*}

As illustrated in \cref{app:fig::text2light_skyAngular}, conditioning on latlong format skydomes exhibits `seams' from discontinuity which become apparent when formatted as sky-angular.
These visually unappealing seams impeded photorealism in IBL rendering.

\subsection{Matches}
\label{app:sec::text2light:::matches}

\begin{figure*}
    \centering
    \includegraphics[width=\textwidth,height=.95\textheight,keepaspectratio]{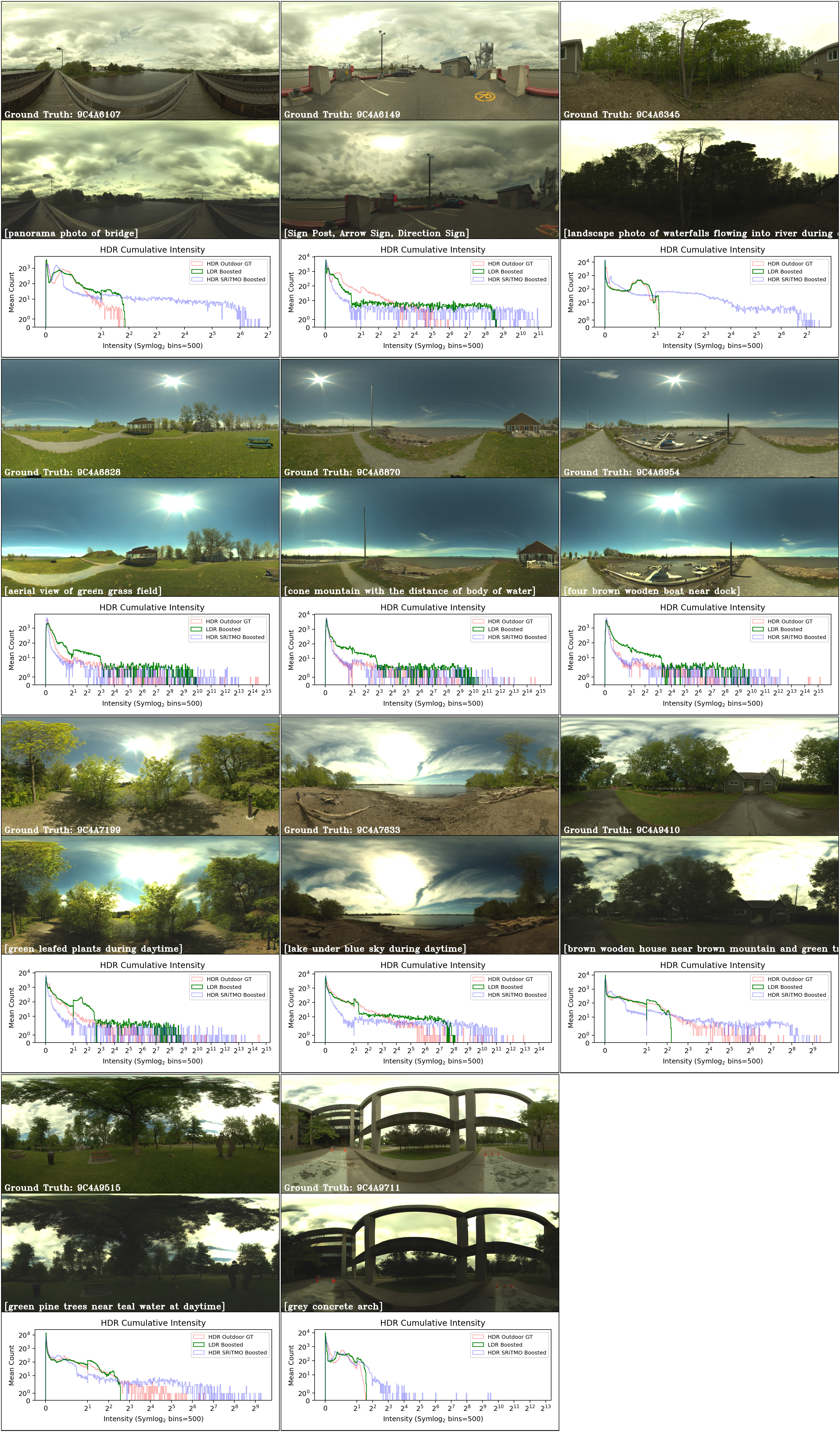}
    \caption{Parametric boosting of unclipped LDR diffusion model output (skipping iTMO))}
    \label{app:fig::text2light_noSRiTMO_2}
\end{figure*}

Given the visual similarity and negligible misalignment between the generated results and LOD, matching was accomplished by FLANN matched ORB keypoints and a scale invariant loss \cref{eq:scale_invariant_loss} \cite{eigen2014depth}.
Matches from a run of the outdoor textual CLIP prompts are included in \cref{app:fig::text2light_LOD1,app:fig::text2light_LOD2,app:fig::text2light_LOD3,app:fig::text2light_LOD4,app:fig::text2light_LOD5,app:fig::text2light_LOD6,app:fig::text2light_LOD7,app:fig::text2light_LOD8,app:fig::text2light_LOD9,app:fig::text2light_LOD10,app:fig::text2light_LOD11,app:fig::text2light_LOD12,app:fig::text2light_LOD13,app:fig::text2light_LOD14,app:fig::text2light_LOD15,app:fig::text2light_LOD16,app:fig::text2light_LOD17,app:fig::text2light_LOD18,app:fig::text2light_LOD19,app:fig::text2light_LOD20,app:fig::text2light_LOD21,app:fig::text2light_LOD22} and histograms are included for various stages of the model's pipeline, including:
\begin{enumerate}
    \item \textbf{Stage 1: LDR}: \\ {The LDR output of the diffusion model (clipped)}
    \item \textbf{Stage 2: HDR SRiTMO}: \\ {The HDR output from the Inverse Tone-Mapping Operator (iTMO) MLP}. Up-scaling by Text2Light's Super Resolution (SR) component is skipped ($\beta{=}1$).
    \item \textbf{HDR SRiTMO Boosted}: \\ {The EDR output from parametric boosting of HDR SRiTMO per \cref{eq:text2light_parametric_boost}}
\end{enumerate}
In \cref{app:fig::text2light_noSRiTMO_1,app:fig::text2light_noSRiTMO_2}, we include the EDR output from applying the parametric boosting directly to the output of the LDR diffusion model (unclipped).

\subsection{Supplemental Results}
\label{app:sec::text2light:::supplemental_results}

\begin{equation}
    \begin{split}
        \mathcal{L}(I_{x},I_{y}) =
        & \text{ }\frac{1}{n}\sum (\ln{I_{x}} - \ln{I_{y}})^2 \\
        & - \frac{1}{n^2} \left( \sum (\ln{I_{x}} - \ln{I_{y}}) \right)^2
    \end{split}
    \label{eq:scale_invariant_loss}
\end{equation}

Text2Light's inverse Tone-Mapping Operator (iTMO) MLP is trained on tone-mapped LOD HDRI ($I'$) using the linear-scale invariant loss in \cref{eq:scale_invariant_loss}, under the assumption $|I'_{real}| \approx \alpha |I'_{fake}|$ in LDR space.
We have demonstrated through \cref{fig:plt_tm} that this assumption is precarious, as all tone-mappers introduce non-linearity with profound impact at high intensities.

To demonstrate the importance of exposure matching, \cref{app:tab::text2light_supplemental} summarizes Text2Light's HDR output before and after exposure matching.
The results show that exposure matching is a crucial step, without which Text2Light's HDR imagery under-performs LDR output for all metrics.
We note that our exposure matching is not manual reproducible through conventional exposure and/or offset sliders.

\begin{figure*}
    \centering
    \includegraphics[width=\textwidth,height=.95\textheight,keepaspectratio]{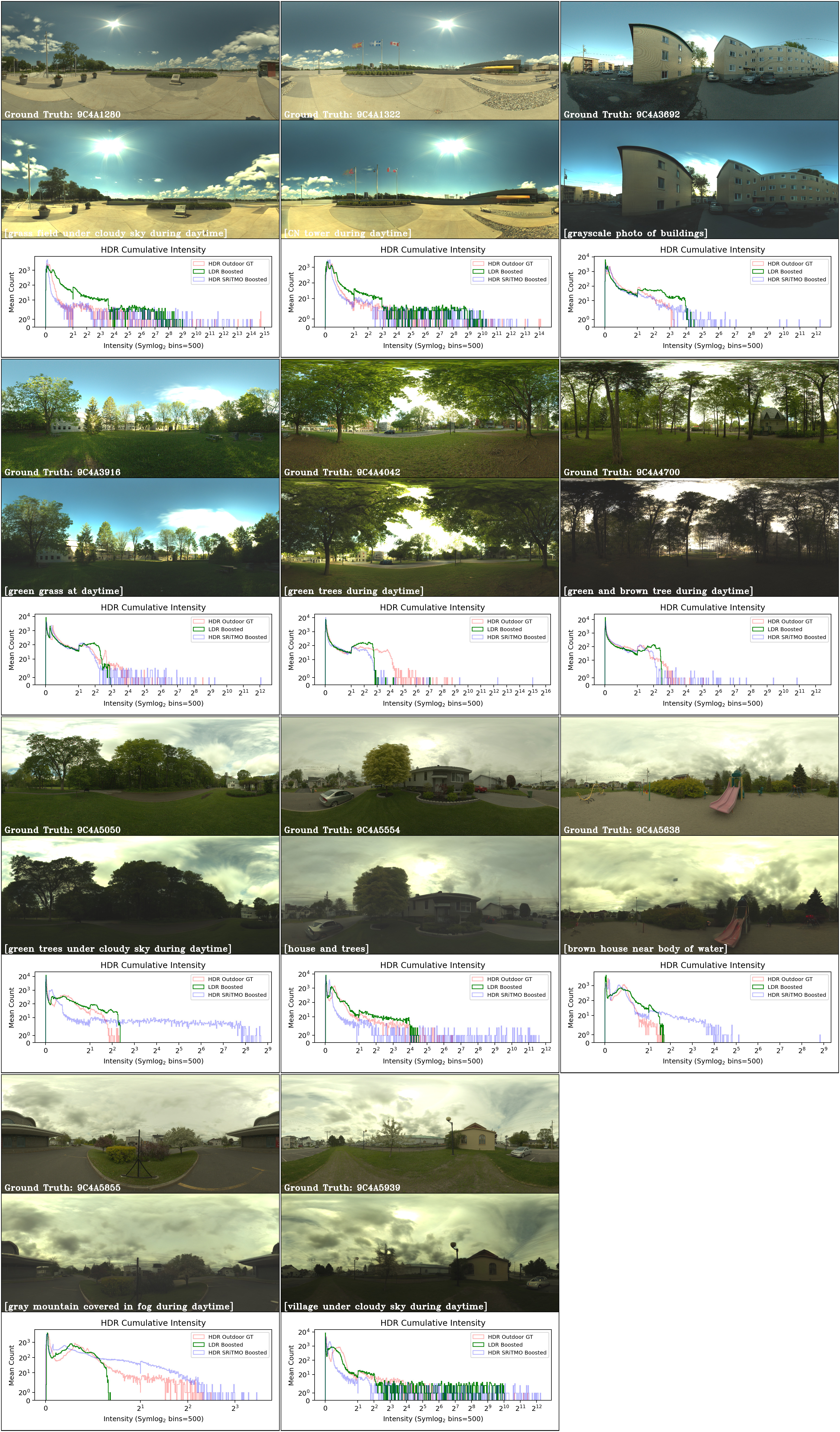}
    \caption{Parametric boosting of the unclipped output of the LDR diffusion model (skipping iTMO)}
    \label{app:fig::text2light_noSRiTMO_1}
\end{figure*}

We further experiment with Text2Light's parametric boosting in \cref{eq:text2light_parametric_boost} and find the results to be inconsistent and to produce incoherent results when imagery has little or no content below the horizon.
To determine the impact of iTMO, we attempted to bypass it using \cref{eq:text2light_parametric_boost} exclusively to decompress the LDR output of the diffusion model to EDR.
We observe that iTMO output has the range $[-8,8]$ and we reflect this distribution by substituting $\max\left( \overline{\overline{I_{b}}}\right)=8$ and adding a linear transform to map the raw $[-1,1]$ LDR output to $[-8,8]$.
Through experimentation, we set the remaining parameters of \cref{eq:text2light_parametric_boost} as: $\vartheta=0.83$, $\rho=8$, $\gamma=0.5$, and $\beta=0.2$.
Using images previously paired with LOD, we produce \cref{app:fig::text2light_noSRiTMO_1,app:fig::text2light_noSRiTMO_1} and quantitatively summarize the results in \cref{app:tab::text2light_supplemental} through \textit{LDR+Boost*}.
The results demonstrate the impact of iTMO to be dubious, given skipping iTMO and augmenting LDR directly to EDR through \cref{eq:text2light_parametric_boost} improves performance across most metrics.

\begin{figure*}
    \centering
    \includegraphics[width=\textwidth,height=.95\textheight,keepaspectratio]{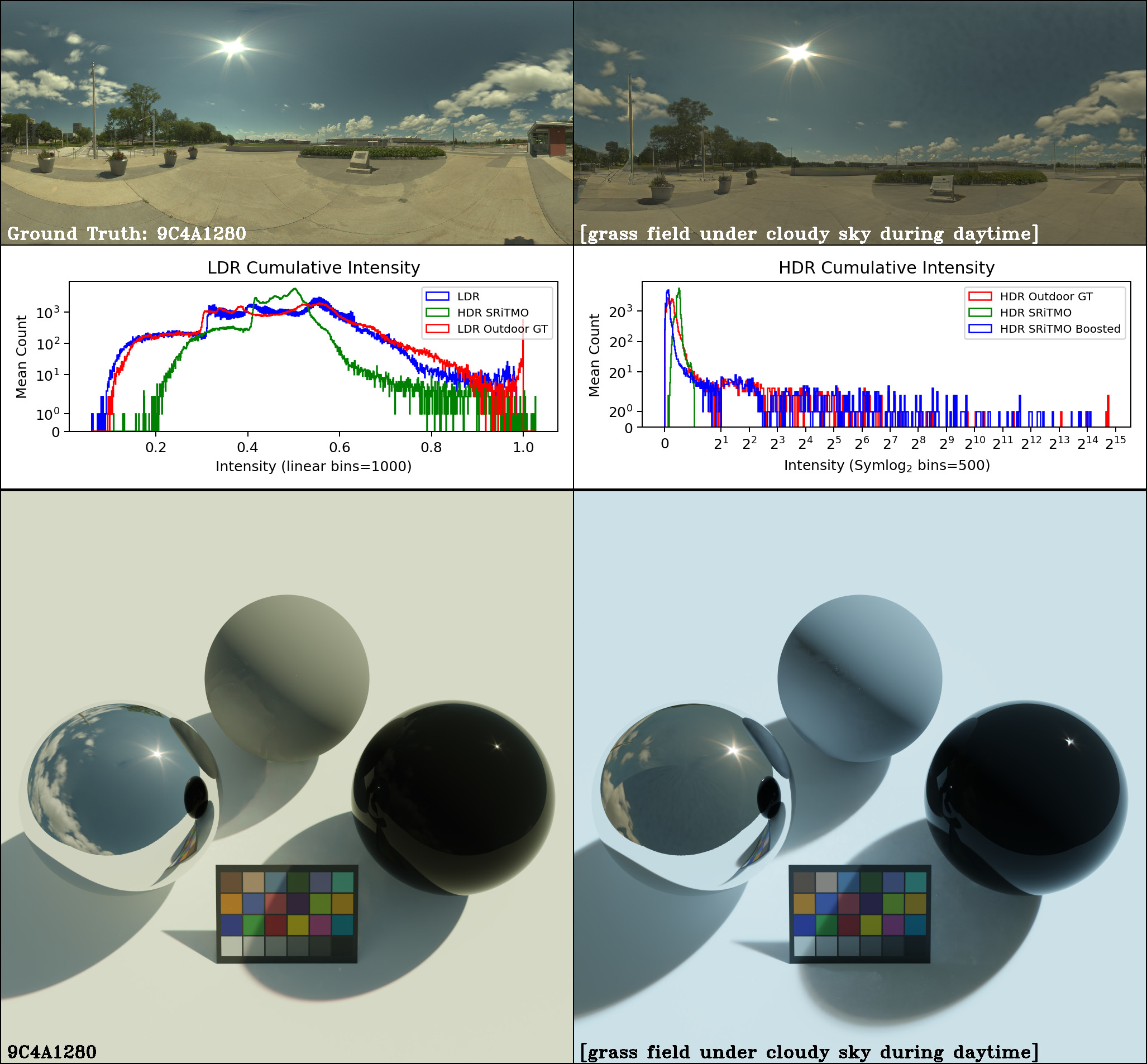}
    \caption{LOD 9C4A1280}
    \label{app:fig::text2light_LOD1}
\end{figure*}

\begin{figure*}
    \centering
    \includegraphics[width=\textwidth,height=.95\textheight,keepaspectratio]{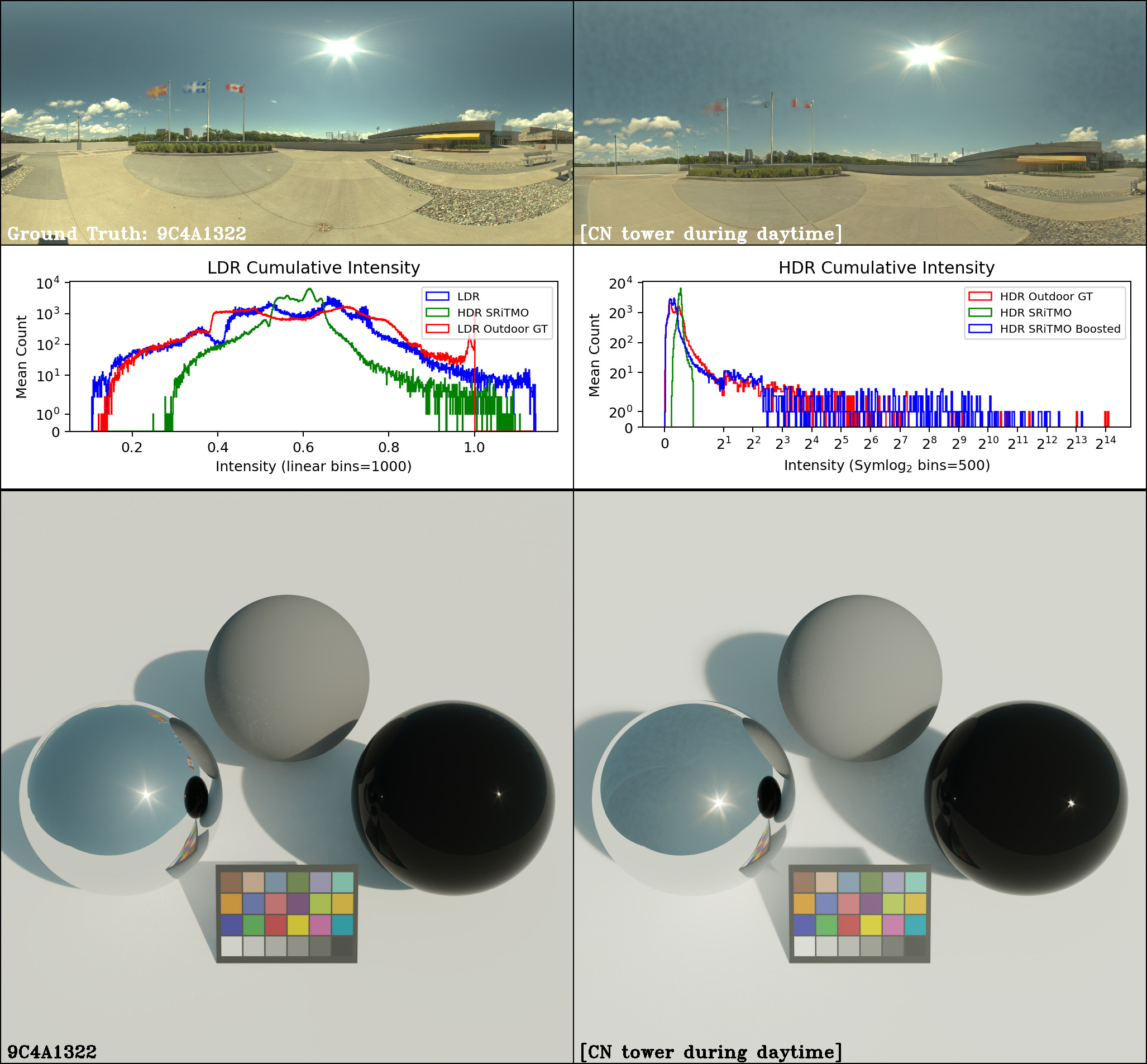}
    \caption{LOD 9C4A1322}
    \label{app:fig::text2light_LOD2}
\end{figure*}

\begin{figure*}
    \centering
    \includegraphics[width=\textwidth,height=.95\textheight,keepaspectratio]{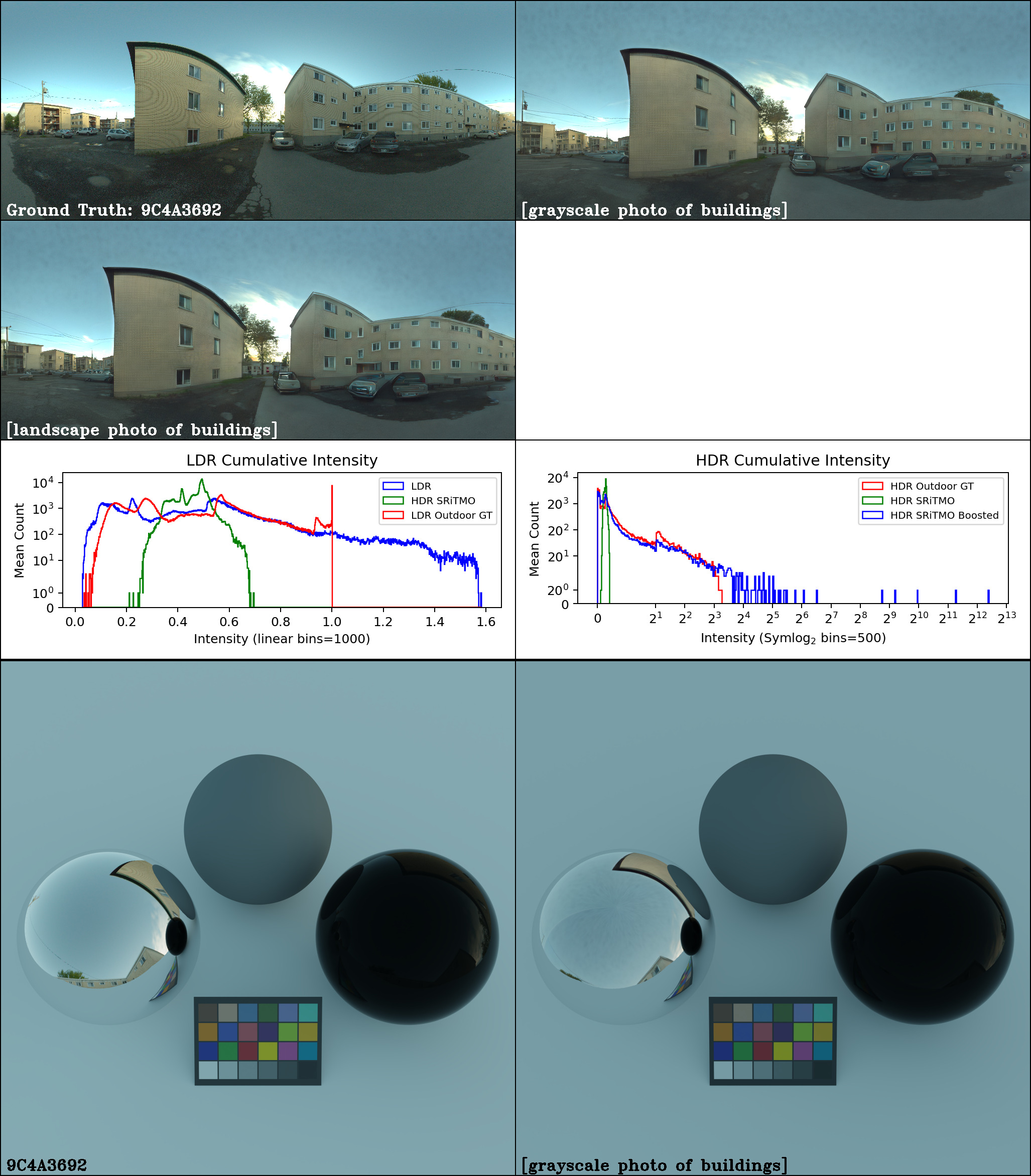}
    \caption{LOD 9C4A3692}
    \label{app:fig::text2light_LOD3}
\end{figure*}

\begin{figure*}
    \centering
    \includegraphics[width=\textwidth,height=.95\textheight,keepaspectratio]{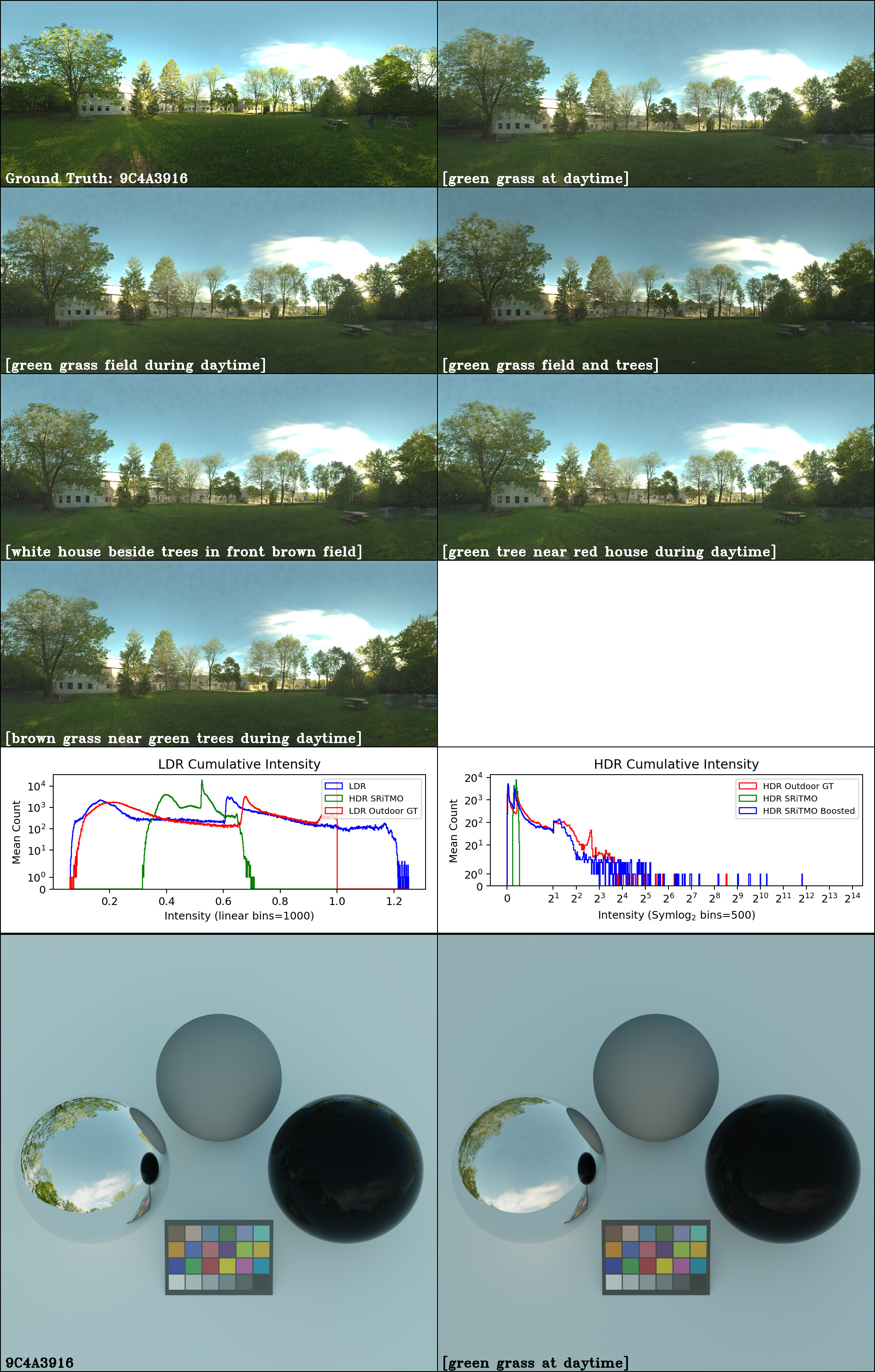}
    \caption{LOD 9C4A3916}
    \label{app:fig::text2light_LOD4}
\end{figure*}

\begin{figure*}
    \centering
    \includegraphics[width=\textwidth,height=.95\textheight,keepaspectratio]{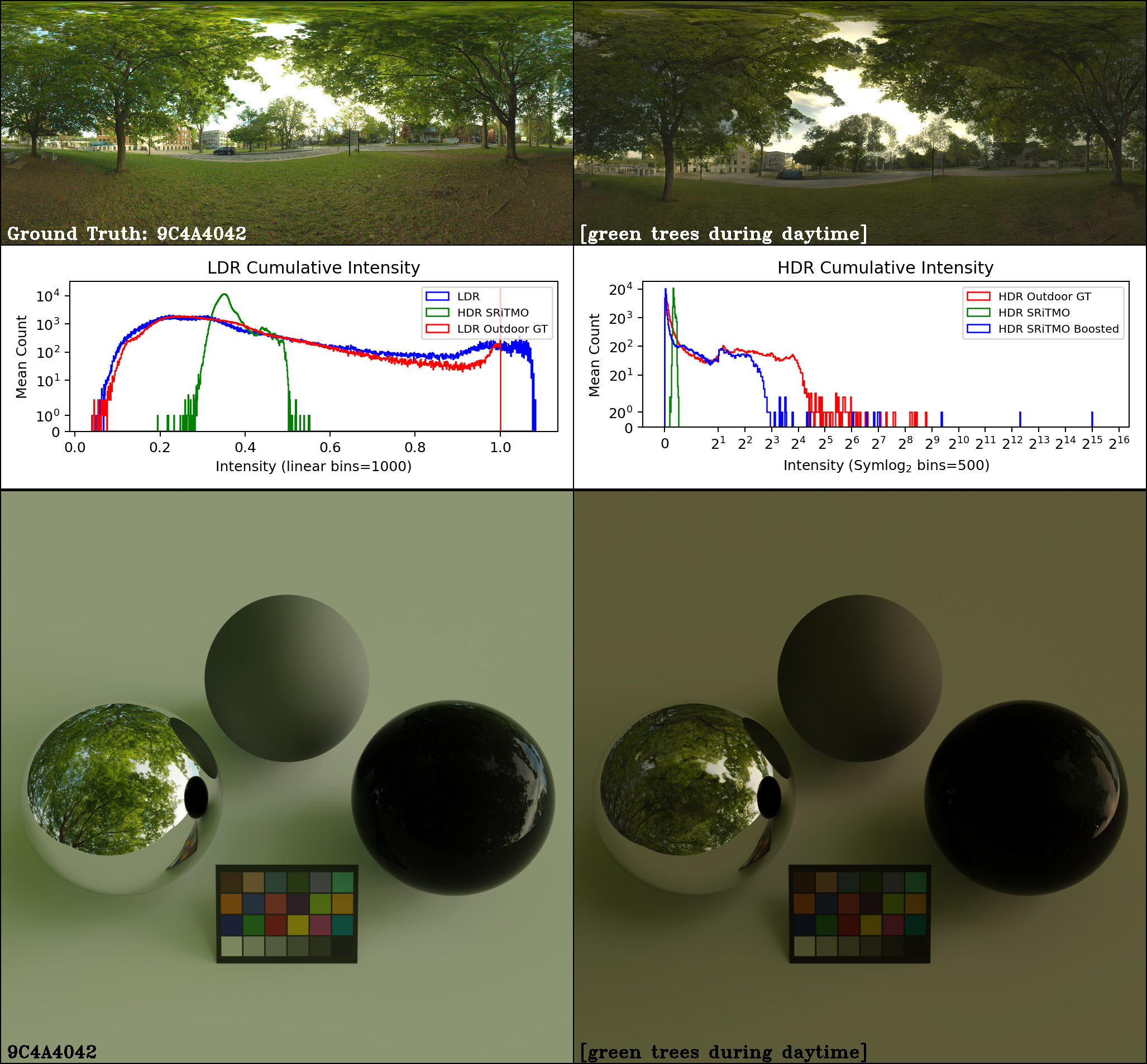}
    \caption{LOD 9C4A4042}
    \label{app:fig::text2light_LOD5}
\end{figure*}

\begin{figure*}
    \centering
    \includegraphics[width=\textwidth,height=.95\textheight,keepaspectratio]{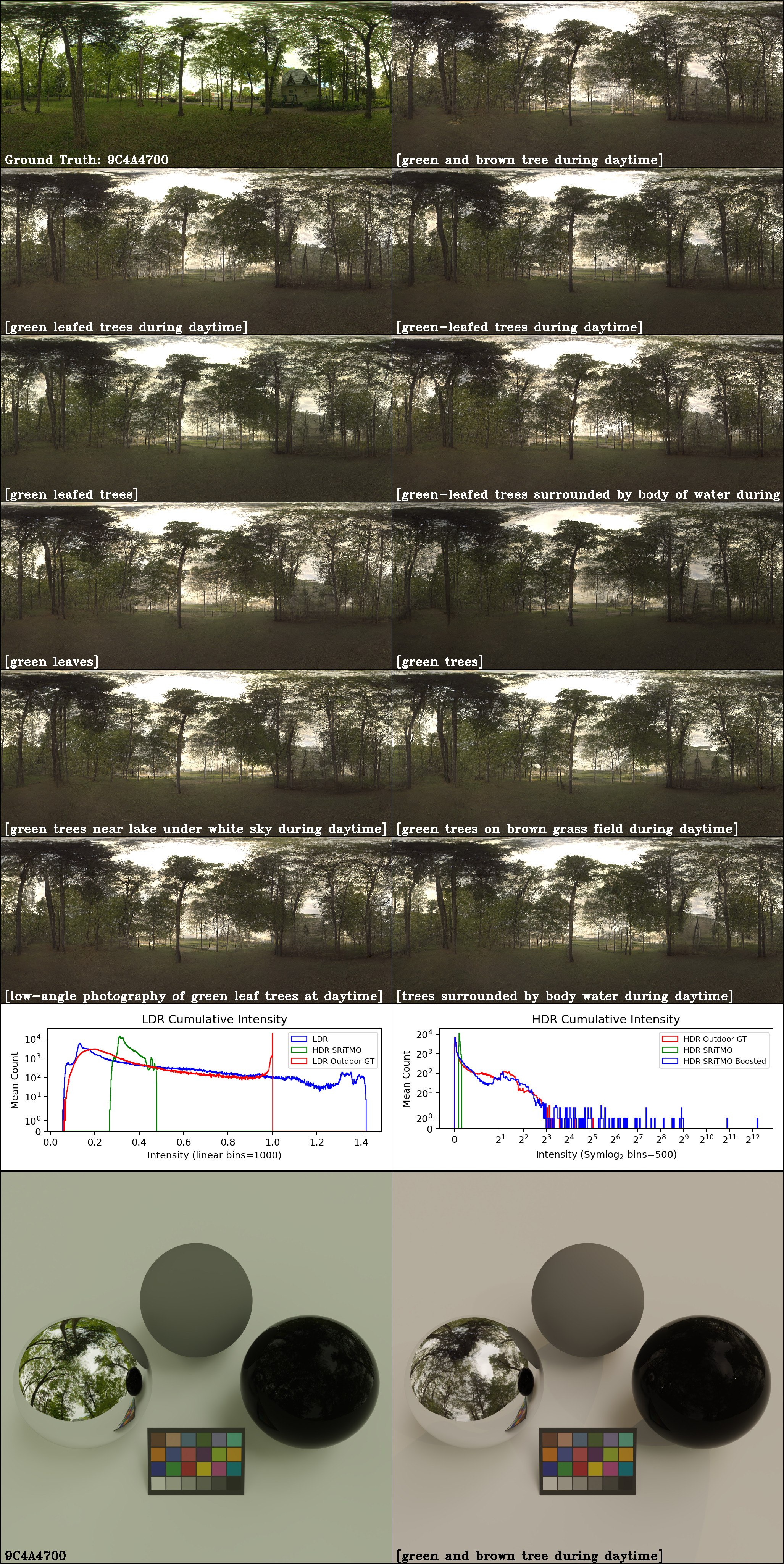}
    \caption{LOD 9C4A4700}
    \label{app:fig::text2light_LOD6}
\end{figure*}

\begin{figure*}
    \centering
    \includegraphics[width=\textwidth,height=.95\textheight,keepaspectratio]{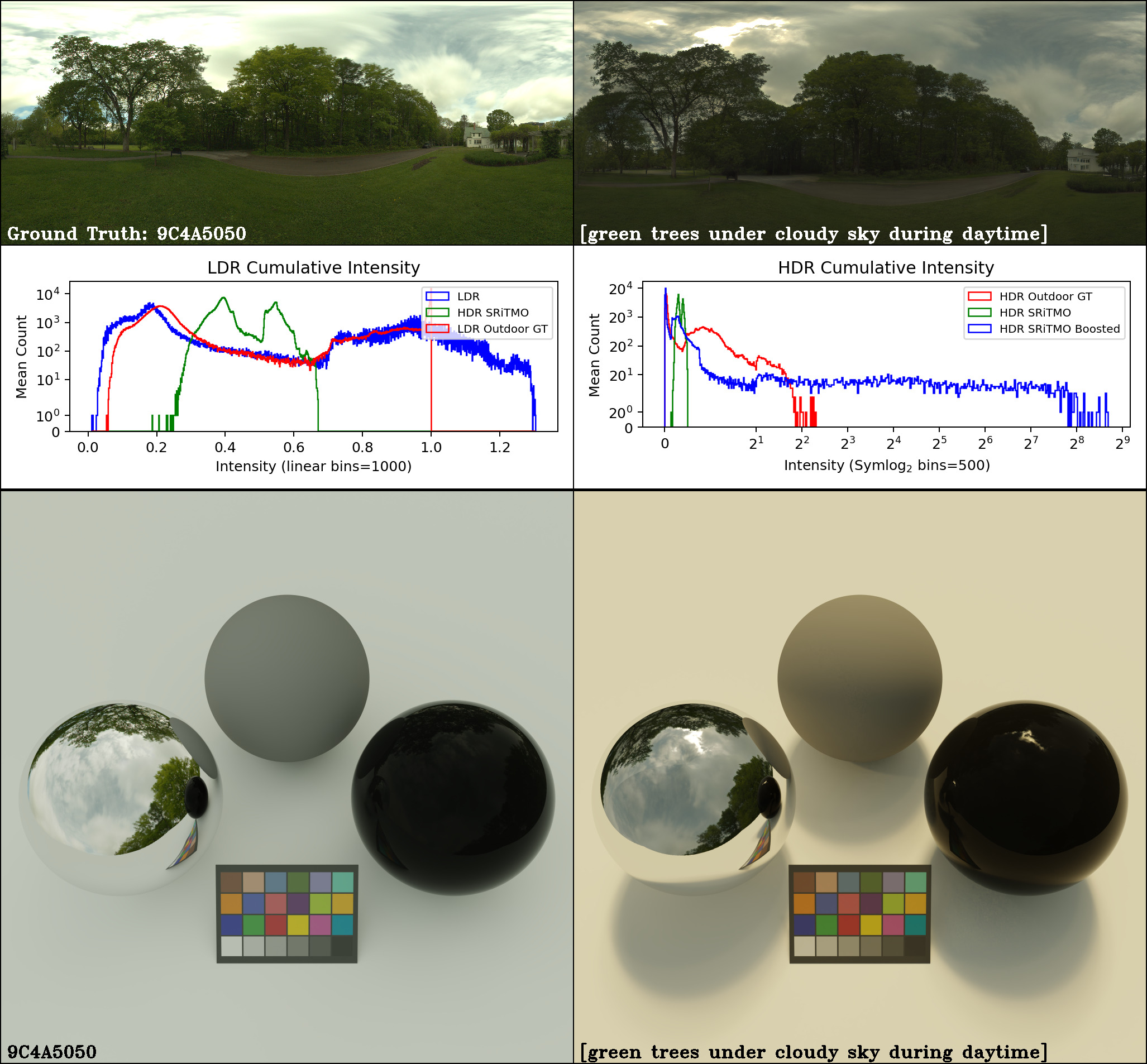}
    \caption{LOD 9C4A5050}
    \label{app:fig::text2light_LOD7}
\end{figure*}

\begin{figure*}
    \centering
    \includegraphics[width=\textwidth,height=.95\textheight,keepaspectratio]{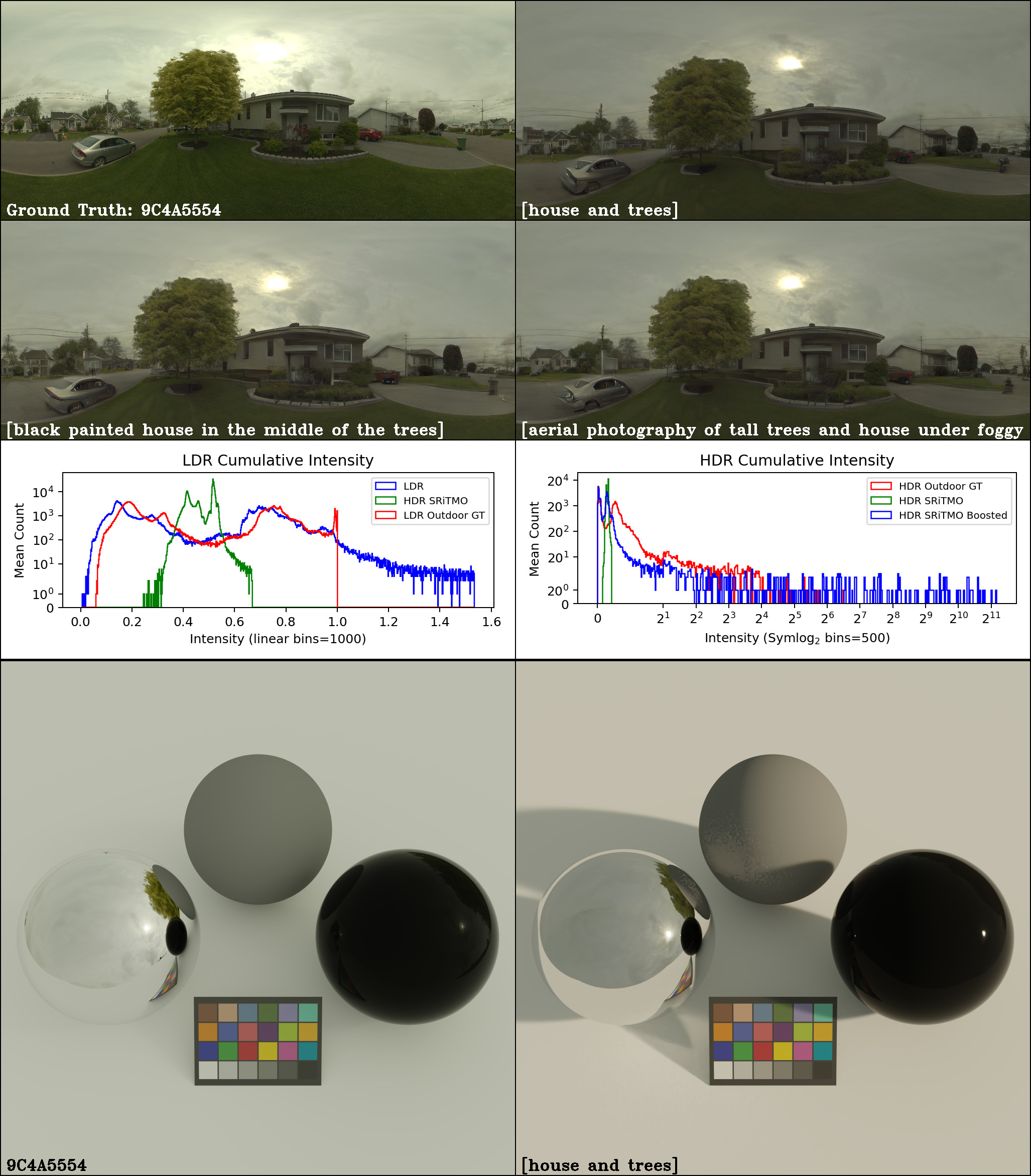}
    \caption{LOD 9C4A5554}
    \label{app:fig::text2light_LOD8}
\end{figure*}

\begin{figure*}
    \centering
    \includegraphics[width=\textwidth,height=.95\textheight,keepaspectratio]{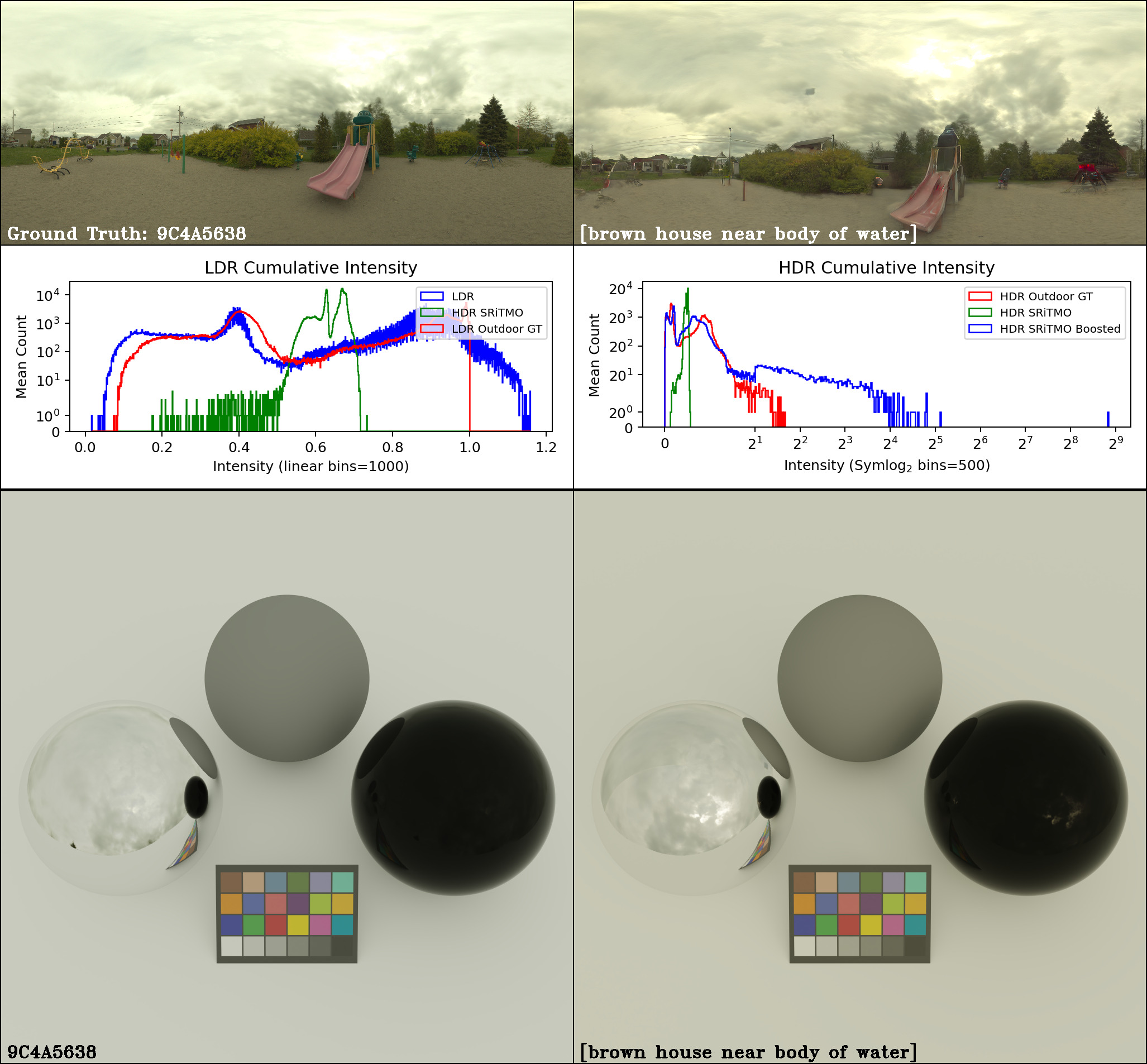}
    \caption{LOD 9C4A5638}
    \label{app:fig::text2light_LOD9}
\end{figure*}

\begin{figure*}
    \centering
    \includegraphics[width=\textwidth,height=.95\textheight,keepaspectratio]{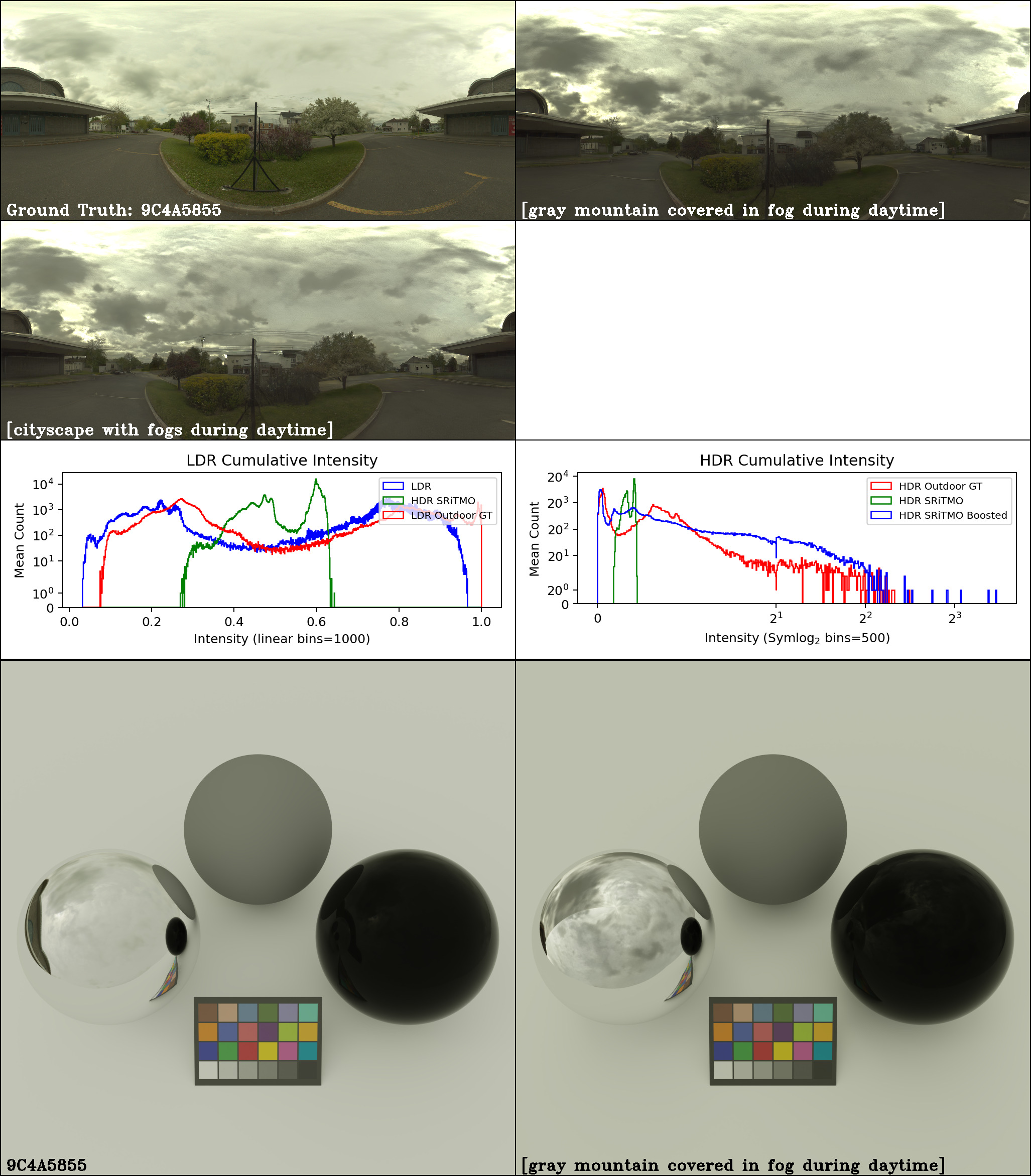}
    \caption{LOD 9C4A5855}
    \label{app:fig::text2light_LOD10}
\end{figure*}

\begin{figure*}
    \centering
    \includegraphics[width=\textwidth,height=.95\textheight,keepaspectratio]{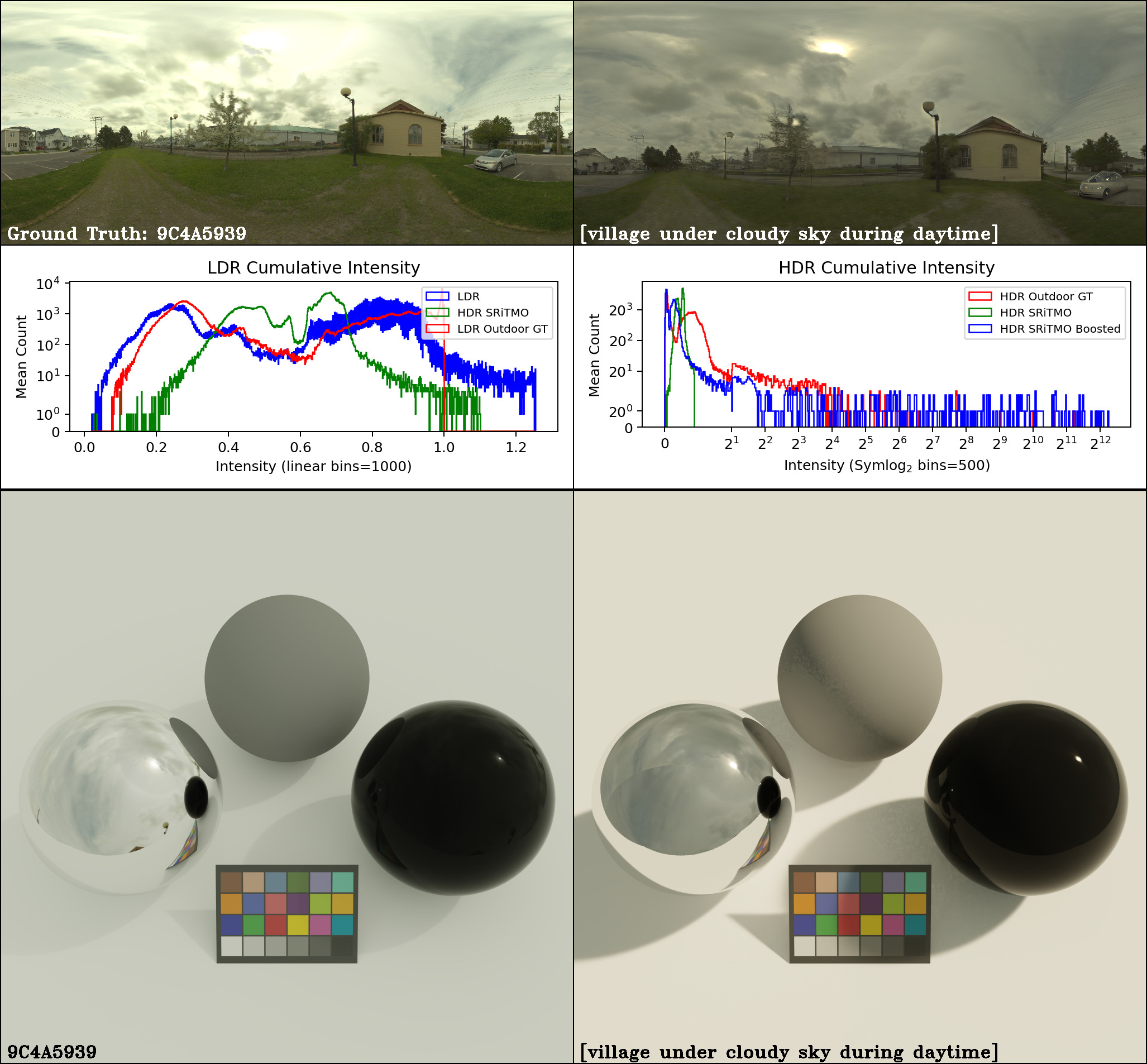}
    \caption{LOD 9C4A5939}
    \label{app:fig::text2light_LOD11}
\end{figure*}

\begin{figure*}
    \centering
    \includegraphics[width=\textwidth,height=.95\textheight,keepaspectratio]{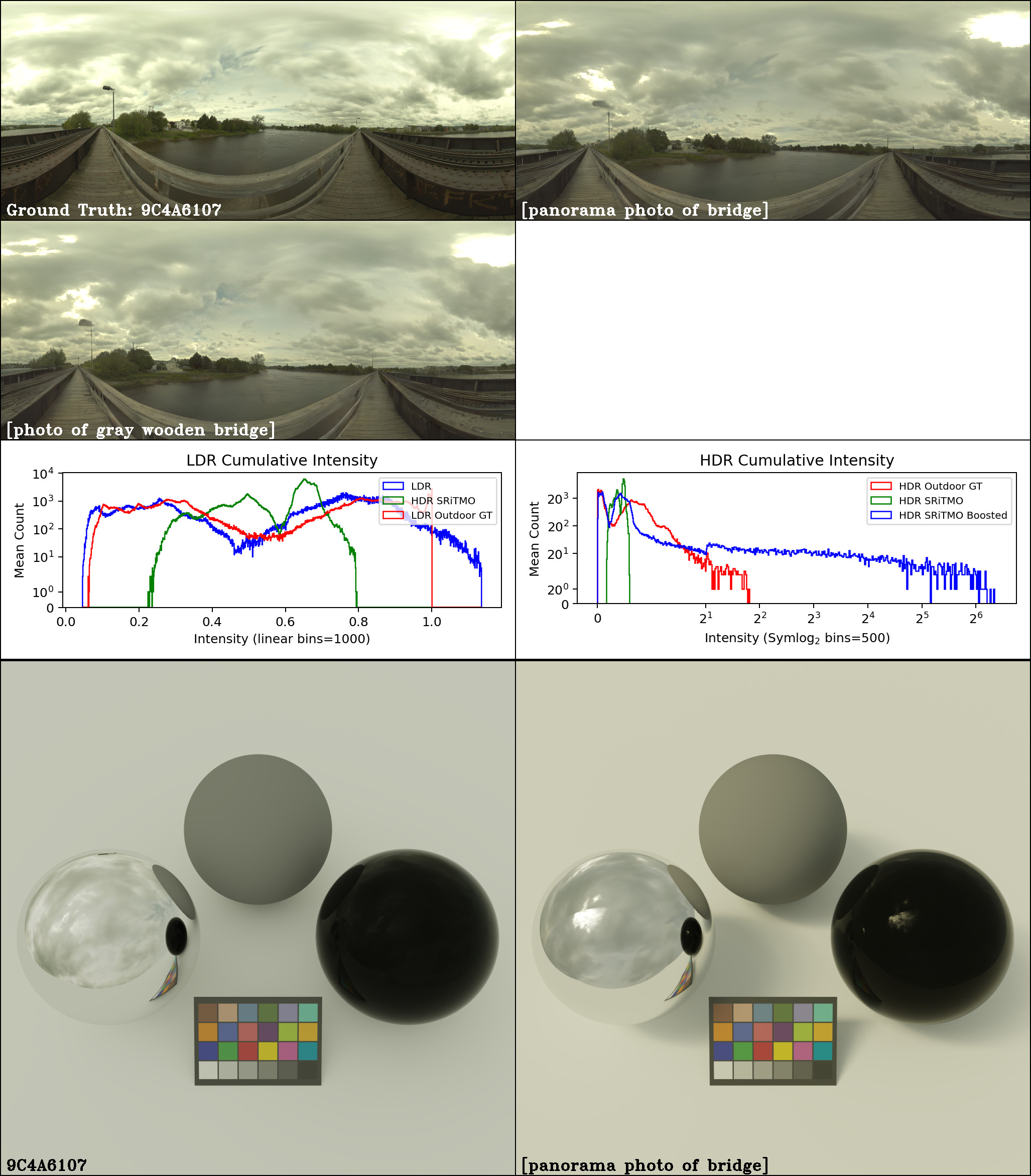}
    \caption{LOD 9C4A6107}
    \label{app:fig::text2light_LOD12}
\end{figure*}

\begin{figure*}
    \centering
    \includegraphics[width=\textwidth,height=.95\textheight,keepaspectratio]{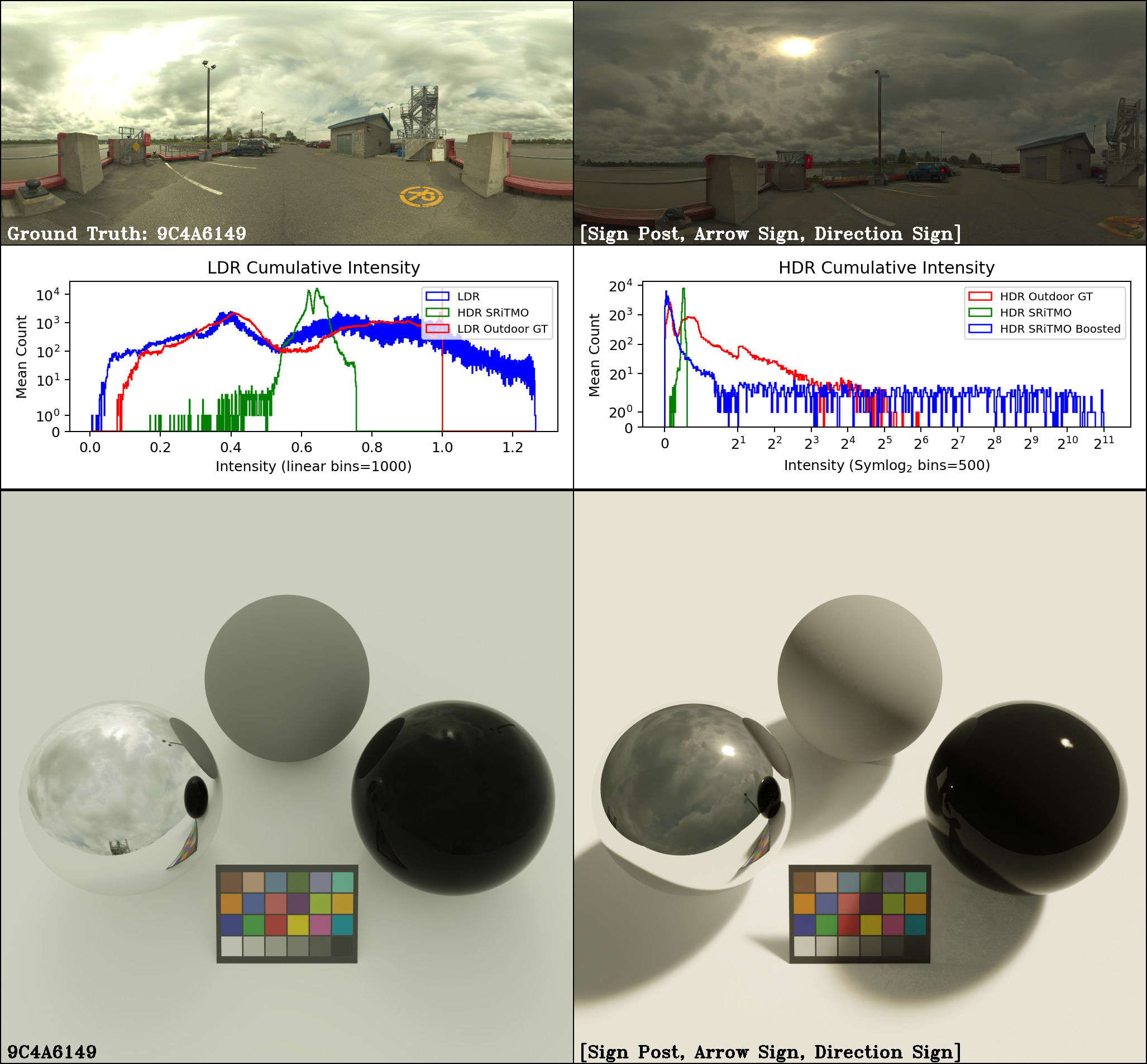}
    \caption{LOD 9C4A6149}
    \label{app:fig::text2light_LOD13}
\end{figure*}

\begin{figure*}
    \centering
    \includegraphics[width=\textwidth,height=.95\textheight,keepaspectratio]{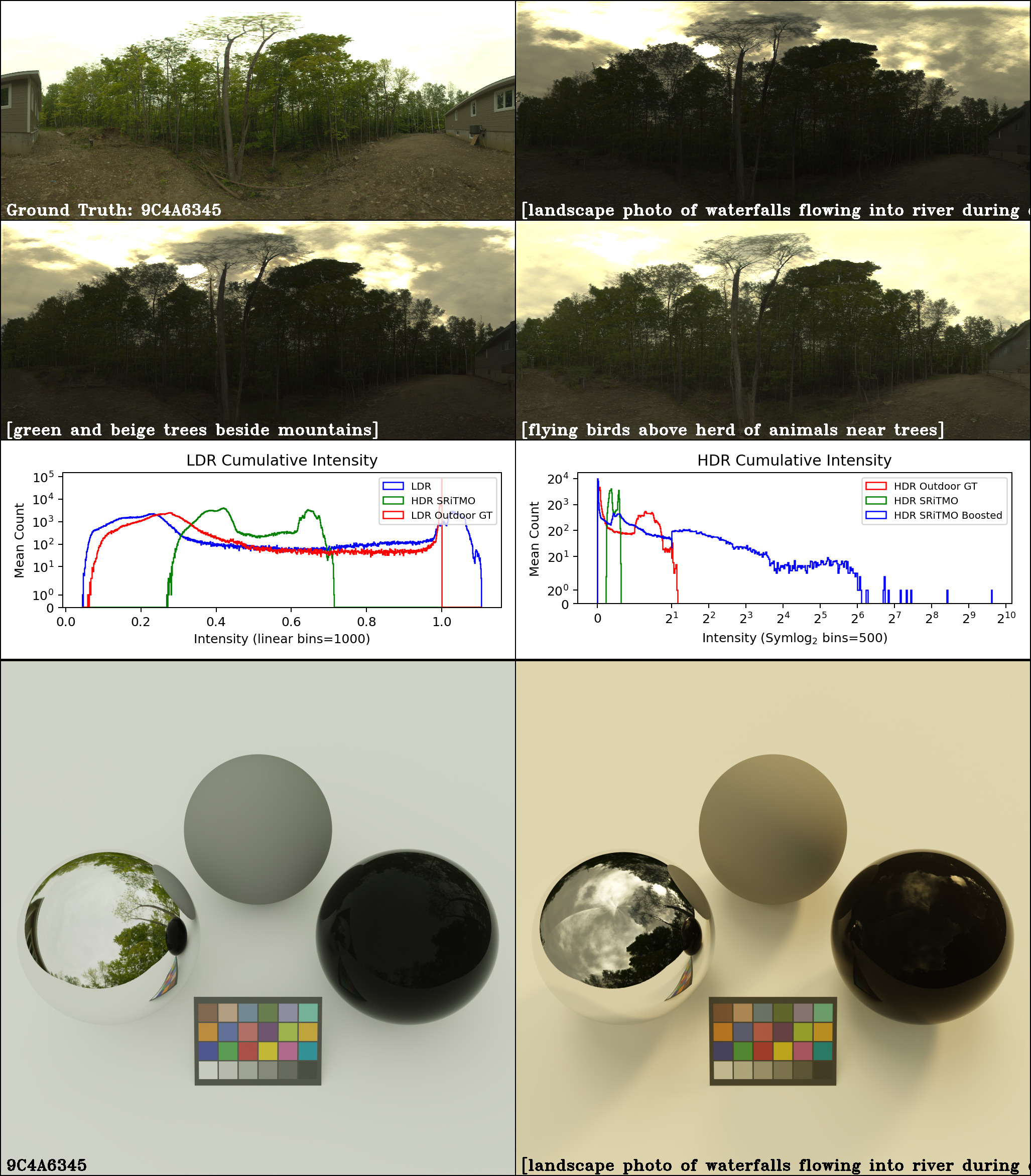}
    \caption{LOD 9C4A6345}
    \label{app:fig::text2light_LOD14}
\end{figure*}

\begin{figure*}
    \centering
    \includegraphics[width=\textwidth,height=.95\textheight,keepaspectratio]{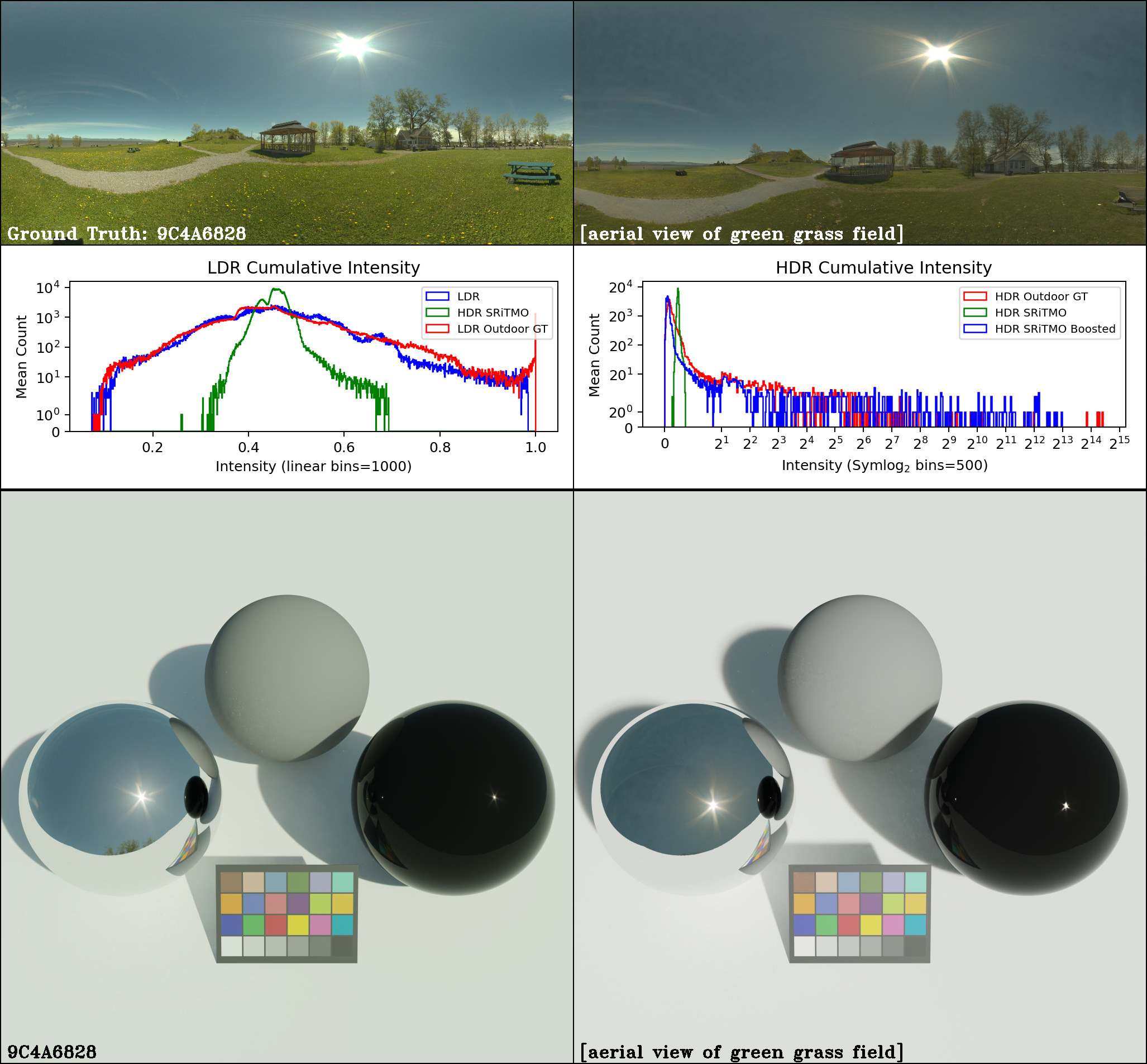}
    \caption{LOD 9C4A6828}
    \label{app:fig::text2light_LOD15}
\end{figure*}

\begin{figure*}
    \centering
    \includegraphics[width=\textwidth,height=.95\textheight,keepaspectratio]{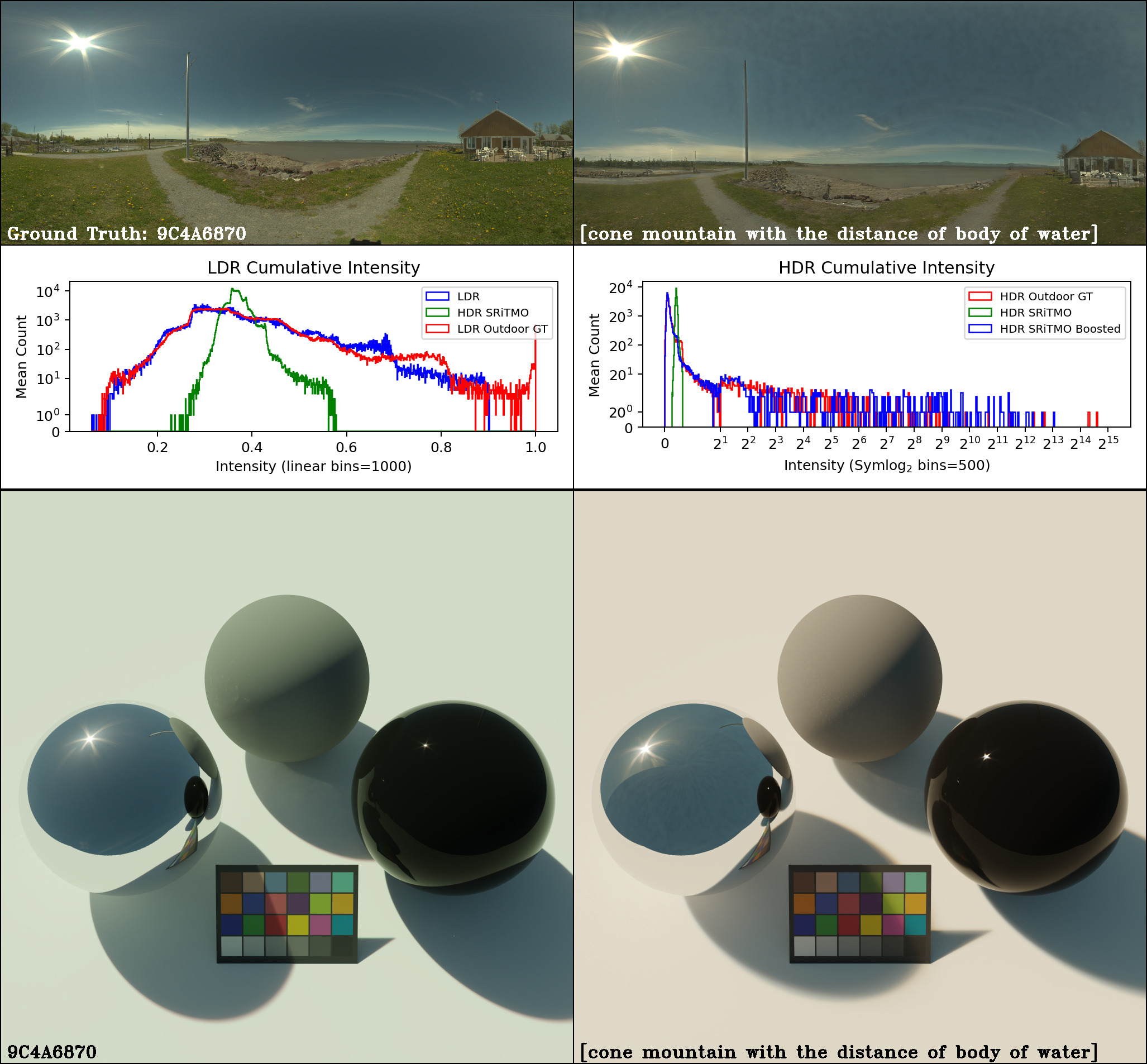}
    \caption{LOD 9C4A6870}
    \label{app:fig::text2light_LOD16}
\end{figure*}

\begin{figure*}
    \centering
    \includegraphics[width=\textwidth,height=.95\textheight,keepaspectratio]{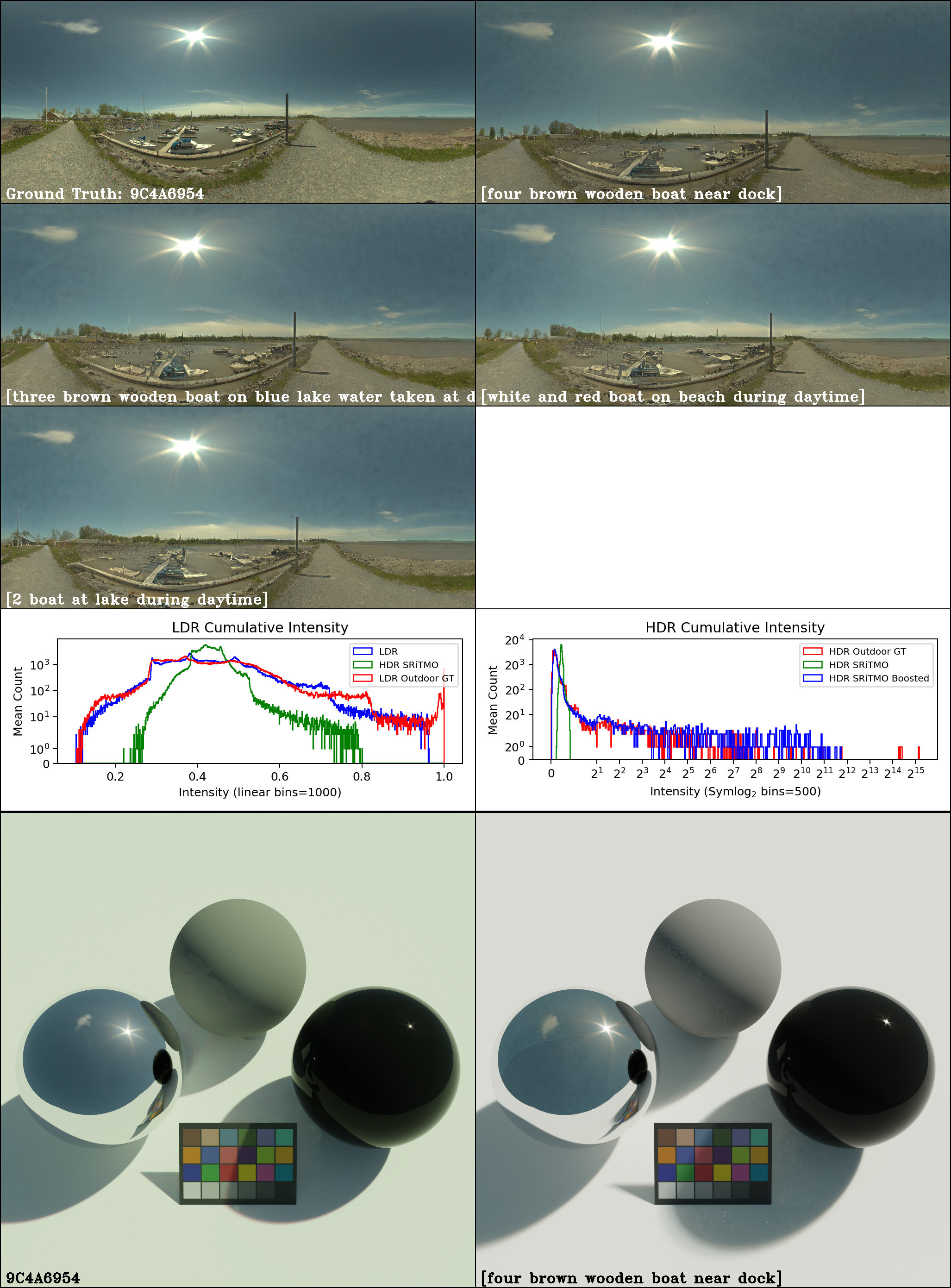}
    \caption{LOD 9C4A6954}
    \label{app:fig::text2light_LOD17}
\end{figure*}

\begin{figure*}
    \centering
    \includegraphics[width=\textwidth,height=.95\textheight,keepaspectratio]{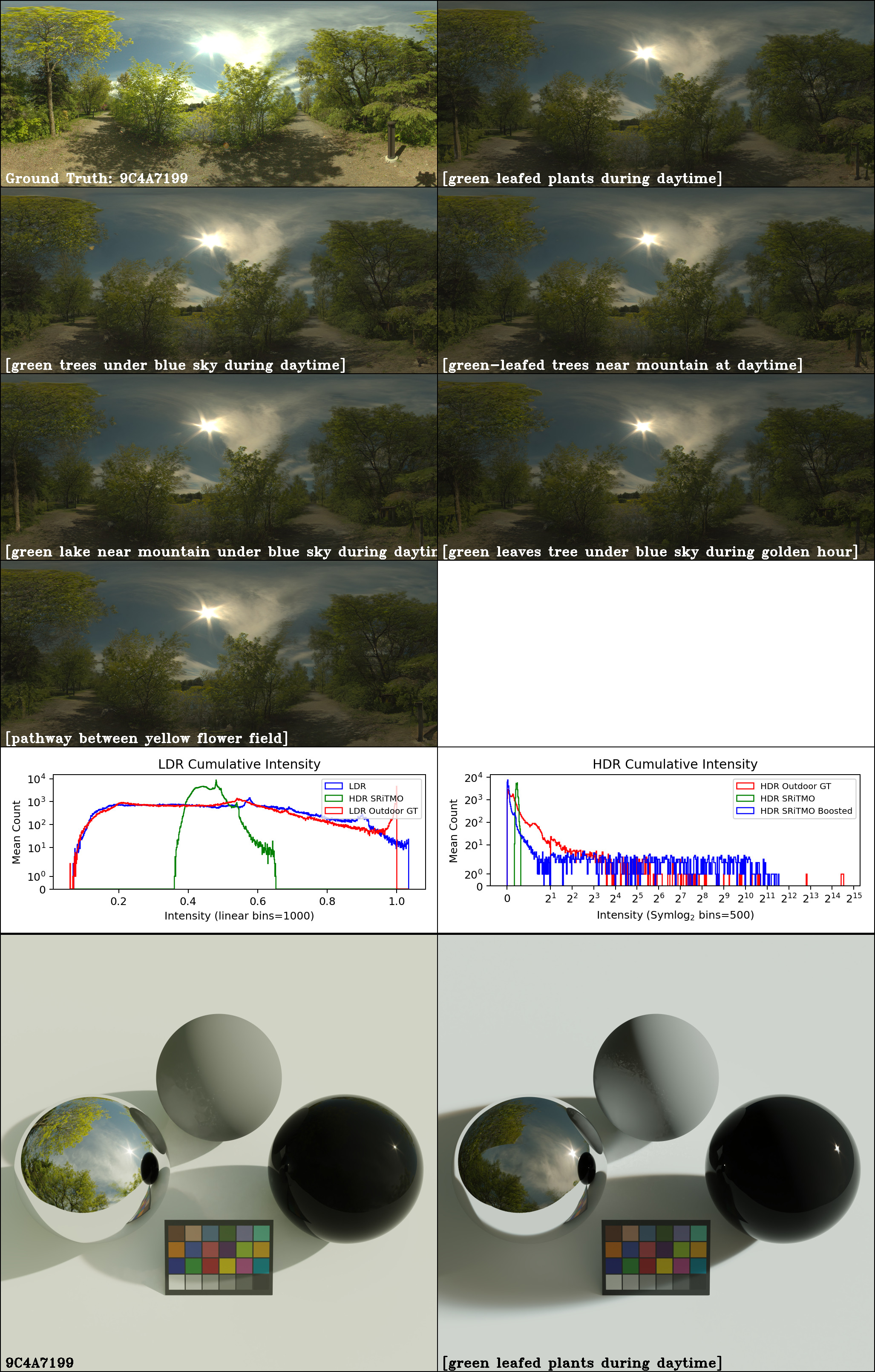}
    \caption{LOD 9C4A7199}
    \label{app:fig::text2light_LOD18}
\end{figure*}

\begin{figure*}
    \centering
    \includegraphics[width=\textwidth,height=.95\textheight,keepaspectratio]{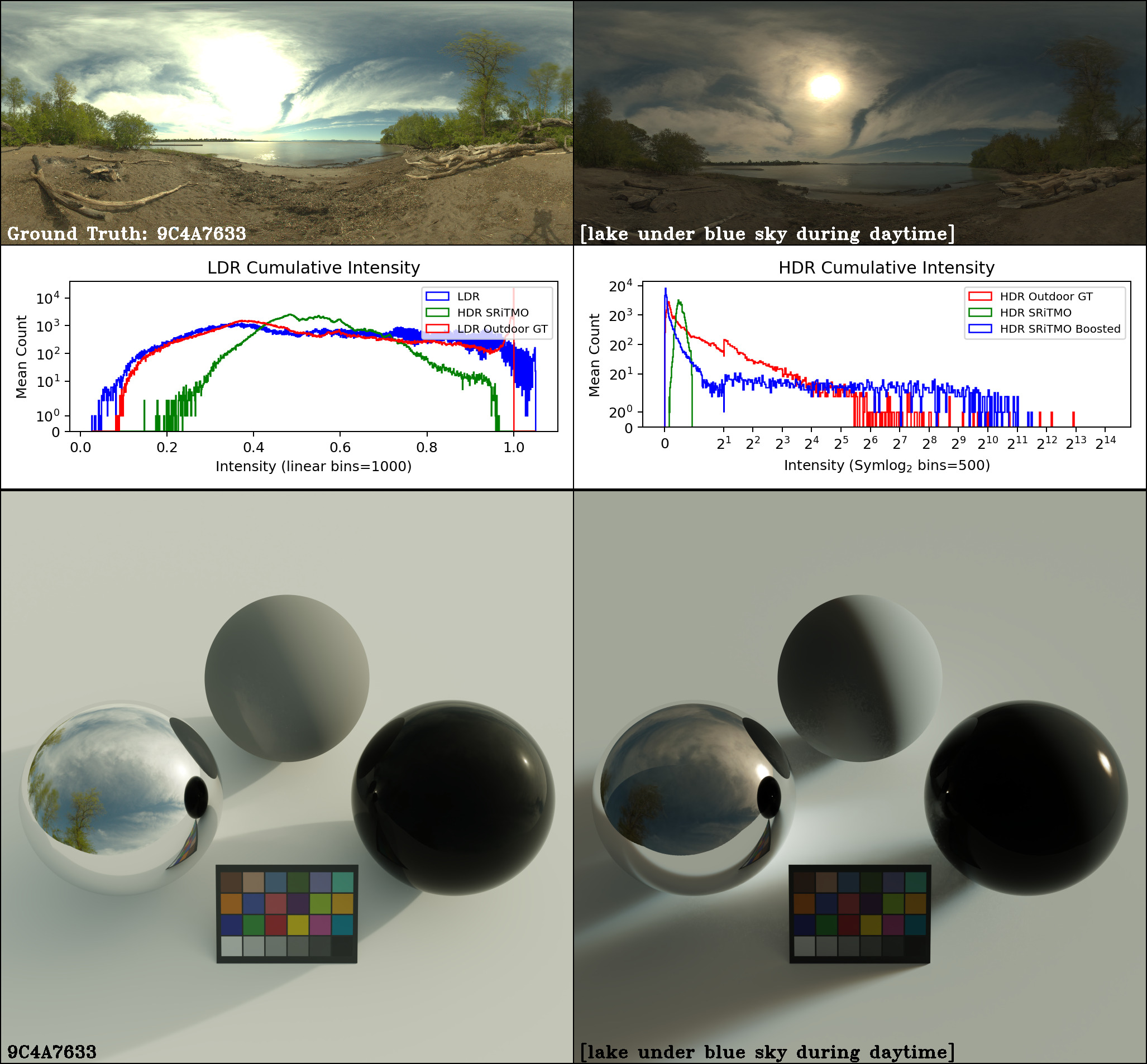}
    \caption{LOD 9C4A7633}
    \label{app:fig::text2light_LOD19}
\end{figure*}

\begin{figure*}
    \centering
    \includegraphics[width=\textwidth,height=.95\textheight,keepaspectratio]{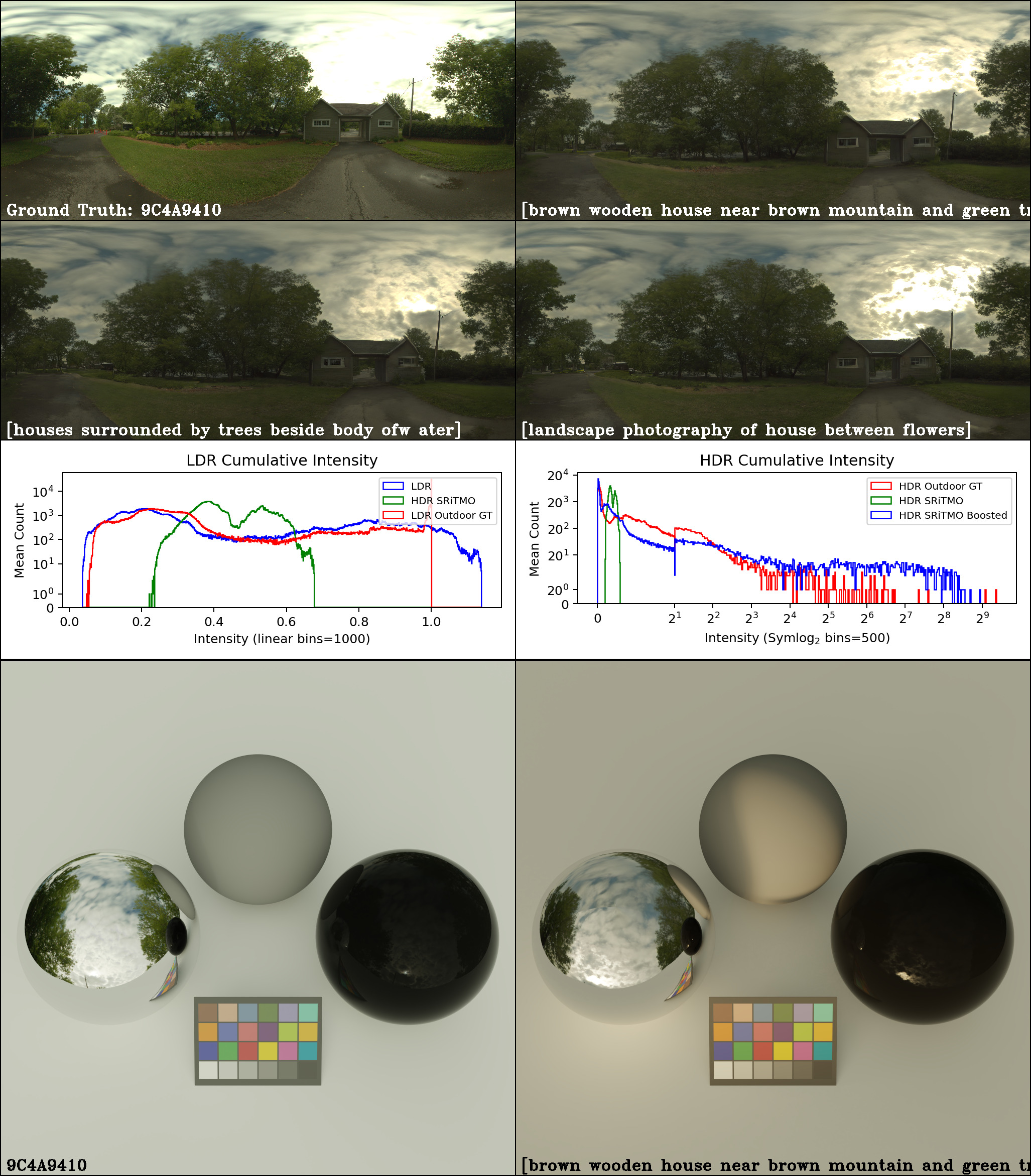}
    \caption{LOD 9C4A9410}
    \label{app:fig::text2light_LOD20}
\end{figure*}

\begin{figure*}
    \centering
    \includegraphics[width=\textwidth,height=.95\textheight,keepaspectratio]{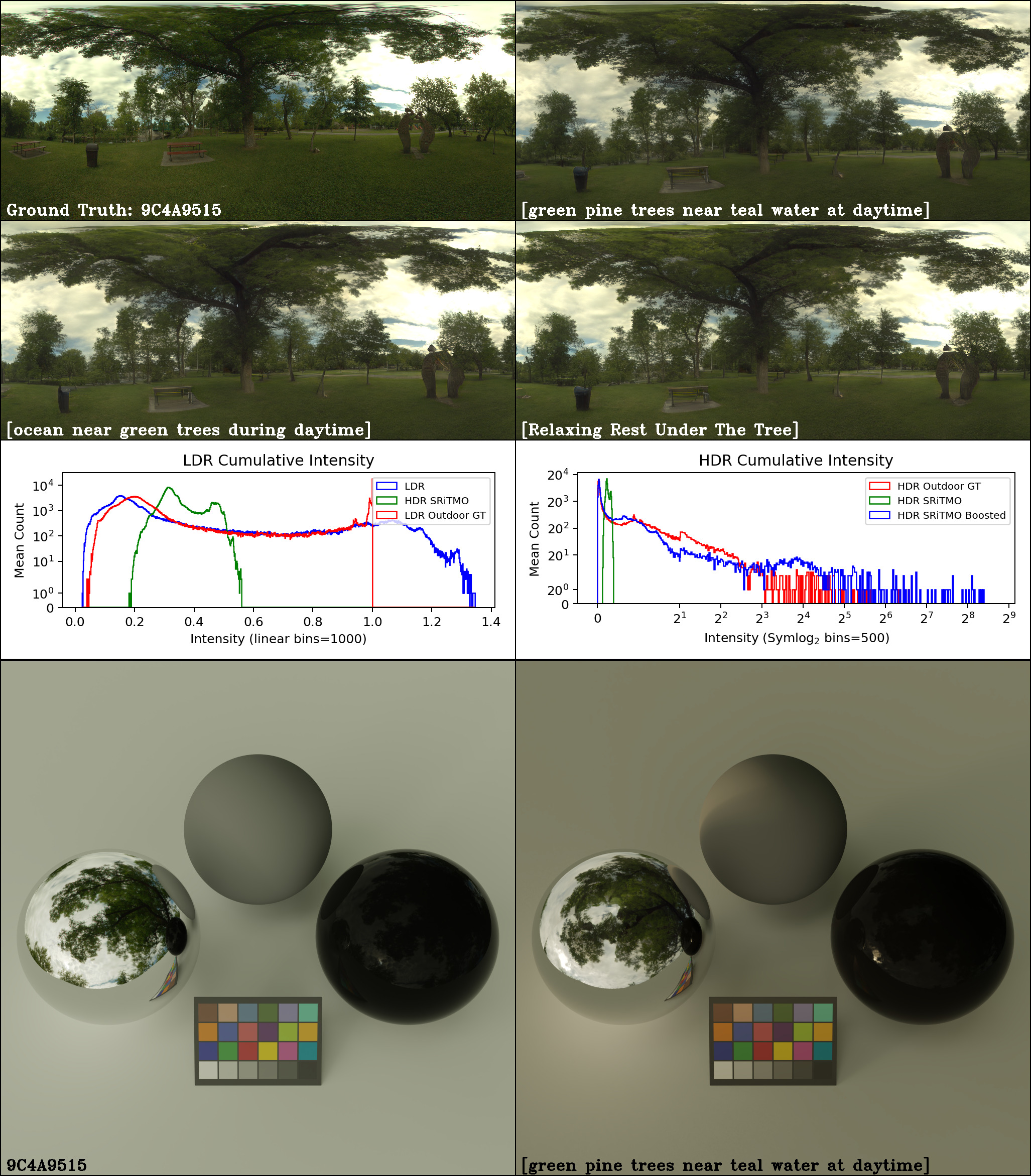}
    \caption{LOD 9C4A9515}
    \label{app:fig::text2light_LOD21}
\end{figure*}

\begin{figure*}
    \centering
    \includegraphics[width=\textwidth,height=.95\textheight,keepaspectratio]{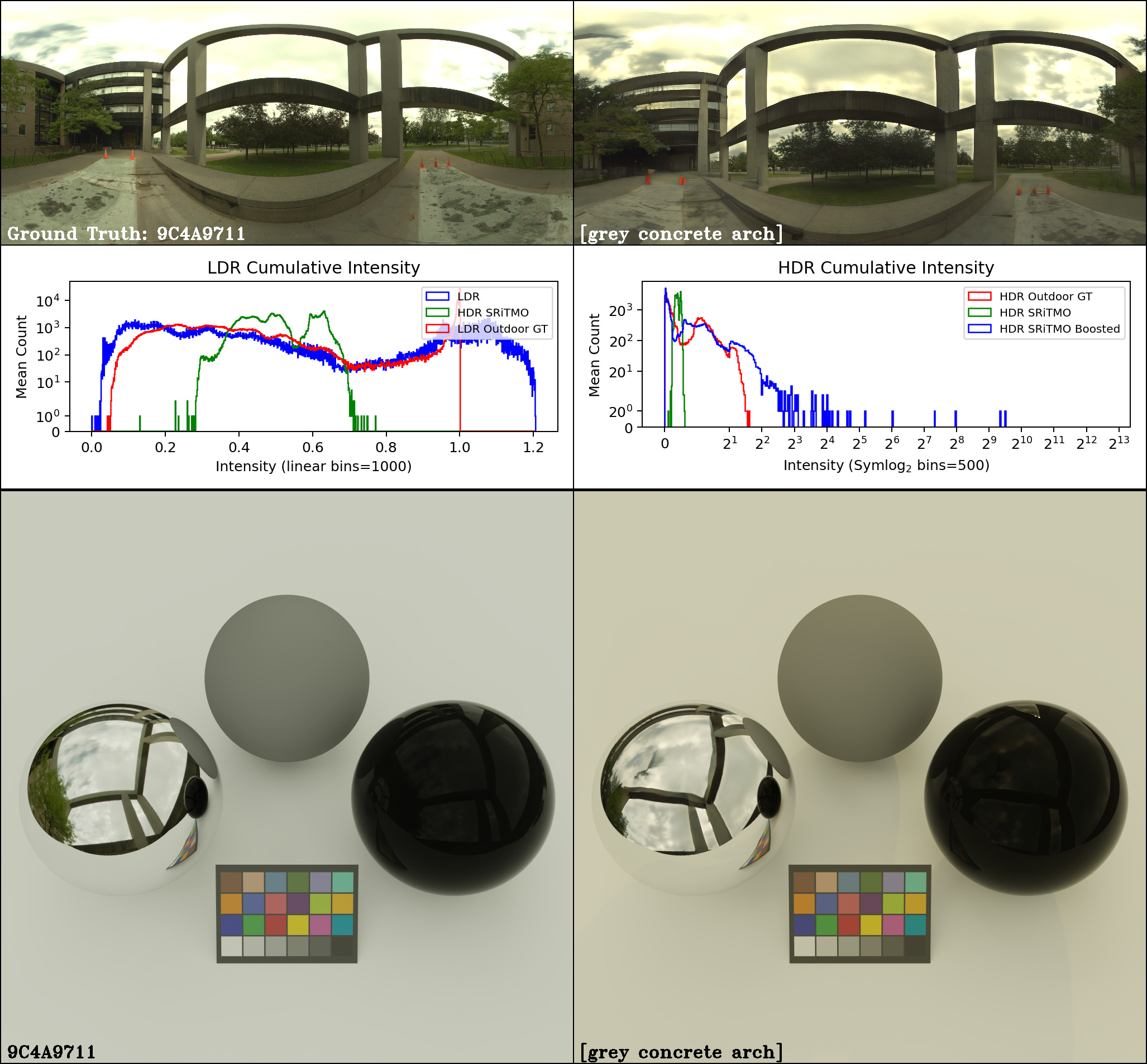}
    \caption{LOD 9C4A9711}
    \label{app:fig::text2light_LOD22}
\end{figure*}

\end{document}